\documentclass{article}
\usepackage{arxiv}
\usepackage[utf8]{inputenc} 
\usepackage[T1]{fontenc}    
\usepackage{hyperref}       
\usepackage{url}            
\usepackage{booktabs}       
\usepackage{amsfonts}       
\usepackage{nicefrac}       
\usepackage{microtype}      
\usepackage{lipsum}
\usepackage{amsthm}
\usepackage{graphicx}
\usepackage{authblk}
{\end{list}}

\def\bi{{\bf i}}

\def\b1{{\bf 1}}

\def\blot{\quad {$\vcenter{\vbox{\hrule height.4pt
             \hbox{\vrule width.4pt height.9ex \kern.9ex \vrule 
width.4pt}
             \hrule height.4pt}}$}}
\usepackage{mathtools}

\usepackage{comment}
\usepackage{multirow}
\usepackage[ruled,norelsize]{algorithm2e}

\SetAlFnt{\small}
\SetAlCapFnt{\small}
\SetAlCapNameFnt{\small}
\SetAlCapHSkip{0pt}
\IncMargin{-\parindent}
\usepackage{algorithmic}
\usepackage{wrapfig}
\usepackage{amsmath}
\allowdisplaybreaks

\renewcommand{\thetheorem}{ \arabic{theorem}}

\newtheorem{definition}{Definition}
\renewcommand{\thedefinition}{\arabic{definition}}
\newtheorem{proposition}{Proposition}

\usepackage{graphicx}
\usepackage{setspace}
\usepackage{verbatim} 
\usepackage[numbers]{natbib} 

\usepackage{latexsym}
\usepackage{amsmath}
\usepackage{nomencl}
\usepackage{amssymb}

\makenomenclature

\usepackage{etoolbox}
\renewcommand\nomgroup[1]{%
  \item[\bfseries
  \ifstrequal{#1}{F}{Functions}{%
  \ifstrequal{#1}{V}{Variables}{%
  \ifstrequal{#1}{S}{Sets}{%
  \ifstrequal{#1}{E}{Empirical Study}}}}%
]}

\usepackage{endnotes}
\let\footnote=\endnote
\let\enotesize=\normalsize
\def\notesname{Endnotes}%
\def\makeenmark{$^{\theenmark}$}
\def\enoteformat{\rightskip0pt\leftskip0pt\parindent=1.75em
	\leavevmode\llap{\theenmark.\enskip}}

\usepackage{amsbsy}    
\usepackage{bbm}
\usepackage{color}
\usepackage{soul}

\begin{document}

\title{Biomanufacturing Harvest Optimization \\with Small Data}

\author[1]{Bo Wang}
\author[ ,1]{Wei Xie \thanks{Corresponding author: w.xie@northeastern.edu}
}
\author[2]{Tugce Martagan}
\author[2]{Alp Akcay}
\author[3]{Bram van Ravenstein}
\affil[1]{Northeastern University, Boston, MA 02115}
\affil[2]{Eindhoven University of Technology,
	5612 AZ Eindhoven, The Netherlands}
\affil[3]{MSD Animal Health,
	5831 AN Boxmeer, The Netherlands}
\maketitle
\begin{abstract}
In biopharmaceutical manufacturing, fermentation processes play a critical role in productivity and profit. A fermentation process uses living cells with complex biological mechanisms, leading to high variability in the process outputs, namely, the protein and impurity levels. By building on the biological mechanisms of protein and impurity growth, we introduce a stochastic model to characterize the accumulation of the protein and impurity levels in the fermentation process. However, a common challenge in the industry is the availability of only a very limited amount of data, especially in the development and early stages of production. This adds an additional layer of uncertainty, referred to as model risk, due to the difficulty of estimating the model parameters with limited data. In this paper, we study the harvesting decision for a fermentation process  (i.e., when to stop the fermentation and collect the production reward) under model risk. We adopt a Bayesian approach to update the unknown parameters of the growth-rate distributions, and use the resulting posterior distributions to characterize the impact of model risk on fermentation output variability. The harvesting problem is formulated as a Markov decision process model with knowledge states that summarize the posterior distributions and hence incorporate the model risk in decision-making. Our case studies at MSD Animal Health demonstrate that the proposed model and solution approach improve the harvesting decisions in real life by achieving substantially higher average output from a fermentation batch along with lower batch-to-batch variability. 
\end{abstract}





\section{Introduction}
\label{sec:introduction}


The biomanufacturing industry has developed several innovative treatments for cancer, adult blindness, and COVID-19 among many other diseases. Despite its increasing success, biomanufacturing is a challenging production environment. Different from classical pharmaceutical manufacturing, biomanufacturing methods use living organisms (e.g., bacteria, viruses, or mammalian cells) during the production processes. These living organisms are custom-engineered to produce highly complex active ingredients for biopharmaceutical drugs. However, the use of living organisms also introduces several operational challenges related to batch-to-batch variability in the production outcomes.

The drug substance manufacturing can be broadly categorized into two main steps: fermentation and purification operations. During the fermentation process, the living organisms grow and produce the desired active ingredients. Specific characteristics of active ingredients (e.g., monoclonal antibodies, biomass, proteins, antigens, etc.) could vary across different drugs. In the remainder of this paper, we refer to the resulting target active ingredient as \textit{protein}. After fermentation, the batch continues with a series of purification operations to comply with stringent regulatory requirements on safety and quality. Our main focus in this study is the fermentation process. 

\begin{figure}
    \centering
    \includegraphics[scale=0.7]
    {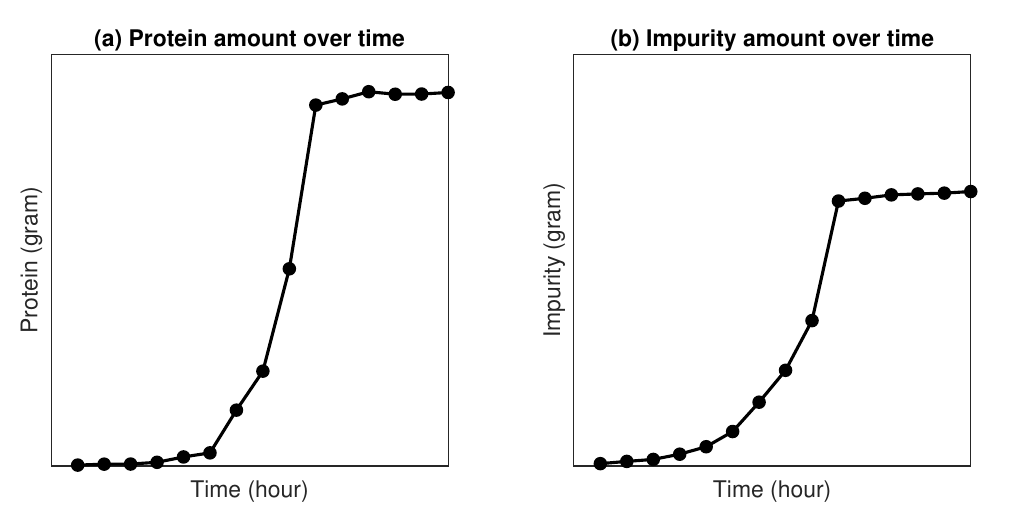}
    \caption{Illustration of fermentation dynamics using industry data from MSD}
    \label{fig:illust}
\end{figure}

The fermentation process is typically carried out inside a stainless steel vessel called \textit{bioreactor}. Bioreactors are equipped with advanced sensors to achieve a highly controlled environment via monitoring of critical process parameters (e.g., cell growth rate, protein accumulation, impurity accumulation, etc.). Figure~\ref{fig:illust} uses industry data to illustrate the main dynamics of a batch fermentation process. 
As the fermentation continues, we observe from Figure~\ref{fig:illust}(a) that the amount of protein produced during fermentation increases exponentially over time. Hence, this specific phase of fermentation is known as the \textit{exponential growth} phase. However, the exponential growth phase continues only for a finite period of time (e.g., several hours or days depending on the application) because of the inherent limitations of biological processes (e.g., limitations in media, cell viability, and growth). After the exponential growth phase, the fermentation enters a \textit{stationary phase} in which the protein production stops and the batch needs to be harvested. In addition, we observe from Figure~\ref{fig:illust}(b) that unwanted \textit{impurities} accumulate inside the bioreactor along with the desired proteins. The specific nature of impurities varies across applications but impurities often represent unwanted byproducts, such as ammonia, dead cells, etc. These impurities are subsequently filtered and eliminated through a series of purification operations. 

\subsection{The Harvesting Problem: Trade-offs and Challenges}

The simultaneous growth of desired proteins and unwanted impurities, as shown in Figure~\ref{fig:illust}, is often known as the \textit{purity--yield trade-off} in fermentation processes. From a practical perspective, the purity--yield trade-off presents a critical challenge in fermentation harvesting (stopping) decisions. In order to achieve a high protein content, the decision maker may be inclined to harvest the fermentation as late as possible. However, waiting too long to harvest can result in higher levels of impurity. As a result, the difficulty (cost) of subsequent purification operations may increase. Therefore, the purity-yield trade-off has financial implications (e.g., expected revenue increases as protein yield increases, but expected cost increases as impurity levels increase). These trade-offs motivate our main research question: \textit{(1) What is an optimal harvesting policy (i.e., when should we stop the fermentation) to maximize the expected profit obtained from a batch?}

In addition to the purity--yield trade-off, process ``uncertainty" imposes another critical challenge on harvesting decisions. In particular, two types of process uncertainty are commonly encountered in biomanufacturing practice: \textit{(i)} inherent stochasticity and \textit{(ii)} model risk. In our problem setting, \textit{inherent stochasticity} represents the uncertainty in the amounts of protein and impurity produced throughout fermentation, and it is often caused by the inherent complexity of biological systems. Because living organisms are used during fermentation, the rate at which proteins and impurities accumulate is random 
(although fermentation is carried out under identical conditions). Therefore, the inherent stochasticity of biological processes motivates our second research question: \textit{(2) How can we develop an analytical model to learn the inherent stochasticity of fermentation processes and incorporate it into optimal harvesting decisions?}

Most often, the inherent stochasticity can not be controlled but can be predicted through historical process data. However, building a reliable prediction model can be challenging when there is only a limited amount of historical process data. 
We refer to the resulting uncertainty in the prediction model itself as the {\it model risk}. The problem of decision-making under limited data (i.e., under model risk) is a critical concern for both research and development (R\&D) projects and industry-scale applications. In biopharmaceutical R\&D projects, each protein is unique such that the scientists re-engineer and manufacture it for the first time. This implies that harvesting decisions are typically made under limited R\&D data. In industry-scale applications, the problem of limited data becomes relevant every time a change occurs in equipment or raw materials. For example, the supplier of raw materials (i.e., medium or seed cells) might change their formulations, the management might purchase a new bioreactor, etc. Such changes have a substantial impact on the output of fermentation which often makes the historical process data obsolete or unreliable. Thus, it is of practical importance to have a harvesting strategy that accounts for the model risk, leading to our third research question:
\textit{ (3) Given a limited amount of historical data, how can we develop a learning mechanism to simultaneously account for inherent stochasticity and model risk while making harvesting decisions?} In addition, the harvesting strategies commonly used in the industry do not account for model risk and may lead to suboptimal decisions. The industry needs a better understanding of how to manage model risk under small data, and how to exploit the structural properties of an optimal policy to facilitate its implementation. 
These observations motivate our final research question: 
{\it (4) What are the structural characteristics of optimal harvesting policies? How does their performance compare to the alternative harvesting policies used in practice?
}

\subsection{Contributions}

\noindent To address the aforementioned research questions, we build a solution framework based on a reinforcement learning model using the theory of Bayesian statistics and Markov decision processes. A key aspect of our work is that we build analytical models that combine the knowledge from life sciences and operations research (OR) to support biomanufacturing decisions under limited historical data and the inherent stochasticity of biological systems. In life sciences research, there are well-known mechanistic models to predict the evolution of fermentation \citep{doran2013bioprocess}. However, existing models do not mathematically capture both aspects of limited process data and the inherent stochasticity of fermentation. Our study is a first attempt to optimize fermentation harvesting decisions in biomanufacturing under limited data,
and combines the knowledge from life sciences and stochastic modeling to derive guidelines that improve industry practices. 

We characterize the control-limit structure of the optimal policy with respect to the impurity level. We also show that the myopic policy which makes the harvesting decisions by looking only one period ahead is optimal under a perfect-information setting and some practically-relevant sufficient conditions. Furthermore, we study how the posterior predictive distributions of the growth rates affect the harvesting decisions under the myopic policy in the presence of model risk. Our framework enables decision makers to do a rigorous assessment of the impact of limited data on harvesting decisions, and provides managerial insights on the value of collecting additional data. This research is an outcome of a multi-year collaboration with MSD Animal Health in Boxmeer, Netherlands. The facility in Boxmeer is a leading biomanufacturing hub in Europe that conducts both biopharmaceutical R\&D and large-scale production. Since September 2019, the developed framework has been used in daily operations to support harvesting decisions. The implementation has resulted in around $50\%$ improvement in batch yield on average. The research outcomes have been recognized as a finalist of the 2022 INFORMS Franz Edelman Prize. 
 


\section{Literature Review}
\label{sec:background}

There is a wide range of papers in operations management with a focus on pharmaceutical industry. For example, \cite{plante1999improving} maximized the expected quality in a pharmaceutical production process by explicitly modeling the quality parameters of raw materials, the parameters of the production process, and the interactions between them. 
In real-world case studies, \cite{martaganPOM} optimized the purification-related decisions for engineer-to-order proteins, and \cite{sahling2019dynamic} determined a master production schedule for weekly demand for a multi-level biopharmaceutical manufacturing process. \cite{subramanian2020pulling} investigated how pharmaceutical manufacturers switch from early‐stage drug discovery and late‐stage drug development. At supply chain level, \cite{zhu2021demand} proposed a forecasting method to predict demand for pharmaceutical products, and \cite{zhao2023pharmaceutical} provided an overview of complex pharmaceutical supply chains. Also at the supply chain level, \cite{xu2023not} investigated a pharmaceutical manufacturer's usage of different types of distributors for speciality drugs. The operations management literature on food processing and agribusiness is also relevant to our work due to the biological nature of the products \citep{lowe2004decision, azoury2013managing, bansal2017product}. For example, \cite{rajaram2004campaign} considered the scheduling of multiproduct batch operations in the food-processing industry to minimize setup and quality costs, and \cite{jahandideh2020production} considered learning across batches and decay in the performance of the catalysts used in the production process. \cite{blackburn2009supply} studied supply chain design strategies for fresh produce after their harvesting. More recently, \cite{ARC2024} addressed a crop harvesting problem in the presence of a time-quality
trade-off. In the remainder of our review, we restrict our scope to studies related to biomanufacturing applications. In particular, our work is most closely related to two streams of research: (1) modeling and control of fermentation based on a {\it known} model that describes the dynamics of the fermentation process, and (2) reinforcement learning approaches to predict and control fermentation processes. 

A vast body of life sciences literature focuses on modeling the biological dynamics of fermentation processes. In particular, predictive models are built to estimate the evolution of fermentation, and then these models are used to guide the search for optimal control strategies. In this context, most studies develop deterministic or stochastic models to predict and control fermentation. Deterministic models typically build kinetic process models (i.e., differential equations) on cell growth and product formation \citep{McNeil08, doran2013bioprocess, Putra18}. These kinetic models are also integrated with optimization models. For example, \cite{chang2016nonlinear} constructed a dynamic flux balance model for a fermentation process. They developed a closed-loop control for feed rate and dissolved oxygen concentration profiles to maximize yield production. 

Data-driven stochastic optimization is relatively understudied to predict and control fermentation processes. Existing studies typically focus on the inherent stochasticity of fermentation. For example, 
\cite{Peroni05} used approximate dynamic programming to maximize yield and minimize process time in fed-batch fermentation. \cite{xing2010modeling} adopted a Markov chain Monte Carlo approach to optimize the kinetics of a fermentation process. More recently, \cite{martagan2016optimal} and \cite{martagan2018managing} developed a Markov decision processes (MDP) model to optimize fermentation operating decisions. However, their optimization models are built based on sufficiently large historical data, and hence are not equipped to capture the impact of model risk (limited historical data) on biomanufacturing decisions. \cite{martagan2022} developed a portfolio of decision support tools to reduce biomanufacturing costs. \cite{koca2023increasing} optimized the timing of so-called bleed-feed decisions, enabling pharmaceutical manufacturers skip intermediary biorector setups in batch fermentation processes. However, they did not consider the model risk.  \cite{xie2022interpretable} considered model risk in an interpretable semantic bioprocess probabilistic knowledge graph for production stability control, but they do not optimize the harvesting decisions. To the best of our knowledge, this paper is the first to simultaneously capture the inherent stochasticity of biological systems and model risk to optimize harvesting decisions in biomanufacturing systems.


Reinforcement learning approaches have been recently developed for bioprocess control. For example, \cite{treloar2020deep} and \cite{nikita230reinforcement} developed model-free deep-Q-network-based reinforcement learning approaches to maintain the cells at target populations by controlling the feeding profiles and maximize the yield by controlling the flow rate. \cite{zheng2020paper} and \cite{Zheng2023_JoC} constructed a model-based reinforcement learning for biomanufacturing control, with a predictive distribution of system response. After that, \cite{zheng2020paper} proposed a simulation-assisted policy gradient algorithm that can efficiently reuse the previous process outputs to facilitate the learning and the search for the optimal policy. Our paper is different from these studies 
in two ways. First, we study the harvesting problem in fermentation, while \cite{zheng2020paper} focus on a chromatography problem. Second, 
we explicitly incorporate the posterior distributions of unknown fermentation-process parameters as knowledge states of the MDP model and study the structure property of optimal policy, which can enhance interpreability and feasibility of the proposed approach for real manufacturing practice.

To be specific, we adopt model-based Bayesian reinforcement learning as our solution approach. Compared to model-free approaches, it allows us to interpret model risk by quantifying the uncertainty in the process parameters. Furthermore, although it is often less computationally efficient than a model-free approach, it can incorporate known process dynamics and any prior information about model risk into decision making. A comprehensive review of Bayesian reinforcement learning methodologies can be found in \cite{ghavamzadeh2016bayesian}.
Since solving the original model-based Bayesian reinforcement learning is notoriously complex due to potentially huge state space, various approximation algorithms have been developed, including offline value approximation \citep{poupart2006analytic}; online near-myopic value and tree search approximation that focus on realized knowledge states in planning \citep{ross2008bayesian,osband2013more,fonteneau2013optimistic}; and exploration bonus based methods where an agent acts according to an optimistic model of MDP under uncertainty \citep{kolter2009near,asmuth2011approaching,asmuth2012bayesian}. 
Motivated by those studies, we develop a model-based Bayesian reinforcement learning approach, which can account for model risk in guiding fermentation harvesting decisions.

\section{Model}
\label{sec:HybridModeling}

Section~\ref{subsec:hybridModeling} introduces a stochastic mechanistic model to represent the protein and impurity accumulation in the fermentation process. Section~\ref{subsec:onlineLearning} presents a Bayesian approach to capture the uncertainty in the unknown parameters of this model, and describes how this uncertainty can be updated with new protein and impurity observations collected during the fermentation process. Finally, Section~\ref{sec:RL-MR} presents an MDP model formulation, accounting for both inherent stochasticity and model risk, to optimize the harvesting decision in the fermentation process. An overview of the mathematical notation is provided in Appendix~\ref{sec: notationtable}.

\subsection{Fermentation Process Modeling}
\label{subsec:hybridModeling}

The accumulation of protein and impurity amount in the exponential-growth phase of a fermentation process is commonly modeled with the so-called {\it cell-growth kinetics} mechanism \citep{doran2013bioprocess}. The cell-growth kinetics mechanism is often represented as an ordinary differential equation. 
To be specific, the protein amount at time $t$, denoted by $p_t$, is given by the functional form  $\dfrac{dp_t}{dt} = \phi p_t$, where $\phi$ is referred to as the \textit{specific growth rate} of protein. Therefore, it is common to assume that the protein amount at time~$t$ follows the functional form $p_t =p_0 \: e^{\phi t}$, where $p_0$ is the starting amount of protein (seed). Similarly, the impurity amount at time~$t$ follows the functional form $i_t = i_0 \: e^{\psi t}$, where $\psi$ is the specific growth rate of impurity and $i_0$ is the starting amount of impurity. In our model, we consider that measurements are performed at discrete time points $\mathcal{T}=\{t: 0, 1, \ldots, T\}$, where $T$ denotes the time point chosen by the decision maker to harvest the fermentation process. The time between two measurements is fixed as one time unit and it can be any finite amount of time (e.g., an hour, a day, or longer), depending on the process characteristics and practical constraints. This leads to a recursive representation $p_{t+1} = p_t e^{\phi}$ for the protein amount and $i_{t+1} = i_t e^{\psi}$ for the impurity amount for $t \in \{0,1,\ldots,T-1\}$. 

Because living biological systems (e.g., cells) are used in the fermentation process, their specific growth rates are random; see for example \cite{templeton2013peak} and \cite{odenwelder2021induced}.
We let $\Phi_t$, $t \in \{0,1,\ldots,T-1\}$, denote the normally distributed independent random variables that represent the specific growth rates of protein in the time interval from time $t$ to $t+1$.  
Let $\mu^{(p)}_c$ and $\sigma^{(p)2}_c$ denote the true (unknown) mean and the variance of $\Phi_t$. Similarly, the random variables $\Psi_t$, $t \in \{0,1,\ldots,T-1\}$, are independent and normally distributed with true mean $\mu^{(i)}_c$ and variance $\sigma^{(i)2}_c$. The modeling of growth rates as normally distributed random variables leads to the recursive representations of the protein and impurity amount given by 
\begin{align}
\label{eq.growthmodel}
\begin{split}
		p_{t+1} & = p_t\cdot e^{\Phi_{t}}, ~~\Phi_{t} \sim \mathcal{N}(\mu^{(p)}_c, \sigma^{(p)2}_c) \\
		i_{t+1} & = i_t\cdot e^{\Psi_{t}}, ~~\Psi_{t} \sim \mathcal{N}(\mu^{(i)}_c, \sigma^{(i)2}_c)
\end{split}
\end{align}
for $t \in \{0,1,\ldots,T-1\}$, accounting for the \textit{inherent stochasticity} in the accumulation of  protein and impurity  in a fermentation process. In \eqref{eq.growthmodel}, we use the notation $\sim$ to mean ``distributed as'' and $\mathcal{N}(a,b)$ to represent a normal distribution with mean $a$ and variance $b$.

The protein and impurity growth rates are assumed independent because there are many biological and chemical factors that randomly influence the ``production speed'' for impurity and protein, i.e., the rate of generating metabolic wastes and antibody proteins \citep{tsao2005monitoring,xing2010modeling}. The independent and normally distributed growth rates are commonly used in the literature to model their random variations over time \citep{wechselberger2013model, mockus2015batch, moller2020model}. We also validated this assumption using historical data in our case study, as described in Section~\ref{subsec:overview}. The stationarity assumption for the growth rate distributions is linked to the fact that the fermentation process has a well-controlled cell culture condition, where the so-called metabolic quasi-steady state is achieved. Thus, the metabolic flux (and hence the corresponding distribution of the protein and impurity growth rates during the exponential growth phase) does not change over time. This assumption is also validated in our case study (Section~\ref{subsec:overview}).

\subsection{Bayesian Learning for the Fermentation Process}
\label{subsec:onlineLearning}

The true parameters of the underlying stochastic model for the protein and impurity growth rates, denoted by  $\pmb{\theta}^c = \{\mu^{(p)}_c, \sigma^{(p)2}_c, \mu^{(i)}_c, \sigma^{(i)2}_c \}$, are unknown and often need to be estimated from a very limited amount of real-world data (especially for new products that are not yet in production). We adopt a Bayesian approach and model the mean and variance of the protein and impurity growth rates as random variables, denoted by $\pmb{\theta} = \{\mu^{(p)}, \sigma^{(p)2}, \mu^{(i)}, \sigma^{(i)2} \}$. 
In the remainder of this section, we describe how we specify the prior distribution for $\pmb{\theta}$, obtain its posterior distribution by updating the prior with new data on protein and impurity accumulation, and characterize the posterior predictive distributions for the protein and impurity growth rates. 

\noindent \textbf{Specification of prior distribution.} For the protein growth rate, we build the joint {\it prior} distribution of $(\mu^{(p)}, \sigma^{(p)2})$ in the following way. First, the marginal distribution of variance $\sigma^{(p)2}$ is chosen as an inverse-gamma distribution with prior parameters $\lambda_{0}^{(p)}$ and $\beta_{0}^{(p)}$, denoted as $\sigma^{(p)2}\sim {\mbox{Inv}\Gamma}(\lambda_{0}^{(p)}, \beta_{0}^{(p)})$. Next, given the value of $\sigma^{(p)2}$, the conditional distribution of mean $\mu^{(p)}$ is assumed to be $\mathcal{N}(\alpha_{0}^{(p)}, \sigma^{(p)2}/\nu_{0}^{(p)})$, where $\alpha_{0}^{(p)}$ and $\nu_{0}^{(p)}$ are also prior parameters. It then follows that the joint  prior distribution of $(\mu^{(p)}, \sigma^{(p)2})$ has a normal-inverse-gamma distribution \citep{Gelman_2004}, i.e., 
\begin{equation}
		\left(\mu^{(p)}, \sigma^{(p)2} \right) \sim \mathcal{N}(\alpha_{0}^{(p)}, \sigma^{(p)2}/\nu_{0}^{(p)})\cdot {\mbox{Inv}\Gamma}(\lambda_{0}^{(p)}, \beta_{0}^{(p)}).
	\label{eq.prior}
\end{equation}
For the impurity growth rate, we obtain the joint prior distribution of $(\mu^{(i)}, \sigma^{(i)2})$ in a similar way by using the prior parameters $\alpha_{0}^{(i)}$, $\nu_{0}^{(i)}$, $\lambda_{0}^{(i)}$ and $\beta_{0}^{(i)}$; i.e., 
\begin{equation}
		\left(\mu^{(i)}, \sigma^{(i)2} \right) \sim  \mathcal{N}(\alpha_{0}^{(i)}, \sigma^{(i)2}/\nu_{0}^{(i)})\cdot {\mbox{Inv}\Gamma}(\lambda_{0}^{(i)}, \beta_{0}^{(i)}).
	\label{eq.priorImp}
\end{equation}
It is well known that normal-inverse-gamma distribution is conjugate when combined with normally distributed observations \citep{Gelman_2004, powell2012optimal}. This enables us to efficiently update the prior distributions with the arrival of new data to obtain the posterior distributions. 



\noindent \textbf{Characterization of the posterior distribution.} Suppose that $(\alpha_{t}^{(p)}, \nu_{t}^{(p)}, \lambda_{t}^{(p)}, \beta_{t}^{(p)})$ represent our belief on the distribution of protein growth rate at time point $t$, and we make an observation of the protein amount $p_{t+1}$ at time point $t+1$. That is, we make an observation $\phi_{t} = \ln (p_{t+1}/p_t)$ as the realization of the normally distributed protein growth rate $\Phi_t$. Then, the posterior distribution of $(\mu^{(p)}, \sigma^{(p)2})$ follows a normal-inverse-gamma distribution, i.e., 
\begin{equation}
		\left( \mu^{(p)}, \sigma^{(p)2} \right) \sim \mathcal{N}(\alpha_{t+1}^{(p)}, \sigma^{(p)2}/\nu_{t+1}^{(p)})\cdot {\mbox{Inv}\Gamma}(\lambda_{t+1}^{(p)}, \beta_{t+1}^{(p)}) 
\end{equation}
with updated parameters
\begin{equation}
		\alpha_{t+1}^{(p)} = \alpha_{t}^{(p)} +  \dfrac{\phi_t - \alpha_{t}^{(p)}}{\nu_{t+1}^{(p)}} ,~ \nu_{t+1}^{(p)} = \nu_{t}^{(p)} + 1,~
		\lambda_{t+1}^{(p)} = \lambda_{t}^{(p)} + \dfrac{1}{2},~ \beta_{t+1}^{(p)} = \beta_{t}^{(p)} +  \dfrac{\nu_{t}^{(p)}(\phi_t - \alpha_{t}^{(p)})^2}{2\nu_{t+1}^{(p)}}.
\label{eq.update2}
\end{equation}


Given our belief $(\alpha_{t}^{(i)}, \nu_{t}^{(i)}, \lambda_{t}^{(i)}, \beta_{t}^{(i)})$ about the distribution of impurity growth rate at time point $t$, since the  observation  $\psi_{t} = \ln (i_{t+1}/i_t)$ is a {\it random} realization of the normally distributed impurity growth rate $\Psi_t$,  the posterior distribution of $(\mu^{(i)}, \sigma^{(i)2})$ is also a normal-inverse-gamma; i.e.,
\begin{equation}
		\left( \mu^{(i)}, \sigma^{(i)2} \right) \sim \mathcal{N}(\alpha_{t+1}^{(i)}, \sigma^{(i)2}/\nu_{t+1}^{(i)})\cdot {\mbox{Inv}\Gamma}(\lambda_{t+1}^{(i)}, \beta_{t+1}^{(i)}),
\end{equation}
with the updated parameters
\begin{equation}
		\alpha_{t+1}^{(i)} = \alpha_{t}^{(i)} +  \dfrac{\psi_t - \alpha_{t}^{(i)}}{\nu_{t+1}^{(i)}} ,~ \nu_{t+1}^{(i)} = \nu_{t}^{(i)} + 1,~
		\lambda_{t+1}^{(i)} = \lambda_{t}^{(i)} + \dfrac{1}{2},~ \beta_{t+1}^{(i)} = \beta_{t}^{(i)} +  \dfrac{\nu_{t}^{(i)}(\psi_t - \alpha_{t}^{(i)})^2}{2\nu_{t+1}^{(i)}}. 
\label{eq.update3}
\end{equation}
For further details on the Bayesian update procedure, we refer the reader to \cite{Gelman_2004}.

\noindent \textbf{Posterior predictive distribution of the growth rates.} Given the Bayesian model described above, the density of the  protein growth rate at time $t$, conditional on the historical protein data (i.e., summarized by the belief parameters $(\alpha_{t}^{(p)}, \nu_{t}^{(p)}, \lambda_{t}^{(p)}, \beta_{t}^{(p)})$), is given by 
\begin{equation}
\label{eq: predPhi}
\mbox{p} \left( \phi_{t} |  \alpha_{t}^{(p)}, \nu_{t}^{(p)}, \lambda_{t}^{(p)}, \beta_{t}^{(p)} \right) = \int \int \mbox{p} \left( \phi_{t} | \mu^{(p)}, \sigma^{(p)2} \right) \mbox{p}(  \mu^{(p)}, \sigma^{(p)2} | \alpha_{t}^{(p)}, \nu_{t}^{(p)}, \lambda_{t}^{(p)}, \beta_{t}^{(p)}) 
d\mu^{(p)} d \sigma^{(p)2},
\end{equation}
where $\mbox{p} \left( \phi_{t} | \mu^{(p)}, \sigma^{(p)2} \right)$ is the density of the normally distributed protein growth rate $\Phi_t$ and $\mbox{p}(  \mu^{(p)}, \sigma^{(p)2} | \alpha_{t}^{(p)}, \nu_{t}^{(p)}, \lambda_{t}^{(p)}, \beta_{t}^{(p)})$ is the joint posterior density of the normal-inverse-gamma distributed $(\mu^{(p)}, \sigma^{(p)2})$. In \eqref{eq: predPhi},  the integral marginalizes out the variables $\mu^{(p)}$ and $\sigma^{(p)2}$, leading to the predictive density of the future observation of the protein growth rate given the current belief parameters on the underlying protein growth model. We let $\widetilde{\Phi}_{t}$ denote the predictive protein growth-rate random variable at time point $t$, and it has the density given in \eqref{eq: predPhi}, accounting for both the model risk and the inherent stochasticity in the protein accumulation.  It can be shown that the random variable 
$\widetilde{\Phi}_{t}$ follows a generalized t-distribution \citep{murphy2007conjugate}, i.e.,
\begin{equation}
\widetilde{\Phi}_{t} \sim t_{2\lambda_{t}^{(p)}} \left( \alpha_{t}^{(p)}, \dfrac{\beta_{t}^{(p)}(1+\nu_{t}^{(p)})}{\nu_{t}^{(p)}\lambda_{t}^{(p)}} \right), \label{eq.predictive}
\end{equation}
where $\widetilde{\Phi}_{t} \sim t_{v}(a,b)$ means that $(\widetilde{\Phi}_{t}-a)/\sqrt{b}$ follows a standard t-distribution with $v$ degrees of freedom. We refer to the term $\beta_{t}^{(p)}(1+\nu_{t}^{(p)})/(\nu_{t}^{(p)}\lambda_{t}^{(p)})$ in \eqref{eq.predictive} as the predictive variance of the protein growth rate, and denote it with  $\widetilde{\sigma}_{t}^{(p)2}$.

The same result holds for the predictive impurity growth-rate  $\widetilde{\Psi}_{t}$ at time point $t$:
\begin{equation}
\widetilde{\Psi}_{t} \sim t_{2\lambda_{t}^{(i)}} \left( \alpha_{t}^{(i)}, \dfrac{\beta_{t}^{(i)}(1+\nu_{t}^{(i)})}{\nu_{t}^{(i)}\lambda_{t}^{(i)}} \right), \label{eq.predictiveImp}
\end{equation}
where we refer to the term $\beta_{t}^{(i)}(1+\nu_{t}^{(i)})/(\nu_{t}^{(i)}\lambda_{t}^{(i)})$ in \eqref{eq.predictiveImp} as the predictive variance of the impurity growth rate and denote it with  $\widetilde{\sigma}_{t}^{(i)2}$. In the remainder of the paper, we use $f_t^{(p)}(\cdot)$ and $f_t^{(i)}(\cdot)$ to denote the posterior predictive density functions of the random variables $\widetilde{\Phi}_{t}$ and $\widetilde{\Psi}_{t}$, respectively.

\subsection{Markov Decision Process Model}
\label{sec:RL-MR}

It is of practical importance to optimize when to harvest the fermentation process under limited historical data. We will formulate this problem as a Markov Decision Processes (MDP) model with Bayesian updates on the parameters of the protein and impurity growth-rate distributions. 

\noindent \underline{Decision Epochs.}
%
We consider a finite-horizon discrete-time model with decision epochs $\mathcal{T}=\{t: 0, 1, \ldots, \bar{T}\}$, representing the time points at which the protein and impurity amounts are measured.\endnote{Although fermentation is a continuous process, we adopt a discrete-time model with measurement times as decision epochs  due to practical constraints (e.g., planning and scheduling requirements), as discussed in Appendix~\ref{sec:SensitivityN}.} The parameter $\bar{T}$ is the time point at which the fermentation must be harvested, if not done yet. We consider $\bar{T}$ as an upper bound on the time of harvest because it is often known when the growth stops (i.e., there are no incentives for continuing the fermentation beyond that point). Also, it gives some level of certainty in the planning of the bioreactor. Note that $T \in \mathcal{T}$, i.e., the time point at which the fermentation is harvested must be a decision epoch and it is at most $\bar{T}$.

\vspace{0.in}
\noindent \underline{Physical States.} The levels of protein and impurity during the fermentation process constitute the physical states. In practice, there is an upper limit on the cell density that can be accommodated by a bioreactor with a certain volume. Thus, it is undesired to continue fermentation beyond a certain level of protein accumulation. We let $\bar{P}$ represent this upper limit on the accumulated protein level at which the fermentation must be harvested. On the other hand, we let $\bar{I}$ denote the maximum impurity value at which the batch is considered as failed. If the accumulated impurity level reaches $\bar{I}$, a predefined value in accordance with regulatory standards on batch quality, the fermentation process must be terminated. At decision epoch $t$, the physical state $\mathcal{S}_t$ is specified by the current protein amount $p_t \in [0,\bar{P}]$ and the current impurity amount $i_t \in [0,\bar{I}]$ in the fermentation process, i.e., $\mathcal{S}_t = (p_t, i_t)$.

\vspace{0.in}
\noindent \underline{Action Space.} 
 At a decision epoch before reaching the stationary phase, we can either continue the fermentation process one more time period (denoted by action $C$) or terminate the fermentation process by harvesting it (denoted by action $H$). The harvest action is the only possible action if: (1) the current protein amount reaches the harvesting limit $\bar{P}$; (2) there is a batch failure, caused by the impurity level reaching the threshold level $\bar{I}$; or (3) the fermentation process reaches the decision epoch $\bar{T}$. The action space can be formalized as $\{H \}$ if $p_t =  \bar{P}$ or $i_t = \bar{I}$ or $t=\bar{T}$. On the other hand, the action space is given by  $\{C, H \}$ if $p_t < \bar{P}$, $i_t < \bar{I}$, and $t<\bar{T}$.

\noindent \underline{Knowledge State.} Since the true parameters $\pmb{\theta}^c$ of the underlying model are unknown and estimated from real-world data, we use the \textit{knowledge state}, specified by the parameters of the posterior distribution of $\pmb{\theta}$, to quantify our current belief about $\pmb{\theta}^c$. 
That is, we specify the posterior-distribution parameters (from Section~\ref{subsec:onlineLearning}) as the knowledge state at decision epoch $t$, denoted by $\mathcal{I}_t = \{ \alpha_{t}^{(p)}, \nu_{t}^{(p)}, \lambda_{t}^{(p)}, \beta_{t}^{(p)}, \alpha_{t}^{(i)}, \nu_{t}^{(i)}, \lambda_{t}^{(i)}, \beta_{t}^{(i)} \}$.

\noindent \underline{Hyper States \& Hyper State Transition:} We introduce the \textit{hyper states} $\mathcal{H}_t \equiv (\mathcal{S}_t, \mathcal{I}_t)$, including both physical state $\mathcal{S}_{t} = (p_{t},i_{t})$ and knowledge state $\mathcal{I}_t$. If the action at decision epoch $t$ is to continue the fermentation, i.e., $a_t = C$, the hyper state transition probability can be specified as
\begin{equation}
\mbox{Pr}(\mathcal{S}_{t+1}, \mathcal{I}_{t+1} | \mathcal{S}_t, \mathcal{I}_t; C) = \mbox{Pr}(\mathcal{S}_{t+1} | \mathcal{S}_t, \mathcal{I}_t) \mbox{Pr}( \mathcal{I}_{t+1} | \mathcal{S}_{t+1}, \mathcal{S}_t, \mathcal{I}_t) \label{eq.hyper_transition}
\end{equation}
where $\mbox{Pr}(\mathcal{S}_{t+1} | \mathcal{S}_t, \mathcal{I}_t)$ represents the probability that the physical state transits to $\mathcal{S}_{t+1}$ (i.e., conditioned on the current physical state and knowledge state), and $\mbox{Pr}( \mathcal{I}_{t+1} | \mathcal{S}_{t+1}, \mathcal{S}_t, \mathcal{I}_t, C)$ represents the probability that the knowledge state transits to $\mathcal{I}_{t+1}$ given the realization of $\mathcal{S}_{t+1}$ (i.e., conditioned on the current knowledge state as well as the realized physical-state transition).

In \eqref{eq.hyper_transition}, the first term $\mbox{Pr}(\mathcal{S}_{t+1} | \mathcal{S}_t, \mathcal{I}_t)$ 
can be determined by the protein and impurity transition equations $p_{t+1}  =  p_t e^{\widetilde{\phi}_t}$ and $i_{t+1}  =  i_t e^{\widetilde{\psi}_t}$, where $\widetilde{\phi}_t$ and $\widetilde{\psi}_t$ are the realizations of the random variables  $\widetilde{\Phi}_t$ and $\widetilde{\Psi}_t$ with distributions specified in \eqref{eq.predictive} and \eqref{eq.predictiveImp}, respectively. The second term $\mbox{Pr}( \mathcal{I}_{t+1} | \mathcal{S}_{t+1}, \mathcal{S}_t, \mathcal{I}_t)$ follows the Bayesian updates for the knowledge states as specified in \eqref{eq.update2} and \eqref{eq.update3} given the realization of the physical states $(p_{t+1},i_{t+1})$ or equivalently the growth rate samples $(\phi_t,\psi_t) = (\ln (p_{t+1}/p_t),\ln (i_{t+1}/i_t))$.

At any decision epoch $t$ with physical states $(p_t, i_t)$, if the action is to harvest (i.e., $a_t = H$), the fermentation process ends. We model this situation by assuming that, if the harvest action is taken, the state of the MDP makes a transition to an {\it absorbing  stopping state} $\Delta$. Thus, the counterpart of \eqref{eq.hyper_transition} for the harvest action can be written as $\mbox{Pr}(\Delta|  \mathcal{S}_t, \mathcal{I}_t; H) = 1$. 

\noindent \underline{Reward:}
At any decision epoch $t$, if the decision is to continue, the immediate cost $c_u$ is charged. 
The cost $c_u$ represents the cost of resources allocated to continue the fermentation process one more time step (e.g., a fixed energy cost for the bioreactor, operator cost, clean room charges). Since the time periods are of equal length, it is natural to assume a constant operating cost $c_u$ per time unit. On the other hand, if the decision is to harvest, the fermentation is terminated and a reward is collected. The reward of the harvest decision depends on the current physical state. Specifically, if the harvest decision is taken because of a failure (i.e., if $i_t=\bar{I}$), then the failure penalty $r_f$ is charged as the cost of losing the batch due to the failure. On the other hand, if there is no failure at the harvesting moment (i.e., if $i_t < \bar{I}$), the harvest reward 
\begin{equation}
\label{eq.reward_linear}
r_h(p_t,i_t) =  c_0 + c_1 p_t - c_2 i_t 
\end{equation}
is collected as an immediate reward. In \eqref{eq.reward_linear}, $c_0>0$ represents the lump-sum reward collected per fermentation batch, while $c_1>0$ and $c_2>0$ represent the marginal reward collected per unit of protein and the marginal cost encountered per unit of impurity, respectively. 
We shall assume that $r_f>c_2 \bar{I}$, reflecting the fact that a failure is costlier than even the worst harvesting outcome. To summarize, given the 
physical states $(p_t,i_t)$ and the action $a_t$, the reward $R(p_t, i_t; a_t)$ at decision epoch $t$ can be written as,
\begin{equation} \label{eq.reward}
R(p_t, i_t; a_t) = 
    \begin{cases}
        -c_u, &   a_t = C, \ \  i_t < \bar{I}\\
            r_h(p_t,i_t), & a_t = H, \ \ i_t < \bar{I} \\
-r_f, ~& a_t = H, \ \ i_t =  \bar{I} 
    \end{cases}.
\end{equation}

\noindent \underline{Policy:} 
Let $\pi$  denote a nonstationary policy $\{\pi_t(\cdot);~ t = 0, 1, \ldots, \bar{T}
\}$, which is a mapping from any hyper state $\mathcal{H}_t$ to an action $a_t$, i.e., $a_t = \pi_t(\mathcal{H}_t)$. Given the policy $\pi$, the expected total discounted reward is
\begin{equation}
\label{eq:obj}
\rho(\pi) = \mbox{E}\left[ \sum_{t=0}^{T} \gamma^t R(p_t,i_t;  \pi_t(\mathcal{H}_t)) \bigg| \mathcal{H}_0, \pi  \right],
\end{equation}
where $\gamma \in (0, 1]$ is the discount factor. Notice that \eqref{eq:obj} represents the expected total discounted reward under the policy $\pi$ from time point $0$ until the termination of the fermentation process. The stopping time $T$ in \eqref{eq:obj} is the decision epoch at which the harvest action is taken; i.e., if $\pi_t(\mathcal{H}_t) = H$, then $T=t$. Our objective is to find the optimal policy $\pi^*$ that maximizes the  expected total discounted reward, i.e., $\pi^\star = \arg\max_{\pi} \rho(\pi)$.

\noindent \underline{Value Function:}
The value function $V_t(\mathcal{H}_t)$ is defined as the expected total discounted reward starting  from the decision epoch $t$ with hyper state $\mathcal{H}_t$ under the optimal policy $\pi^*$, i.e.,
\begin{equation*}
V_t(\mathcal{H}_t) =  \mbox{E}
\left[ \sum_{\ell=t}^{T} \gamma^{\ell} R(p_\ell, i_\ell; \pi^*_\ell(\mathcal{H}_{\ell}) ) \bigg| \mathcal{H}_t  \right].
\end{equation*}
The value function $V_t(\mathcal{H}_t)$, or equivalently  $V_t(p_{t}, i_{t}, \mathcal{I}_{t})$, represents the maximum expected total discounted reward starting from  decision epoch $t$ with physical state $(p_t,i_t)$ and knowledge state $\mathcal{I}_t$,  
and it can be recursively written as
\begin{equation}
    \label{eq: valuefunction}
	V_t(p_t, i_t, \mathcal{I}_{t})= \begin{cases} 
	{\max}\left\{ r_h(p_t, i_t), -c_u + \gamma \mbox{E}\left[
	V_{t+1}(p_{t+1}, i_{t+1}, \mathcal{I}_{t+1})\right]\right\} & {\rm if \ } p_t <\bar{P} {\rm \ and \ } i_t < \bar{I} \\
	r_h(p_t, i_t) & {\rm if \ } p_t = \bar{P} {\rm \ and \ } i_t  < \bar{I} \\
	- r_f & {\rm if \ } i_t  = \bar{I} \\
	\end{cases}
\end{equation}
for $t=0,1\ldots,\bar{T}-1$.

At the decision epoch $\bar{T}$, the value function is equal to
\begin{equation}
    \label{eq: terminalvalue}
	V_{\bar{T}}(p_{\bar{T}}, i_{\bar{T}}, \mathcal{I}_{\bar{T}})
	= 
	\begin{cases}
	r_h(p_{\bar{T}},i_{\bar{T}}),
	~~ &\mbox{~if~} i_{\bar{T}} <  \bar{I} \\
	-r_f,
	~~ &\mbox{~if~}  i_{\bar{T}} =  \bar{I}
	\end{cases} 
\end{equation}
because the only feasible action is to harvest if the time point $\bar{T}$ is reached, and either the harvesting reward or the failure cost is charged depending on the impurity amount at decision epoch $\bar{T}$.
Recall that the hyper state transits to the absorbing stopping state $\Delta$ after the harvest action, and the value function is equal to zero for a process already at the stopping state, i.e., $V_t(\mathcal{H}_t) = 0$ if $\mathcal{H}_t=\Delta$ at any $t$. So, the transition to the stopping state $\Delta$ is omitted in \eqref{eq: valuefunction} and \eqref{eq: terminalvalue}. 

\section{Analysis} \label{Sec:Analysis}

Section~\ref{subsec:varianceDecomposition} presents a characterization of the variability in the posterior predictive distribution of the growth rates. Section~\ref{subsec:structuralAnalysisPhysicalState} provides some analytical properties of the optimal policy. Note that our MDP model is a variant of the classical optimal stopping problem \citep{ferguson2000optimal}. A well-known class of policies for optimal-stopping problems is look-ahead policies. Motivated by its simplicity for applying in practice, 
we consider the {\it one-step look-ahead policy} (referred to as myopic policy) in Section~\ref{subsec: PerfInfo}. 
Finally, Section~\ref{sec:optimizationAlgorithm-main} discusses our solution approach to obtain the optimal policy. All the proofs and the algorithm procedure are provided in the Appendix.

\subsection{Growth-Rate Variability under Model Risk}
\label{subsec:varianceDecomposition}

Recall that the uncertainty in the protein and impurity growth rates comes from two sources: the inherent stochasticity of the fermentation and the model risk. Conditional on the historical data collected until the decision epoch $t$, the predictive protein growth rate $\widetilde{\Phi}_{t}$ and the predictive impurity growth rate  $\widetilde{\Psi}_{t}$ are the random variables that the decision maker uses to model the growth rates, and these random variables account for both sources of uncertainty (see Section~\ref{subsec:onlineLearning}). The objective of this section is to quantify the contribution of each source of uncertainty to the 
predictive variance of the random variables $\widetilde{\Phi}_{t}$ and $\widetilde{\Psi}_{t}$, denoted with $\widetilde{\sigma}_{t}^{(p)2}$ and $\widetilde{\sigma}_{t}^{(i)2}$, respectively.

Let $\mathcal{D}_t = \{ (\phi^{(0)},\psi^{(0)}),(\phi^{(1)},\psi^{(1)}),\ldots, (\phi^{({J_t})},\psi^{(J_t)})\}$ denote the historical data on past realizations of the growth rates available at the $t$-th decision epoch of the fermentation process. It is possible that the data size $J_t$ can be greater than $t$ as the historical data $\mathcal{D}_t$ may also include the growth-rate realizations from the previous fermentation processes. Recall from Section~\ref{subsec:onlineLearning} that the knowledge states can be recursively written as a function of the historical data $\mathcal{D}_t$; see equations \eqref{eq.update2} and \eqref{eq.update3}. By applying the commonly used improper prior that assumes the initial belief states $\alpha_{0}^{(p)},\nu_{0}^{(p)},\lambda_{0}^{(p)}, \beta_{0}^{(p)},\alpha_{0}^{(i)}, \nu_{0}^{(i)}, \lambda_{0}^{(i)}, \beta_{0}^{(i)}$ are all equal to 0, the knowledge states  can be obtained as
\begin{equation}
		\alpha_{t}^{(p)} = \bar{\phi},~ \nu_{t}^{(p)} = J_t,~ \lambda_{t}^{(p)} = \dfrac{J_t}{2},~
		\beta_{t}^{(p)} = \dfrac{1}{2}\sum_{j=1}^{J_t}(\phi^{(j)} - \bar{\phi})^2, 
		\nonumber
	\end{equation}
	\begin{equation}
		\alpha_{t}^{(i)} = \bar{\psi},~ \nu_{t}^{(i)} = J_t,~ \lambda_{t}^{(i)} = \dfrac{J_t}{2},~
		\beta_{t}^{(i)} = \dfrac{1}{2}\sum_{j=1}^{J_t}(\psi^{(j)} - \bar{\psi})^2,
		\nonumber
\end{equation}
where $\bar{\phi} = \sum_{j=1}^{J_t}\phi^{(j)}/J_t$ and $\bar{\psi} = \sum_{j=1}^{J_t}\psi^{(j)}/J_t$. The posterior predictive variances are then given by
\begin{equation}
\widetilde{\sigma}_{t}^{(p)2} = \dfrac{J_t+1}{(J_t-2)J_t}\sum_{j=1}^{J_t}(\phi^{(j)} - \bar{\phi})^2,~~
\widetilde{\sigma}_{t}^{(i)2} = \dfrac{J_t+1}{(J_t-2)J_t}\sum_{j=1}^{J_t}(\psi^{(j)} - \bar{\psi})^2 \label{eq.sigma_tilde}
\end{equation} 
for $J_t>2$. Next, by considering the randomness in the historical data $\mathcal{D}_t$, Proposition~\ref{theo:var_decomp}(i) establishes the expectation and variance of $\widetilde{\sigma}_{t}^{(p)2}$ (i.e., similar to the characterization of the expectation and variance for the sample variance of a set of realizations from a specific population). For a particular realization of the historical data set $\mathcal{D}_t$, Proposition~\ref{theo:var_decomp}(ii) characterizes the predictive variance $\widetilde{\sigma}_{t}^{(p)2}$ as the sum of two closed-form terms that represent the variability in the growth rate due to the inherent stochasticity of the fermentation process and the model risk, respectively. 

\begin{proposition}
\label{theo:var_decomp}

(i) $\mbox{E}\left[\widetilde{\sigma}_{t}^{(p)2}\right] = \sigma^{(p)2}_c + \dfrac{(2J_t-1)\sigma^{(p)2}_c}{(J_t^2-2J_t)}$ and $\mbox{Var}\left[\widetilde{\sigma}_{t}^{(p)2}\right] = \dfrac{2(J_t^3+J_t^2-J_t-1)\sigma^{(p)4}_c}{J_t^4 - 4J_t^3 + 4J_t^2}$. 

(ii) Conditional on the historical data $\mathcal{D}_t$, the predictive variance $\widetilde{\sigma}_{t}^{(p)2}$ for the protein growth rate can be decomposed into two components $\hat{\sigma}_{t}^{(p)2} = \dfrac{\beta_{t}^{(p)}}{\lambda_{t}^{(p)} - 1}$ and $\check{\sigma}_{t}^{(p)2} = \dfrac{\beta_{t}^{(p)}}{(\lambda_{t}^{(p)} - 1)\nu_{t}^{(p)}}$, representing the variability of the protein growth rate due to inherent stochasticity and the model risk, respectively; i.e., $\widetilde{\sigma}_{t}^{(p)2} = \hat{\sigma}_{t}^{(p)2} + \check{\sigma}_{t}^{(p)2}$
with 
$\hat{\sigma}_{t}^{(p)2} = \dfrac{\sum_{j=1}^{J_t}(\phi^{(j)} - \bar{\phi})^2}{J_t - 2}
~~~ \mbox{and} ~~~
\check{\sigma}_{t}^{(p)2} = \dfrac{\sum_{j=1}^{J_t}(\phi^{(j)} - \bar{\phi})^2}{J_t^2 - 2 J_t}$.

\end{proposition}

We notice from Proposition~\ref{theo:var_decomp}(i) that the bias $\mbox{E}\left[\widetilde{\sigma}_{t}^{(p)2} - \sigma^{(p)2}_c \right] =  \dfrac{(2J_t-1)\sigma^{(p)2}_c}{J_t^2-2J_t} > 0$ for $J_t > 2$.
Therefore, under model risk, on average the predictive variance $\widetilde{\sigma}_{t}^{(p)2}$ will be greater than the true variance $\sigma^{(p)2}_c$. Furthermore, as the amount of historical data $J_t$ increases, $\mbox{Var}\left[\widetilde{\sigma}_{t}^{(p)2}\right]$ converges to zero, and the predictive variance $\widetilde{\sigma}_{t}^{(p)2}$ will converge to $\sigma^{(p)2}_c$, which represents the protein growth-rate variance under perfect information. 
Given a particular realization of the historical data set $\mathcal{D}_t$, Proposition~\ref{theo:var_decomp}(ii) is useful in practice as it allows making a judgment on how the overall uncertainty in the protein growth rate is affected from the inherent stochasticity of the fermentation process and from the model risk.
Notice that the ratio of model risk to the inherent stochasticity, i.e., $\check{\sigma}_{t}^{(p)2}/\hat{\sigma}_{t}^{(p)2} = 1/\nu_{t}^{(p)}$,
only depends on the shape parameter $\nu_{t}^{(p)}$
which is
equal to
$J_t$.
This intuitively shows that the model risk becomes smaller (relative to the inherent stochasticity of the process) as the size of the historical data increases.

Notice that Proposition~\ref{theo:var_decomp} also applies to the impurity growth model. To be specific, the same results hold for the predictive variance $\widetilde{\sigma}_{t}^{(i)2}$ of the impurity growth rate, as the functional form and the underlying modeling assumptions are the same as in the protein growth model. For brevity, we do not repeat those results in the paper.

\subsection{Analytical Properties of the Optimal Policy}
\label{subsec:structuralAnalysisPhysicalState}

We start our analysis by first showing the monotonicity of the value function, and then present sufficient conditions for the existence of a \textit{control-limit policy} with respect to the impurity level. 

\begin{proposition}\label{theo:physical}
    Given the knowledge state $\mathcal{I}_t$, the value function $V_t(p_t,i_t, \mathcal{I}_t)$ is a non-increasing function of the impurity level $i_t$ and a non-decreasing function of the protein level $p_t$.
\end{proposition}

Based on the monotonicity properties presented in Theorem~\ref{theo:physical}, we can derive  sufficient conditions for the existence of a control-limit policy with respect to the impurity level as follows.

\begin{proposition}\label{theo:policy_i} 
    At any decision epoch $t$ with a given protein level 
    and knowledge state, there exists a critical threshold $i_t^\star$ such that the optimal decision is to harvest for the impurity level $i_t\ge i_t^\star$ if the following condition holds for all $i_t^+ > i_t^- \ge 0$:
	\begin{equation}
	c_2 (i_t^+ - i_t^-) \le \gamma r_f \left[ \mbox{Pr}\left(i_{t+1} \le \bar{I} | i_t^- \right) - \mbox{Pr}\left(i_{t+1} \le \bar{I} | i_t^+ \right) \right]  - \gamma c_1\bar{P} \mbox{Pr}\left(i_{t+1} \le \bar{I} | i_t^- \right)  
	\label{eq.optimalPolicyCondition}
	\end{equation}
	where $\mbox{Pr}\left(i_{t+1} < \bar{I} | i_t \right) = \mbox{Pr}\left(i_t e^{\widetilde{\Psi}_t} < \bar{I}  \right) = \int_{-\infty}^{\ln{\bar{I}} - \ln{i_t}} f_{t}^{(i)}(\psi_t)d\psi_t $ is the probability that the process failure doesn't occur in the time period that starts with the impurity level $i_t$ at decision epoch $t$.
\end{proposition}

Proposition~\ref{theo:policy_i} presents the existence of a critical threshold $i_t^*$ with respect to the impurity level: for a given protein and knowledge state, if we harvest at a certain impurity level, we will also harvest at any higher level of impurity. That is, at a given knowledge state, the physical-state space can be split into a harvest zone and a continue zone indicating the optimal action. The notations $i_t^+$ and $i_t^-$ in Theorem~\ref{theo:policy_i} represent any two distinct values of the impurity state that must satisfy the sufficient condition~\eqref{eq.optimalPolicyCondition} to assure the optimality of a control-limit policy with respect to the impurity level. If the condition~\eqref{eq.optimalPolicyCondition} is satisfied for all $i_t^+ > i_t^- \ge 0$, the optimal policy is guaranteed to be a control-limit policy with a critical threshold on the impurity level.

Note that the condition~\eqref{eq.optimalPolicyCondition} is more likely to be satisfied as the relative value of the failure penalty $r_f$ increases compared to  $c_1$ and $c_2$. In practice, it is common that $r_f$ is much larger compared to $c_1$ and $c_2$, reflecting the fact that failures are undesired because of strict safety concerns, loss of reputation, and extra rework. In fact, the condition~\eqref{eq.optimalPolicyCondition} holds for the realistic instances we studied in our case study, where the optimal policy indeed follows a threshold-type policy with respect to the impurity level (see Figure~\ref{fig:emp_boundary}).
However, we observe that the optimal policy does not follow a threshold-type structure with respect to the protein level in realistic instances (see Section~\ref{subsec:empirical1} for an additional discussion on optimal policies based on an industry case study).

\subsection{Myopic Policy}
\label{subsec: PerfInfo}
In this section, we consider a one-step look-ahead policy as a practically appealing alternative policy. We refer to it as the myopic policy because it makes the harvesting decisions by only comparing the reward of harvesting at the current decision epoch with the expected reward of harvesting at the next decision epoch.  We study the myopic policy because it can be implemented by only maintaining a posterior distribution of the unknown growth-rate distribution parameters. It can be relevant to many small-sized biomanufacturing companies that may not have the necessary infrastructure or expertise to compute the optimal policy.

\subsubsection{Myopic Policy under Perfect Information.}
%
We first consider the case where the true parameters of the underlying stochastic model are known, referred to as the \textit{perfect-information} case. That is, the parameters $\mu^{(p)}_c$ and $\sigma^{(p)2}_c$ for the protein growth rate and the parameters $\mu^{(i)}_c$ and $\sigma^{(i)2}_c$ for the impurity growth rate are known by the decision maker. The analysis in this section assumes that the probability of a negative growth rate is negligible (i.e., the realizations of the growth rate random variables $\Phi_t$ and $\Psi_t$ are always non-negative), which is often the case in practice with a standard deviation of the normally distributed growth rates expected to be much smaller than their mean values. 

Let $A$ denote the physical state space at which harvesting is at least as good as continuing for exactly one more time period and then harvesting:
\begin{equation} \label{eq:myopic_policy}
A = \left\{ (p, i): r_h(p, i) \ge -c_u +  \gamma  \mbox{E}[R(p^\prime, i^\prime; H) | p, i] \right\}.
\end{equation}
The expectation in \eqref{eq:myopic_policy} is with respect to the underlying true growth model \eqref{eq.growthmodel}, and $(p^\prime, i^\prime)$ denotes the protein and impurity levels of the next decision epoch after the continue action is taken in the current decision epoch at state $(p, i)$. 

\begin{definition}[Myopic Policy under Perfect Information]
\label{def: MPMR}
The policy that takes the harvest decision the first time the protein and impurity levels enter a state in $A$ is defined as the myopic policy under perfect information.  
\end{definition}

Our objective is to establish when the myopic policy is optimal in the perfect-information setting. Proposition~\ref{prop: myopic_under_PI} establishes the optimality of the myopic policy under some specific conditions.

\begin{proposition}
\label{prop: myopic_under_PI} 
Consider a fermentation process starting with protein level $p_0$ and impurity level $i_0$. The myopic policy is the optimal policy under perfect information, if $p_0 \in [\underline{p}, \bar{P}]$ with
\begin{equation}
\underline{p} = \exp\left\{ \ln{\bar{P}} - \mu_c^{(p)} - \sigma_c^{(p)2} - \sigma_c^{(p)}\Phi^{-1} \left[ \dfrac{e^{- \mu_c^{(p)} - \sigma_c^{(p)2}/2}}{\gamma \Phi\left( \dfrac{\ln{\bar{I}} - \ln{i_0} - \mu_c^{(i)}}{\sigma_c^{(i)}} \right) } \right]  \right\} \label{eq:sec4lem1},
\end{equation}
and 
\begin{align}
\dfrac{c_2}{\gamma}(i^+ - i) \le& r_f \left[ \Phi\left( \dfrac{\ln{\bar{I}} - \ln{i} - \mu_c^{(i)}}{\sigma_c^{(i)}} \right) - \Phi\left( \dfrac{\ln{\bar{I}} - \ln{i^+} - \mu_c^{(i)}}{\sigma_c^{(i)}} \right) \right] \nonumber\\
&- c_2 \bar{I} e^{\mu_c^{(i)} + \sigma_c^{(i)2}/2} \left[ \Phi\left( \dfrac{\ln{\bar{I}} - \ln{i} - \mu_c^{(i)} - \sigma_c^{(i)2} }{\sigma_c^{(i)}} \right) - \Phi\left( \dfrac{\ln{\bar{I}} - \ln{i^+} - \mu_c^{(i)} - \sigma_c^{(i)2} }{\sigma_c^{(i)}} \right) \right] \label{eq:sec4lem2}
\end{align}
for all $i, i^+ \in [i_0, \bar{I}]$ with $i<i^+$.
\end{proposition}
The conditions in Proposition~\ref{prop: myopic_under_PI} assure that the physical state space $A$ is a closed set, meaning that when the process enters into set $A$ it never leaves the set $A$. Therefore, the myopic policy turns out to be optimal. The conditions in Proposition~\ref{prop: myopic_under_PI} hold when the starting protein level and the failure cost are sufficiently large. Intuitively, this result can be interpreted as follows: It is wise to harvest now and not to take a costly failure risk in  future periods if a sufficient amount of protein is already accumulated. 

While the equation \eqref{eq:sec4lem1} is the closed-form characterization of a starting protein-level lower bound for the optimality of the myopic policy, the condition   \eqref{eq:sec4lem2} is not immediately intuitive. To make it easier to interpret, we  simplify \eqref{eq:sec4lem2} by applying a Taylor series approximation. To be specific, by applying first-order Taylor series approximation of the nonlinear terms on both sides of inequality \eqref{eq:sec4lem2}, it can be approximated as 
\begin{align}
r_f \ge & c_2\bar{I} \left[ \dfrac{\sqrt{2\pi} \sigma_c^{(i)}}{\gamma} + e^{\mu_c^{(i)} + \sigma_c^{(i)2}/2} \right]. \label{eq:sec4lem2approx}
\end{align}
Note that the inequality \eqref{eq:sec4lem2approx} is more likely to hold as the failure cost $r_f$ increases or the maximum purification cost $c_2 \bar{I}$ decreases. We provide the details of this approximation in Appendix~B.

\subsubsection{Myopic Policy under Model Risk.}
\label{subsec: myopicinformation policy}

In this section, the true parameters of the underlying stochastic model for the growth rates are not known anymore. That is, there is a model risk. Our objective is to investigate how the model risk affects the harvesting decisions of the myopic policy.
When there is model risk, the expected reward of harvesting at the next decision epoch is calculated by using the posterior predictive distributions of the growth rates (i.e., the distributions characterized in \eqref{eq.predictive} and \eqref{eq.predictiveImp}). 
Let 
\begin{equation}
    \widetilde{h}(\alpha^{(p)}, \widetilde{\sigma}^{(p)}, \alpha^{(i)}, \widetilde{\sigma}^{(i)}; p, i) = r_h(p, i) + c_u -  \gamma  \mbox{E}[R(p^\prime, i^\prime; H) | p, i, \alpha^{(p)}, \widetilde{\sigma}^{(p)}, \alpha^{(i)}, \widetilde{\sigma}^{(i)}],
\end{equation}
where $\alpha^{(p)}$ and $\alpha^{(i)}$ are the means and $\widetilde{\sigma}^{(p)}$ and $\widetilde{\sigma}^{(i)}$ are the standard deviations of the  posterior predictive distribution for the protein and impurity growth rates, respectively.
We first define the myopic policy under model risk.
\begin{definition}[Myopic Policy under Model Risk]
\label{def: MPMR} 
The policy that takes the harvest decision the first time the inequality $\widetilde{h}(\alpha^{(p)}, \widetilde{\sigma}^{(p)}, \alpha^{(i)}, \widetilde{\sigma}^{(i)}; p, i)\geq 0$ holds is defined as the myopic policy under model risk.  
\end{definition}

\begin{figure}[!h]{
		\centering
		\includegraphics[scale=0.525]{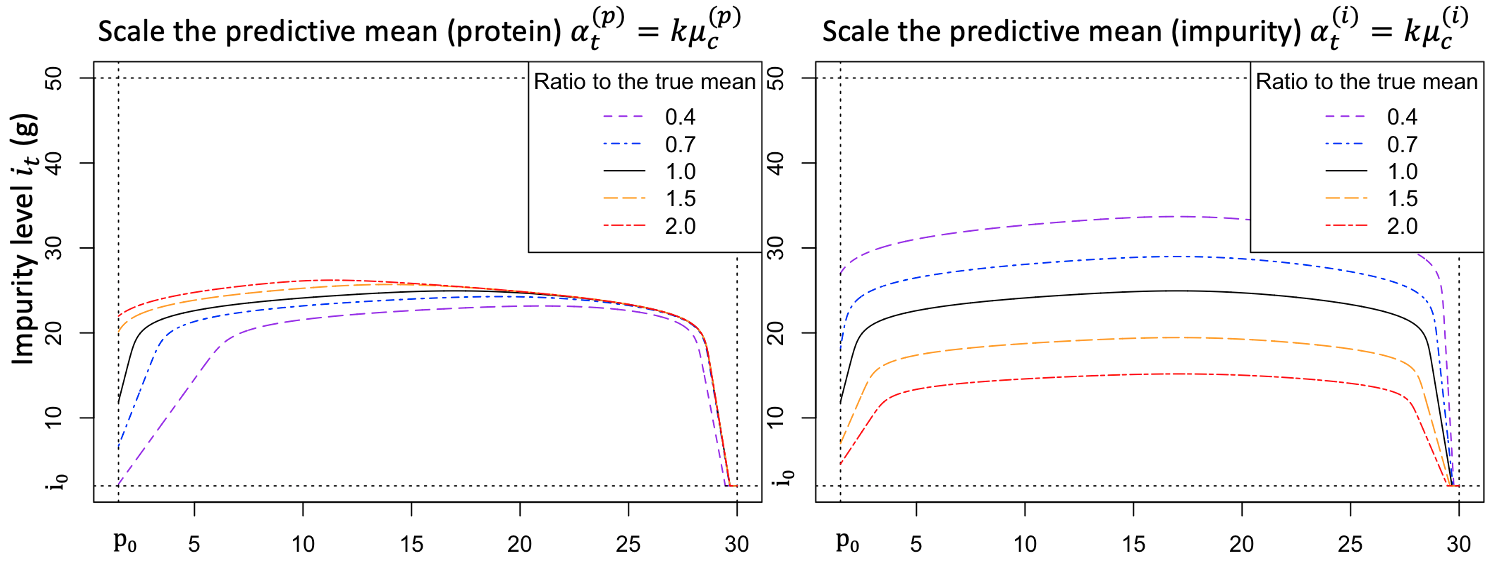}
		\includegraphics[scale=0.525]{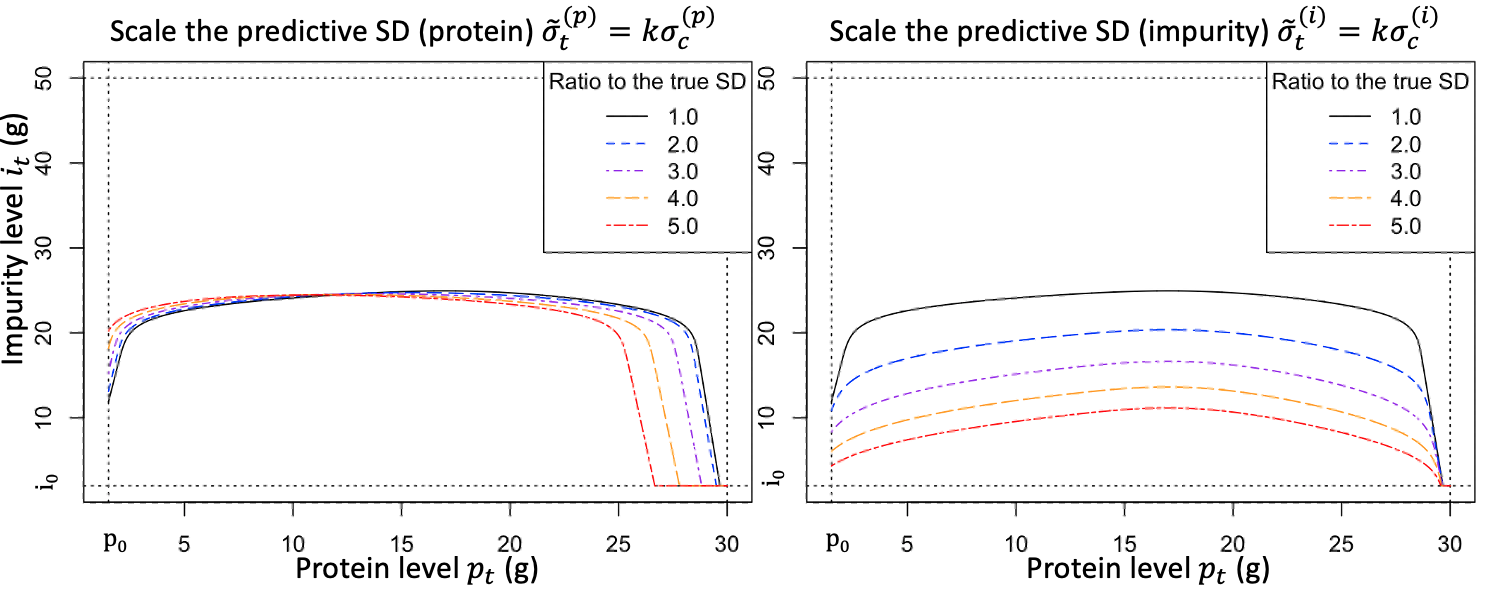}
		\caption{Illustration of how the harvest boundary is affected by the model risk (based on the case study parameters presented in Section~\ref{subsec:overview}).} 
		\label{fig:boundary}
		}
\end{figure}

We will study the effect of model risk on the harvesting decisions of the myopic policy by investigating how the so-called {\it harvest boundary}, which is given by 
\[
\{(p, i):  \widetilde{h}(\alpha^{(p)}, \widetilde{\sigma}^{(p)}, \alpha^{(i)}, \widetilde{\sigma}^{(i)}; p, i) = 0 \},
\]
is influenced by the parameters of the posterior predictive distributions. Recall that each posterior predictive distribution was t-distributed. To be able to generate some further insights in the remainder of this section, we use the normal approximations of these posterior predictive distributions (we later numerically investigate the effect of normal approximation for smaller than 30 data points, and find that the impact of this approximation on the performance of the myopic policy is negligible). When  the posterior predictive distributions are normal, since the true growth rates also follow normal distribution, we can focus on the difference between the posterior predictive distribution parameters and their true counterparts to generate insights on the affect of this difference on the shape of the harvest boundary. We present Figure~\ref{fig:boundary} as an illustration. 

Figure~\ref{fig:boundary} plots the harvest boundary for various levels of model risk. To be specific, Figure~\ref{fig:boundary} introduces a scaling factor $k$ that links each parameter to its true counterpart, and plots the harvest boundaries for some relevant values of $k$. For example, $k=1$ represents the case where each predictive-distribution parameter reduces to its true counterpart, e.g., ${\alpha}_{t}^{(p)}$, the mean of the predictive distribution of the protein growth rate, becomes equal to $\mu^{(p)}_c$, the true mean of the protein growth rate. Therefore, the resulting harvest boundary becomes equivalent to the harvest boundary of the myopic policy under perfect information. Recall from Proposition~\ref{prop: myopic_under_PI} that the myopic policy is optimal under perfect information when certain conditions hold. For the illustrative example in Figure~\ref{fig:boundary}, we  know $\underline{p}$ from \eqref{eq:sec4lem1} is equal to $17.34$, and condition~\eqref{eq:sec4lem2} is satisfied. Thus, for protein levels greater than $17.34$, Figure~\ref{fig:boundary} already gives an insight about how close the harvest boundary of the myopic policy under model risk is to the harvest boundary of the optimal policy under perfect information.

Figure~\ref{fig:boundary}~(top,~left) shows that as the predictive mean of protein ${\alpha}_{t}^{(p)}$ increases, the harvest boundary moves up, indicating that it becomes less beneficial to harvest immediately when the potential future gain in protein amount is high. Also, the difference on the left part (i.e., small amount of protein $p_t$) is larger than on the right (i.e., large amount of protein $p_t$), since as protein approaches the harvesting limit, an increase in ${\alpha}_{t}^{(p)}$ becomes less influential on the harvesting decision. 
Figure~\ref{fig:boundary}~(top,~right) shows that as the predictive mean of impurity ${\alpha}_{t}^{(i)}$ increases, the harvest boundary moves down. This is intuitive because a higher ${\alpha}_{t}^{(i)}$ implies a higher probability of a fermentation failure, so a smaller amount of current impurity level is sufficient to trigger a harvest action. 
According to Figure~\ref{fig:boundary}~(bottom,~left), as the parameter $\widetilde{\sigma}_{t}^{(p)}$ increases, the decision boundary shifts to the left. That is, with higher uncertainty of protein growth, we tend to harvest with a smaller amount of protein especially when it is close to the maximum protein level. On the other hand, as the parameter $\widetilde{\sigma}_{t}^{(i)}$ increases, Figure~\ref{fig:boundary}~(bottom,~right) shows that the decision boundary moves down. A larger $\widetilde{\sigma}_{t}^{(i)}$ implies higher uncertainty in the impurity growth rate, and we tend to harvest with a smaller amount of impurity to avoid a batch failure. 

Proposition~\ref{theo:boudary_move_mean} formalizes our observations from Figure~\ref{fig:boundary} by establishing the monotonicity of the function $\widetilde{h}(\alpha^{(p)}, \widetilde{\sigma}^{(p)}, \alpha^{(i)}, \widetilde{\sigma}^{(i)}; p, i)$ that characterizes the harvest boundary with respect to the parameters of the predictive growth-rate distributions.

\begin{proposition}\label{theo:boudary_move_mean}
The function $\widetilde{h}(\alpha^{(p)}, \widetilde{\sigma}^{(p)}, \alpha^{(i)}, \widetilde{\sigma}^{(i)}; p, i)$ is 
\begin{itemize}
    \item[(i)] increasing in $\alpha^{(i)}$;
    \item[(ii)] decreasing in $\alpha^{(p)}$;
    \item[(iii)] increasing in $\widetilde{\sigma}^{(i)}$, if $\ln{\bar{I}} - \ln{i} > \alpha^{(i)}$ and
    \begin{equation}
    \widetilde{\sigma}^{(i)} \Phi\left( \dfrac{\ln{\bar{I}} - \ln{i} - \alpha^{(i)} - \widetilde{\sigma}^{(i)2} }{\widetilde{\sigma}^{(i)}} \right) 
    - \phi\left( \dfrac{\ln{\bar{I}} - \ln{i} - \alpha^{(i)} - \widetilde{\sigma}^{(i)2} }{\widetilde{\sigma}^{(i)}} \right) > 0; \label{eq:thm6-eq1}
    \end{equation}
    \item[(iv)] decreasing in $\widetilde{\sigma}^{(p)}$, if and only if
    \begin{equation}
    \widetilde{\sigma}^{(p)} \Phi\left( \dfrac{\ln{\bar{P}} - \ln{p} - \alpha^{(p)} - \widetilde{\sigma}^{(p)2} }{\widetilde{\sigma}^{(p)}} \right) 
    - \phi\left( \dfrac{\ln{\bar{P}} - \ln{p} - \alpha^{(p)} - \widetilde{\sigma}^{(p)2} }{\widetilde{\sigma}^{(p)}} \right) > 0. \label{eq:thm6-eq2}
    \end{equation}
\end{itemize}
\end{proposition}

Proposition~\ref{theo:boudary_move_mean} can be useful in understanding how the model risk would influence the harvest boundary of the myopic policy. For example, if the predictive mean $\alpha^{(i)}$ of the impurity growth rate increases, since $\widetilde{h}(\alpha^{(p)}, \widetilde{\sigma}^{(p)}, \alpha^{(i)}, \widetilde{\sigma}^{(i)}; p, i)$ is increasing by Proposition~\ref{theo:boudary_move_mean}(i), the harvest zone of the physical-state space, denoted by $\{(p, i): \widetilde{h}(\alpha^{(p)}, \widetilde{\sigma}^{(p)}, \alpha^{(i)}, \widetilde{\sigma}^{(i)}; p, i) \ge 0 \}$, will enlarge, while the continue zone $\{(p, i): \widetilde{h}(\alpha^{(p)}, \widetilde{\sigma}^{(p)}, \alpha^{(i)}, \widetilde{\sigma}^{(i)}; p, i) < 0 \}$ will shrink.
In other words, the decision maker tends to be more willing to take the harvest decisions for larger $\alpha^{(i)}$. 
Similar insights can be derived for all other predictive parameters. 
Notice that the harvest boundary can behave differently  as a function of the predictive standard deviations (i.e., $\widetilde{\sigma}^{(i)}$ and $\widetilde{\sigma}^{(p)}$) at different parts of the physical-state space, since the monotonicity properties with respect to $\widetilde{\sigma}^{(i)}$ and $\widetilde{\sigma}^{(p)}$ depend on whether the conditions \eqref{eq:thm6-eq1} and \eqref{eq:thm6-eq2} hold for particular values of $(p,i)$ in the physical-state space.

\begin{proposition}\label{theo:boundary_consistency}
As the size of the historical data $\mathcal{D}_t$ approaches infinity (i.e., as $J_t \to \infty$), the myopic policy under model risk becomes equivalent to the myopic policy under perfect information. 
\end{proposition}

Proposition~\ref{theo:boundary_consistency} is useful in practice as it tells us that the myopic policy under model risk becomes similar to the myopic policy under perfect information, which we already know to be optimal under the conditions in Proposition~\ref{prop: myopic_under_PI}. 
Intuitively, this represents the situation with a sufficient amount of historical data such that the underlying fermentation process is already learned accurately.

\subsection{Reinforcement Learning under Model Risk (RL with MR)}
\label{sec:optimizationAlgorithm-main}

In this section, we introduce our solution approach
to compute the optimal policy that minimizes the objective function in~\eqref{eq:obj}. Different from the myopic policy in Section~\ref{subsec: PerfInfo}, 
we now consider the effect of learning from future data on harvesting decisions  (in a forward-looking manner). Recall that the harvest action at physical states $(p_t,i_t)$ ends the fermentation with a deterministic (known) reward. Therefore, it is only needed to estimate the total reward associated with the continue action and following the optimal policy thereafter, denoted by the corresponding Q-function
\begin{equation}
\label{eqn: Qcont}
	Q_t(p_{t}, i_{t}, \mathcal{I}_{t}; C)
	\triangleq
	-c_u + \gamma \mbox{E}\left[
	\max_{a_{t+1}\in \mathcal{A}} Q_{t+1}(p_{t+1}, i_{t+1}, \mathcal{I}_{t+1}; a_{t+1})\right] 
\end{equation}
for $t=0,1,\ldots,\bar{T}-1$. On the other hand, the 
Q-function associated with the harvest action is denoted with  
\begin{equation}
\label{eqn: Qharv}
	Q_t(p_{t}, i_{t}, \mathcal{I}_{t};  H)
	\triangleq 
	\begin{cases}
	r_h(p_t, i_t),~~&\mbox{~if~} i_t < \bar{I} \\
	-r_f,~~&\mbox{~if~} i_t = \bar{I}
	\end{cases} 
\end{equation}
for $t=0,1,\ldots,\bar{T}$. Recall that harvesting is the only feasible action for $i_t=\bar{I}$, $p_t =\bar{P}$, or $t=\bar{T}$, and the harvest action takes the state of the system to a cost-free absorbing state.  For a fermentation process that has not yet been harvested at decision epoch $t$, the value function is given by $V_t(p_t,i_t, \mathcal{I}_{t}) = \max_{a_t} Q_t(p_t,i_t, \mathcal{I}_{t}; a_t)$. Theoretically, the optimal policy  that maps {\it any} possible hyper state to an action can be obtained through backward dynamic programming (this is also referred to as offline planning).

However, computing this dynamic program is notoriously difficult and also not necessary in practice, given that the optimal policy is only needed starting from a specific physical state (which evolves by visiting certain states more likely than others) in the real-life execution of the fermentation process. Therefore, we adopt a solution approach that executes the policy in an online manner, which means that we focus on estimating the Q-function in \eqref{eqn: Qcont} at a particular current state $(p_t,i_t,\mathcal{I}_t)$, and decide to continue or harvest the fermentation process by comparing it with the harvesting reward in  \eqref{eqn: Qharv}. After the selected action is executed, the next decision epoch starts with a new hyper-state at which the entire procedure is repeated. 
The details of the 
solution procedure is provided 
in Appendix~\ref{sec:optimizationAlgorithm}.

\section{Numerical Analysis}
\label{Sec:CaseStudy}

We present a case study motivated by the implementation at MSD Animal Health. 
To protect confidentiality, we disguised MSD's original data and used representative values to generate insights. 


\subsection{
Experiment Setting and Analysis Overview} \label{subsec:overview}

The starting protein and impurity for each batch are $p_0 = 1.5, i_0 = 2.0$ grams. 
The harvesting limit on protein is $\bar{P} = 30$ grams, the batch failure impurity threshold is $\bar{I} = 50$ grams, and the maximum time for a batch to reach the stationary phase is $\bar{T} = 8$ hours. To represent general practice, we considered the following cost and reward structures: harvest reward $r_h(p,i) = 10p - i$, the one-step operation cost $c_u = 2$, and the batch failure penalty $r_f = 880$ with no discounting. These cost structures are identified based on input received from our industry partners (see Appendix~\ref{sec:campaign} for a sensitivity analysis of costs). Based on historical production data, we consider the following protein and impurity growth parameters $\mu^{(p)}_c = 0.488, \sigma^{(p)}_c = 0.144$, $\mu^{(i)}_c = 0.488, \sigma^{(i)}_c =0.144$ as underlying truth. The realized values of growth rates ranged between $0.2$ and $0.6$ in our production data. The protein and impurity generation model in (\ref{eq.growthmodel}) was validated with real-world fermentation data.\endnote{The data had a sample size of 24. We conducted the Kolmogorov–Smirnov normality test and obtained p-value 0.6416, which supports the normality assumption on the growth rate. We also conducted the ANOVA test on the constant mean assumption and the Levene's test on the constant variance assumption over time with p-values 0.251 and 0.5035, respectively.} Consistent with practice, the length of a time period in our discrete-time model is equal to six hours (i.e., 8 decision epochs during a maximum of 48 hours of
fermentation). Appendix~\ref{sec:SensitivityN} presents an additional analysis related to the frequency of decision epochs. 


\noindent \textbf{Analysis overview.} We use the case study to generate insights for practitioners. For this purpose, we consider various practically-relevant strategies as a benchmark and compare their performance:

\begin{itemize}

    \item[(1)] \textbf{Common practice} (CP). This strategy represents a common rule of thumb used in the industry. It uses a `fixed threshold' approach, harvesting when protein and impurity levels exceed certain predetermined threshold values. In our case study, CP harvests when the impurity amount exceeds $60\%$ of the maximum amount permitted $\bar{I}$  or when the protein amount reaches the limit $\bar{P}$ or when the fermentation process transitions to the stationary phase.
    
     \item[(2)] \textbf{Reinforcement learning with model risk (RL with MR).} The decision-maker considers both the inherent stochasticity of fermentation processes and the model risk caused by limited historical data. RL with MR represents the optimal policy under our proposed Bayesian decision-making framework based on both model risk and inherent uncertainty (Section~\ref{sec:optimizationAlgorithm-main}).

      \item[(3)] \textbf{Perfect information MDP (PI-MDP).} To establish a benchmark, we consider the setting where the decision-maker has perfect information on the underlying true model of fermentation dynamics (i.e., no model risk, but only inherent stochasticity). Theoretically, this setting represents the best possible performance that can be achieved with an infinite amount of historical data. We obtain the PI-MDP policy by using the same approach of RL with ML (Appendix~\ref{sec:optimizationAlgorithm}) with one key difference: The true model parameters $\mu^{(p)}_c$, $\mu^{(i)}_c$, $\sigma^{(p)}_c$ and $\sigma^{(i)}_c$ are assumed known, and hence the state space only includes the physical states and there is no sampling from the information states (i.e., sampling of the growth rates is done from the true fermentation model).

    \item[(4)] \textbf{Reinforcement learning ignoring model risk (RL ignoring MR).} 
    This strategy represents the case where the decision maker ignores the model risk by simply using the point estimates of the unknown true parameters $\mu^{(p)}_c$, $\mu^{(i)}_c$, $\sigma^{(p)}_c$ and $\sigma^{(i)}_c$ obtained from limited historical data. That is, the point estimates are used as if they were the true parameters. We obtain the `RL ignoring MR' policy similar to how we solve the PI-MDP, but by replacing the unknown true parameters with their maximum likelihood estimates. We consider this policy because it represents a common approach to using historical data for model calibration, but ignoring the effects of estimation errors and statistical learning on decisions.
    
    
    \item[(5)] \textbf{Myopic policy.} The decision-maker considers both the inherent stochasticity and the model risk (as described in Section~\ref{subsec: myopicinformation policy}). However, the harvesting decision is made by only comparing the harvesting reward at the current decision epoch with the expected reward of harvesting in the next decision epoch. We propose the myopic policy as a relevant benchmark because it is easy to implement, especially for companies that do not have the infrastructure for Bayesian learning implementations.

\end{itemize}


\subsection{Performance Comparison} \label{sec:numericresults}
  We now 
  evaluate the performance of benchmark strategies described in Section~\ref{subsec:overview}. 
  In particular, we focus on the expected total reward $\widehat{\rho}^c(\pi)$, and the standard deviation $\widehat{\mbox{SD}}^c(\pi)$ under different sizes of historical data (i.e., $J_0 = 3, 10, 20$) and harvest policy $\pi$, calculated with 100 simulation replications as described in Appendix~\ref{sec:optimizationAlgorithm}. Table~\ref{table:compare_0} reports the performance of the considered strategies with the starting state $p_0=1.5$ and $i_0=2$. Recall that harvesting policies $\pi$ under both PI-MDP and CP are independent of data size $J_0$. The column labeled ``\% of PI-MDP" in Table~\ref{table:compare_0} uses the perfect information setting (PI-MDP) as a benchmark to assess the performance of the considered strategies. 

\begin{table}[hbt!]
\caption{The mean and standard deviation of the total reward achieved by different strategies.}
\label{table:compare_0}
\centering
\begin{tabular}{|lcccc|}
\hline
\multicolumn{1}{|l|}{}                 & $\widehat{\rho}^c(\pi)$   & \multicolumn{1}{c|}{\% of PI-MDP}    & $\widehat{\mbox{SD}}^c(\pi)$     & \% of PI-MDP  \\ \hline
\multicolumn{1}{|l|}{PI-MDP} & 177.23 & \multicolumn{1}{c|}{100.0\%} & 125.86 & 100.0\% \\ 
\multicolumn{1}{|l|}{CP}  & 97.40  & \multicolumn{1}{c|}{55.0\%}  & 318.19 & 252.8\% \\
\hline \hline
$J_0$ = 3                                  &        &                              &        &         \\ \hline
\multicolumn{1}{|l|}{RL ignoring MR}   & 132.00 & \multicolumn{1}{c|}{74.5\%}  & 221.68 & 176.1\% \\
\multicolumn{1}{|l|}{Myopic Policy}    & 151.22 & \multicolumn{1}{c|}{85.3\%}  & 166.14 & 132.0\% \\
\multicolumn{1}{|l|}{RL with MR}       & 159.16 & \multicolumn{1}{c|}{89.8\%}  & 128.76 & 102.3\% \\ \hline \hline
$J_0$ = 10                                 &        &                              &        &         \\ \hline
\multicolumn{1}{|l|}{RL ignoring MR}   & 143.47 & \multicolumn{1}{c|}{81.0\%}  & 196.39 & 156.0\% \\
\multicolumn{1}{|l|}{Myopic Policy}    & 168.21 & \multicolumn{1}{c|}{94.9\%}  & 166.31 & 132.1\% \\
\multicolumn{1}{|l|}{RL with MR}       & 171.63 & \multicolumn{1}{c|}{96.8\%}  & 126.68 & 100.7\% \\ \hline \hline
$J_0 = 20$                                 &        &                              &        &         \\ \hline
\multicolumn{1}{|l|}{RL ignoring MR}   & 153.98 & \multicolumn{1}{c|}{86.9\%}  & 165.79 & 131.7\% \\
\multicolumn{1}{|l|}{Myopic Policy}    & 175.34 & \multicolumn{1}{c|}{98.9\%}  & 127.74 & 101.5\% \\
\multicolumn{1}{|l|}{RL with MR}       & 176.30 & \multicolumn{1}{c|}{99.5\%}  & 126.26 & 100.3\% \\ \hline
\end{tabular}
\end{table}

We observe from Table~\ref{table:compare_0} that the strategy RL with MR provides substantial benefits (in terms of the performance metrics $\widehat{\rho}^c$ and $\widehat{\mbox{SD}}^c$) compared to all other strategies. In addition, we notice that the average reward of all strategies increases while variability decreases as the number of historical data increases. In this specific case study, we also observe that CP does not perform well compared to PI-MDP. For practitioners, these results emphasize the business value (and the potential impact) of accounting for the model risk in harvesting decisions. In addition, we observe that the strategy RL with MR results in a lower standard deviation $\widehat{\mbox{SD}}^c$ compared to current practice CP, even when the amount of historical data is small $(J_0=3)$. This observation is interesting because the objective of the optimization model is to maximize the expected total reward (not to minimize variability). We obtained a similar result from the implementation at MSD (i.e., variability reduced after the implementation, as discussed in Section~\ref{sec:impact}). For managers, the results reported in Table 1 underscore the importance of considering both model risk and inherent stochasticity of fermentation systems to achieve higher profits. 

\begin{figure}
    \centering
    \includegraphics[scale=0.85]{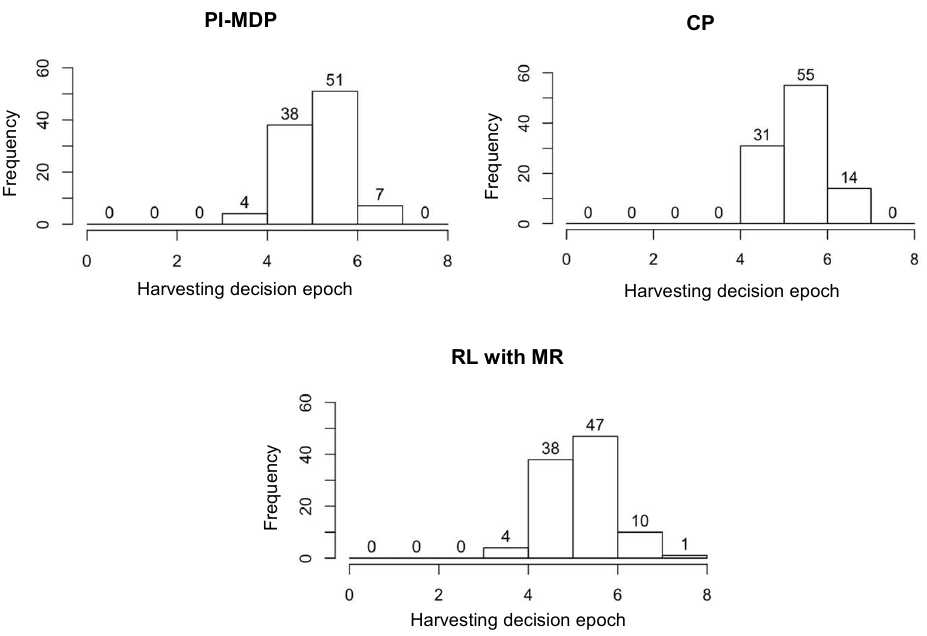}
    \caption{The frequency of the number of decision epochs at which the harvesting decision is made for the case with $J_0=3$  (out of 100 simulation replications).}
    \label{fig:enter-label}
\end{figure}

Figure~\ref{fig:enter-label} provides an analysis of the harvesting times under PI-MDP, CP and RL with MR strategies to better understand the reason for the performance difference between these strategies.
Specifically, we record the decision epoch at which the harvesting decision is realized in each of the 100 simulation replications (similar to obtaining the results in Table~\ref{table:compare_0} as described in Appendix~\ref{subsec: performance metrics}), and we plot the histogram of these realized harvesting decision epochs (recall that the time between two decision epochs is six hours). As a managerial insight, Figure~\ref{fig:enter-label} indicates that the poor performance of the CP strategy in our case study is due to late harvesting decisions, as can be seen by comparing the harvesting decision epochs under CP with those under PI-MDP. 
This can be explained by the tendency of the CP policy to collect as much protein as possible by ignoring impurity-related costs and failure risk.
However, our proposed `RL with MR' strategy takes into account the model risk, and the distribution of harvesting moments becomes similar to that of the PI-MDP, leading to the superior performance of `RL with MR' as observed in Table~\ref{table:compare_0}.

\subsection{Impact of Limited Historical Data on Harvesting Decisions} \label{subsec:empirical1}

\begin{figure}[h!]{
		\centering
		\includegraphics[scale=0.525]{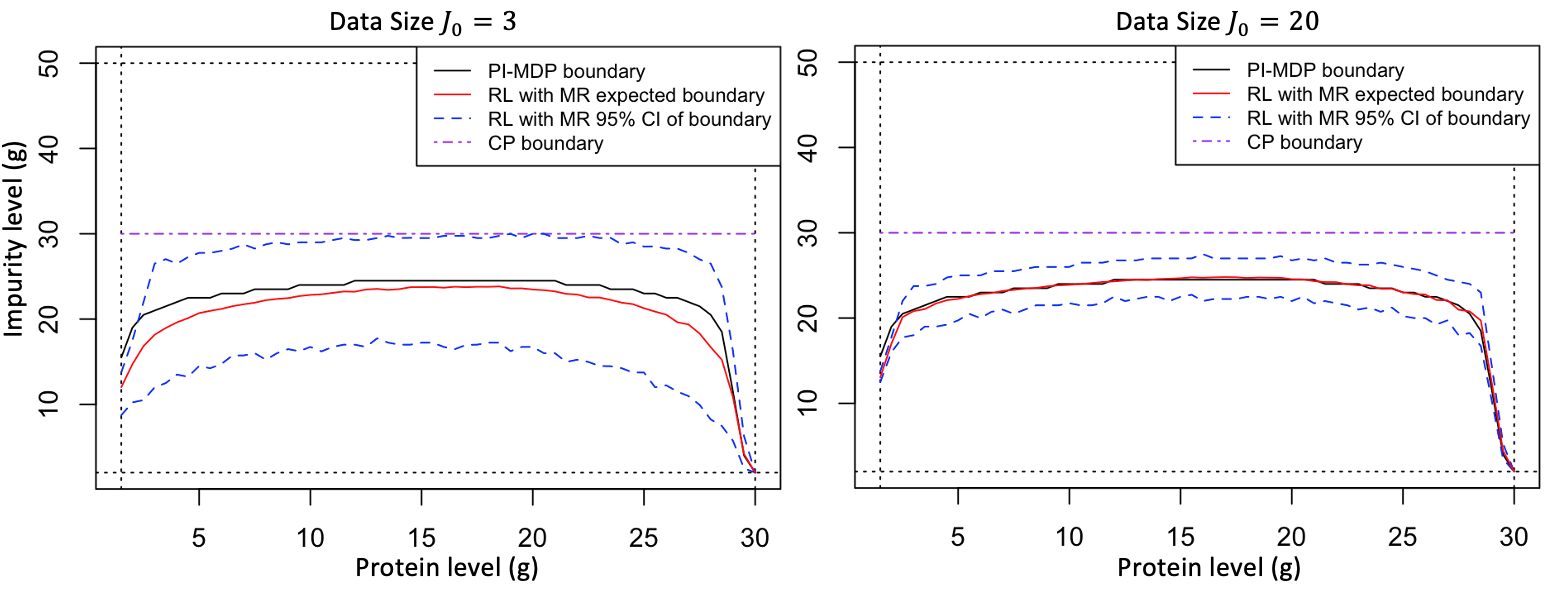}
		\caption{Optimal harvesting thresholds (above curve denotes harvest region) with $J_0 \in \{3, 20\}$ under the strategies PI-MDP, RL with MR, and CP.
		\label{fig:emp_boundary}}}
		\vspace{-0.1in}
\end{figure}

We illustrate the impact of model risk on harvesting decisions. For this purpose, our analysis considers two different sizes of historical data, $J_0 = 3, 20$. We use the strategy PI-MDP as a benchmark to represent the case with perfect information. 
Figure~\ref{fig:emp_boundary}
represents the optimal harvesting policy under the strategy PI-MDP  
for the physical states $p_t \in [1.5, 30]$ and $i_t \in [2.0, 50]$. Moreover, Figure~\ref{fig:emp_boundary} shows the fixed-threshold based CP, and the mean harvesting threshold under the strategy RL with MR  with its corresponding $95\%$ confidence band. 


Figure~\ref{fig:emp_boundary} indicates how the optimal harvesting threshold moves as the number of historical data $J_0$ increases from $3$ to $20$. As the number of historical data increases, we see that the $95\%$ confidence band of the harvest boundary shrinks and becomes closer to the one under the optimal harvest policy with perfect information. Finally, the structure of the RL with MR policy in Figure~\ref{fig:emp_boundary} verifies our analytical finding on the optimality of a threshold-type policy with respect to the impurity level, but shows the similar structure does not hold with respect to the protein level. Intuitively, if the protein level is too close to zero, the impurity/protein ratio is already too high and it becomes beneficial to stop the process by harvesting. On the other hand, if the protein level is too close to its upper limit, harvesting becomes more advantageous again, as continuing the fermentation would only add more impurities with little gain in protein.

\subsection{Impact of Prior Information}

In industrial applications, contextual information may be available either through expert judgments or historical data from similar processes, and this information can provide an informed starting point to build prior information on unknown process parameters. Our results so far assumed non-informative priors when there is no historical data (see Appendix~\ref{sec:optimizationAlgorithm}). Considering the availability of historical data of size $J_0$ (collected from the fermentation process of interest), we built the prior distribution by updating the non-informative prior with this historical data of size $J_0$ as described in Section~\ref{subsec:onlineLearning}. Thus, the change of values in Table~\ref{table:compare_0} with respect $J_0$ shows how the amount of correct prior information influences the total rewards. For example, we observe in Table~\ref{table:compare_0} that the amount of prior information obtained from $J_0=3$ data points leads to $89.8\%$ of the true optimal reward for the `RL with MR' policy, while this value reaches $99.5\%$ when the prior information is obtained from $J_0=20$ data points from the fermentation process. Therefore, the analysis in $J_0$ can also help practitioners understand how much data is needed to achieve high performance. 

The prior information may not always be an accurate representation of reality, e.g., two fermentation processes that are thought to be similar (e.g., using similar seed cultures) may eventually behave very differently. In Table~\ref{table:compare_prior}, we investigate how the total rewards and harvesting times would change if the prior information were obtained from historical data generated from another fermentation process with mean protein and impurity growth rates equal to $k \mu_c^{(p)}$ and $k \mu_c^{(i)}$, respectively, for $k\in\{0.25, 0.5, 1, 2, 4\}$. Here, $k$ represents a deviation factor of the historical data from the true fermentation process (notice that $k=1$ means that the historical data has been collected from the true fermentation process and it leads to the results in Table~\ref{table:compare_0}).

\begin{table}[hbt!]
\caption{The mean and standard deviation of the total rewards and harvesting decision epochs under the RL with MR policy starting with different prior information obtained from $J_0=3$ data points.}
\label{table:compare_prior}
\centering
\begin{tabular}{|lcccc|}
\hline
\multicolumn{1}{|l|}{$k$}                 & $\widehat{\rho}^c(\pi)$   & \multicolumn{1}{c|}{$\widehat{\mbox{SD}}^c(\pi)$}    & Avg. $T$     & Std. dev. $T$  \\ \hline
\multicolumn{1}{|l|}{$0.25$ }       & 122.60 & \multicolumn{1}{c|}{221.07}  & 5.45 & 1.05 \\
\multicolumn{1}{|l|}{$0.5$}       & 147.57 & \multicolumn{1}{c|}{166.15}  & 5.45 & 0.77 \\ 
 \multicolumn{1}{|l|}{$1$}       & 159.16  & \multicolumn{1}{c|}{128.76}  & 5.42 & 0.82 \\ 
\multicolumn{1}{|l|}{$2$}       & 127.22 & \multicolumn{1}{c|}{69.10}  & 4.69 & 0.90 \\ 
\multicolumn{1}{|l|}{$4$ }       & 34.86 & \multicolumn{1}{c|}{23.42}  & 1.98 & 0.98 \\ \hline
\end{tabular}
\end{table}

Table~\ref{table:compare_prior} allows us to understand how the quality of the prior information affects the total rewards and the harvesting times. For example, building the prior by using a data set that comes from a process that has half of the mean growth rates of the current process (i.e., $k=0.5$) leads to a smaller reduction in rewards than building the prior by using a data set from a process with twice the mean growth rates (i.e., $k=2$). A similar observation can be made for $k=0.25$ and $k=0.4$ to compare how the prior information affects the total rewards. As a managerial insight, we observe that overestimating the mean growth rates seems to be worse than underestimating them for the problem instances considered in Table~\ref{table:compare_prior}. This is because overestimating the mean growth rates leads to premature harvesting decisions in our problem configuration, as shown in Table~\ref{table:compare_prior}.

\section{Implementation at MSD} \label{Sec:Implement}

We quantify the real-world impact obtained at MSD's daily operations in Section~\ref{sec:impact} and elaborate on the implementation process in Section~\ref{sec:imp}.

\subsection{Impact} \label{sec:impact}

\begin{figure}[h]
    \centering
\includegraphics[scale=0.75]{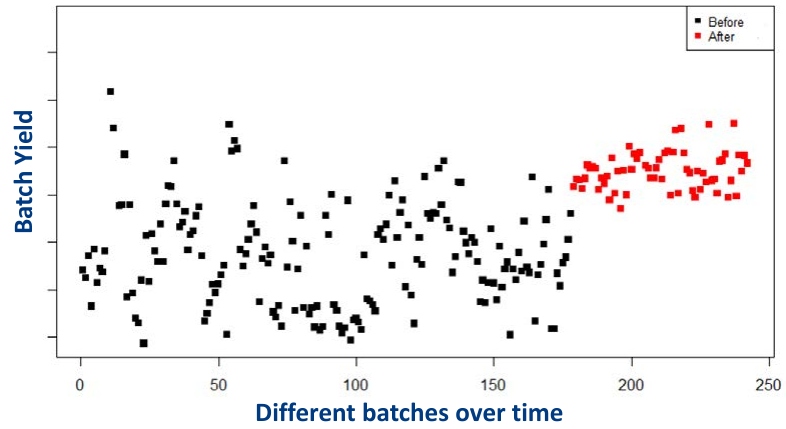}
    \caption{(Color online) Performance of different batches of the same product produced over time: Black dots represent performance before implementation. Red dots represent performance after implementation.}
    \label{fig:Yresults}
\end{figure}

The optimization framework has been implemented at MSD since 2019. The project focused on various products manufactured in Boxmeer and had a significant impact on business metrics. As an example, Figure~\ref{fig:Yresults} shows the implementation results for a specific product. The x-axis represents time and the y-axis represents batch yield (the value of y-axis starts from 0, but its scale is not shown for confidentiality reasons). In this figure, ``batch yield'' represents $p-i$, which is the total amount of protein minus the impurities present in the batch at the time of harvest. The black dots in Figure~\ref{fig:Yresults} indicate the batch yield before implementation, while the red dots correspond to the batch yield after implementation. The figure shows that the average batch yield for this product increased by approximately 50\%, while the batch-to-batch variability decreased significantly after implementation.

Prior to implementation, decisions were made based on domain knowledge and common industry guidelines (e.g., the so-called ``common practice'' strategy based on fixed thresholds and rules of thumb, as described in Section~\ref{Sec:CaseStudy}). However, these prior approaches did not systematically exploit any optimization framework. After implementing the OR-based framework, the company was able to make better use of the historical data and dynamically optimize operational decisions. 

\begin{figure}[h]
    \centering
\includegraphics[scale=0.75]{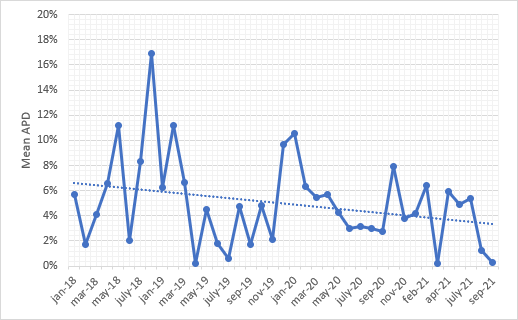}
    \caption{Impact of the learning-by-doing framework across all products considered in the implementation process. 
    }
    \label{fig:Yerror}
\end{figure}

Recall that Figure~\ref{fig:Yresults} illustrates the results obtained for one particular product. We focused on this product due to the availability of a large number of historical data, allowing us to see the benefit gained from the proposed optimization framework when the inherent fermentation uncertainty is more important relative to the model risk. We now present insights based on all products within the scope of the implementation during 2019-2021 (including products with limited data). In this setting, our objective was to quantify the impact of the learning-by-doing framework. For this purpose, we first collected information on the ``expected" batch yield (ex-ante) and the ``actual" batch yield (ex-post) obtained for each batch produced in 2019-2021. The expected batch yield represents our predicted value of the batch yield under a certain harvesting policy used for that batch; whereas the actual batch yield denotes the realized batch yield at the time of harvest. Then, we calculated the \textit{absolute percentage difference} (APD) between expected and actual values for each batch (where the denominator captures the expected yield). We defined the measure APD to understand how our prediction capability changed over time as a result of the learning-by-doing framework. For ease of exposition, Figure~\ref{fig:Yerror} plots the mean APD values on a monthly basis (i.e., the average of APD values across all batches produced in a certain month). In this figure, the implementation started around January 2019. We see a clear downward trend in Figure~\ref{fig:Yerror}, indicating that our predictive capability continuously improved after implementing the data-driven decision framework. We quantified the impact in terms of batch yield (rather than financial figures) for confidentiality reasons.

\subsection{Overview of the Implementation Process} \label{sec:imp}

\noindent \textbf{The team and project timeline.} The research project has been conducted in close collaboration with a university team and a team of practitioners from MSD in the Netherlands. The university team brought in expertise on operations research, whereas the MSD team provided expertise in life sciences. A multi-disciplinary team from MSD (e.g., bioreactor operators, chemical and biological engineers, and middle/upper management) contributed to the project.

The project went through three major phases: model development, validation, and implementation. The project started in early 2018 with data collection and the development of the optimization framework. Prior to the implementation, the optimization model and the corresponding policies were validated based on \textit{(i)} discussions with practitioners, and \textit{(ii)} small-scale test runs. 
The control-limit structure of optimal policies facilitated these discussions, as these policies were ``explainable" and their intuition aligned with the current practice. Following these discussions, small-scale test runs were performed. The results obtained from both computational experiments and real-world test runs established a foundation for industry-scale implementation.

\noindent \textbf{Scope and aspects related to data.} The implementation focused on industry-scale production orders with limited historical process data. As a common characteristic, these products were typically high-mix, low-volume batches produced only a few times a year. Moreover, we encountered challenges with limited data when new equipment or raw materials were used.  
Available data typically involved around $10$ batches. 
However, some products had as little as one or two data sets. In some cases, we had \textit{no} data to build the prior distribution because of a change in raw materials or equipment. All batches exhibited inherent stochasticity and model risk. 


\noindent \textbf{Implementation challenges.} The major challenge in this project was related to data collection. In some cases, data was not available in a digital format. 
We also encountered a few special cases with no historical data. 
%
One of the strengths of the project is its multi-disciplinary approach which combines life sciences and operations research. However, multi-disciplinary projects have their own challenges. For example, the concept of Markov decision processes may be difficult for scientists who have no background in operations research. Similarly, developing a thorough understanding of the fermentation processes was challenging for the university team (with no background in biological and chemical engineering). Therefore, both the university and MSD teams had regular meetings to learn from each other and co-design the model.

It is also important to facilitate the knowledge transfer to other facilities. When the scope of the implementation expands to other facilities in the future, it can be challenging to identify the right products (and facilities) that would obtain the highest benefit from the optimization framework. For this purpose, MSD developed a dashboard that collects information from all batches produced globally. This dashboard reports the APD values of selected batches (as illustrated in Figure~\ref{fig:Yerror}) from their global network, thereby identifying opportunities for future implementations.

\section{Conclusions} \label{sec:concl}

Limitations in historical process data (model risk) are often perceived as a common industry challenge in biomanufacturing. Yet the implications of model risk on optimal costs and harvesting decisions have not been fully understood. Our work provides one of the first attempts at modeling and optimization of fermentation systems under model risk (caused by limited historical data) and inherent stochasticity (caused by the uncertain nature of biological systems). 

We developed an MDP model to guide fermentation harvesting decisions under a learning-by-doing framework. In particular, we used a Bayesian approach where the decision-maker sequentially collects real-world data on fermentation dynamics and updates the beliefs on state transitions. As a salient feature, the MDP model combines the knowledge from life sciences and operations research, and is equipped to capture the complex dynamics of fermentation processes under limited data. 
We studied the analytical properties of the optimal policy and the myopic policy, and characterized the impact of model risk on biomanufacturing harvesting decisions. To illustrate the use of the optimization model, we present a case study from MSD Animal Health. The implementation at MSD has shown that linking operations research with life sciences drives substantial productivity improvements. We hope that our results will inspire the global biomanufacturing industry and stimulate new research at the intersection of operations research and life sciences.

The developed optimization framework is generic and addresses common industry challenges. Therefore, it can be easily implemented at other production lines and facilities. 
The long-term vision is to encourage the worldwide use of such optimization models at other facilities. Moreover, it will be interesting to explore the potential applications in other industries. For example, the food and agriculture industries face similar challenges and can benefit from the developed optimization model. In addition, future research could explore optimal sampling decisions based on the costs and marginal benefits of additional data collection efforts. Another research direction could extend our framework to continuous-time models to support real-time fermentation decisions.

\section*{Acknowledgment}
This research was funded by the Dutch Science Foundation (NWO-VENI Scheme) and the National Institute of Standards and
Technology (Grant nos. 70NANB17H002 and 70NANB21H086). We would like to thank the Master's students Thijs Diessen and Len Hermsen for their assistance during the project.

\theendnotes


\bibliographystyle{unsrt} 
\bibliography{proj_ref,proposal} 

\begin{thebibliography}{10}

\bibitem{doran2013bioprocess}
P.M. Doran.
\newblock {\em Bioprocess Engineering Principles}.
\newblock Elsevier, Amsterdam, 2013.

\bibitem{plante1999improving}
Robert Plante, Herbert Moskowitz, Jen Tang, and Jeff Duffy.
\newblock Improving quality via matching: A case study integrating supplier and
  manufacturer quality performance.
\newblock {\em Manufacturing \& Service Operations Management}, 1(1):36--49,
  1999.

\bibitem{martaganPOM}
Tugce Martagan, Ananth Krishnamurthy, Peter~A Leland, and Christos~T
  Maravelias.
\newblock Optimal purification decisions for engineer-to-order proteins at
  aldevron.
\newblock {\em Production and Operations Management}, 25(12):2003--2005, 2016.

\bibitem{sahling2019dynamic}
Florian Sahling and Gerd~J Hahn.
\newblock Dynamic lot sizing in biopharmaceutical manufacturing.
\newblock {\em International Journal of Production Economics}, 207:96--106,
  2019.

\bibitem{subramanian2020pulling}
Annapoornima~M Subramanian, Moren L{\'e}vesque, and Vareska Van De~Vrande.
\newblock ``{P}ulling the plug''': {T}ime allocation between drug discovery and
  development projects.
\newblock {\em Production and Operations Management}, 29(12):2851--2876, 2020.

\bibitem{zhu2021demand}
Xiaodan Zhu, Anh Ninh, Hui Zhao, and Zhenming Liu.
\newblock Demand forecasting with supply-chain information and machine
  learning: Evidence in the pharmaceutical industry.
\newblock {\em Production and Operations Management}, 30(9):3231--3252, 2021.

\bibitem{zhao2023pharmaceutical}
Hui Zhao.
\newblock Pharmaceutical supply chains and drug shortages.
\newblock In {\em Tutorials in Operations Research: Advancing the Frontiers of
  OR/MS: From Methodologies to Applications}, pages 228--245. INFORMS, 2023.

\bibitem{xu2023not}
Liang Xu, Vidya Mani, and Hui Zhao.
\newblock ``{N}ot a box of nuts and bolts'': Distribution channels for
  specialty drugs.
\newblock {\em Production and Operations Management}, 32(7):2283--2303, 2023.

\bibitem{lowe2004decision}
Timothy~J Lowe and Paul~V Preckel.
\newblock Decision technologies for agribusiness problems: A brief review of
  selected literature and a call for research.
\newblock {\em Manufacturing \& Service Operations Management}, 6(3):201--208,
  2004.

\bibitem{azoury2013managing}
Katy~S Azoury and Julia Miyaoka.
\newblock Managing production and distribution for supply chains in the
  processed food industry.
\newblock {\em Production and Operations Management}, 22(5):1250--1268, 2013.

\bibitem{bansal2017product}
Saurabh Bansal and Mahesh Nagarajan.
\newblock Product portfolio management with production flexibility in
  agribusiness.
\newblock {\em Operations Research}, 65(4):914--930, 2017.

\bibitem{rajaram2004campaign}
Kumar Rajaram and Uday~S Karmarkar.
\newblock Campaign planning and scheduling for multiproduct batch operations
  with applications to the food-processing industry.
\newblock {\em Manufacturing \& Service Operations Management}, 6(3):253--269,
  2004.

\bibitem{jahandideh2020production}
Hossein Jahandideh, Kumar Rajaram, and Kevin McCardle.
\newblock Production campaign planning under learning and decay.
\newblock {\em Manufacturing \& Service Operations Management}, 22(3):615--632,
  2020.

\bibitem{blackburn2009supply}
Joseph Blackburn and Gary Scudder.
\newblock Supply chain strategies for perishable products: the case of fresh
  produce.
\newblock {\em Production and Operations Management}, 18(2):129--137, 2009.

\bibitem{ARC2024}
Saurabh Bansal, Phil Coles, Dongsheng Li, and Karthik Natrajan.
\newblock Redesigning harvesting processes and improving working conditions in
  agribusiness.
\newblock Working paper, 2024.

\bibitem{McNeil08}
B.~McNeil and L.~M. Harvey.
\newblock {\em Practical Fermentation Technology}.
\newblock 2008.
\newblock John Wiley \& Sons, New York.

\bibitem{Putra18}
Meilana~Dharma Putra and Ahmed~E. Abasaeed.
\newblock A more generalized kinetic model for binary substrates fermentations.
\newblock {\em Process Biochemistry}, 75:31--38, 2018.

\bibitem{chang2016nonlinear}
Liang Chang, Xinggao Liu, and Michael~A Henson.
\newblock Nonlinear model predictive control of fed-batch fermentations using
  dynamic flux balance models.
\newblock {\em Journal of Process Control}, 42:137--149, 2016.

\bibitem{Peroni05}
C.~V. Peroni, N.~S. Kaisare, and J.~H. Lee.
\newblock Optimal control of fed-batch bioreactor using simulation based
  approximate dynamic programming.
\newblock {\em IEEE Transactions on Control Systems Technology}, 13:786--790,
  2005.

\bibitem{xing2010modeling}
Zizhuo Xing, Nikki Bishop, Kirk Leister, and Zheng~Jian Li.
\newblock Modeling kinetics of a large-scale fed-batch cho cell culture by
  markov chain monte carlo method.
\newblock {\em Biotechnology Progress}, 26(1):208--219, 2010.

\bibitem{martagan2016optimal}
Tugce Martagan, Ananth Krishnamurthy, and Christos~T Maravelias.
\newblock Optimal condition-based harvesting policies for biomanufacturing
  operations with failure risks.
\newblock {\em IIE Transactions}, 48(5):440--461, 2016.

\bibitem{martagan2018managing}
T.G Martagan, A.~Krishnamurthy, and P.~Leland.
\newblock Managing trade-offs in protein manufacturing: How much to waste?
\newblock {\em Manufacturing \& Service Operations Management}, 22(2):223--428,
  2020.

\bibitem{martagan2022}
Tugce Martagan, Ivo Adan, Marc Baaijens, Coen Dirckx, Oscar Repping, Bram van
  Ravenstein, and PK~Yegneswaran.
\newblock Merck animal health uses operations research methods to transform
  biomanufacturing productivity for lifesaving medicines.
\newblock {\em Franz Edelman Special Issue of INFORMS Journal on Applied
  Analytics}, 53(1):85--95, 2023.

\bibitem{koca2023increasing}
Yesim Koca, Tugce Martagan, Ivo Adan, Lisa Maillart, and Bram van Ravenstein.
\newblock Increasing biomanufacturing yield with bleed--feed: Optimal policies
  and insights.
\newblock {\em Manufacturing \& Service Operations Management}, 25(1):108--125,
  2023.

\bibitem{xie2022interpretable}
Wei Xie, Bo~Wang, Cheng Li, Dongming Xie, and Jared Auclair.
\newblock Interpretable biomanufacturing process risk and sensitivity analyses
  for quality-by-design and stability control.
\newblock {\em Naval Research Logistics (NRL)}, 69(3):461--483, 2022.

\bibitem{treloar2020deep}
Neythen~J Treloar, Alex~JH Fedorec, Brian Ingalls, and Chris~P Barnes.
\newblock Deep reinforcement learning for the control of microbial co-cultures
  in bioreactors.
\newblock {\em PLOS Computational Biology}, 16(4):e1007783, 2020.

\bibitem{nikita230reinforcement}
Saxena Nikita, Anamika Tiwari, Deepak Sonawat, Hariprasad Kodamana, and
  Anurag~S. Rathore.
\newblock Reinforcement learning based optimization of process chromatography
  for continuous processing of biopharmaceuticals.
\newblock {\em Chemical Engineering Science}, 230:116171, 2021.

\bibitem{zheng2020paper}
Hua Zheng, Wei Xie, and M.~Ben Feng.
\newblock Green simulation assisted reinforcement learning with model risk for
  biomanufacturing learning and control.
\newblock In {\em Proceedings of the 2020 Winter Simulation Conference}, page
  337–348, 2020.

\bibitem{Zheng2023_JoC}
Hua Zheng, Wei Xie, Ilya~O. Ryzhov, and Dongming Xie.
\newblock Policy optimization in dynamic bayesian network hybrid models of
  biomanufacturing processes.
\newblock {\em INFORMS Journal on Computing}, 35(1):66–82, 2023.

\bibitem{ghavamzadeh2016bayesian}
Mohammad Ghavamzadeh, Shie Mannor, Joelle Pineau, and Aviv Tamar.
\newblock Bayesian reinforcement learning: A survey.
\newblock {\em arXiv preprint arXiv:1609.04436}, 2016.

\bibitem{poupart2006analytic}
Pascal Poupart, Nikos Vlassis, Jesse Hoey, and Kevin Regan.
\newblock An analytic solution to discrete bayesian reinforcement learning.
\newblock In {\em Proceedings of the 23rd International Conference on Machine
  Learning}, pages 697--704, 2006.

\bibitem{ross2008bayesian}
Stephane Ross, Brahim Chaib-draa, and Joelle Pineau.
\newblock Bayesian reinforcement learning in continuous pomdps with application
  to robot navigation.
\newblock In {\em 2008 IEEE International Conference on Robotics and
  Automation}, pages 2845--2851, 2008.

\bibitem{osband2013more}
Ian Osband, Daniel Russo, and Benjamin Van~Roy.
\newblock ({M}ore) efficient reinforcement learning via posterior sampling.
\newblock In {\em Advances in Neural Information Processing Systems}, pages
  3003--3011, 2013.

\bibitem{fonteneau2013optimistic}
Raphael Fonteneau, Lucian Bu{\c{s}}oniu, and R{\'e}mi Munos.
\newblock Optimistic planning for belief-augmented markov decision processes.
\newblock In {\em 2013 IEEE Symposium on Adaptive Dynamic Programming and
  Reinforcement Learning}, pages 77--84, 2013.

\bibitem{kolter2009near}
J~Zico Kolter and Andrew~Y Ng.
\newblock Near-bayesian exploration in polynomial time.
\newblock In {\em Proceedings of the 26th annual international conference on
  machine learning}, pages 513--520, 2009.

\bibitem{asmuth2011approaching}
John Asmuth and Michael~L Littman.
\newblock Approaching {B}ayes-optimalilty using monte-carlo tree search.
\newblock In {\em Proceedings of the 21st International Conference on Automated
  Plannning and Scheduling.}, 2011.

\bibitem{asmuth2012bayesian}
John Asmuth, Lihong Li, Michael~L Littman, Ali Nouri, and David Wingate.
\newblock A {B}ayesian sampling approach to exploration in reinforcement
  learning.
\newblock {\em arXiv preprint arXiv:1205.2664}, 2012.

\bibitem{templeton2013peak}
Neil Templeton, Jason Dean, Pranhitha Reddy, and Jamey~D Young.
\newblock Peak antibody production is associated with increased oxidative
  metabolism in an industrially relevant fed-batch cho cell culture.
\newblock {\em Biotechnology and Bioengineering}, 110(7):2013--2024, 2013.

\bibitem{odenwelder2021induced}
Daniel~C Odenwelder, Xiaoming Lu, and Sarah~W Harcum.
\newblock Induced pluripotent stem cells can utilize lactate as a metabolic
  substrate to support proliferation.
\newblock {\em Biotechnology Progress}, 37(2):e3090, 2021.

\bibitem{tsao2005monitoring}
Y-S Tsao, AG~Cardoso, RGG Condon, M~Voloch, P~Lio, JC~Lagos, BG~Kearns, and
  Z~Liu.
\newblock Monitoring chinese hamster ovary cell culture by the analysis of
  glucose and lactate metabolism.
\newblock {\em Journal of Biotechnology}, 118(3):316--327, 2005.

\bibitem{wechselberger2013model}
Patrick Wechselberger, Patrick Sagmeister, and Christoph Herwig.
\newblock Model-based analysis on the extractability of information from data
  in dynamic fed-batch experiments.
\newblock {\em Biotechnology Progress}, 29(1):285--296, 2013.

\bibitem{mockus2015batch}
Linas Mockus, John~J Peterson, Jose~Miguel Lainez, and Gintaras~V Reklaitis.
\newblock Batch-to-batch variation: a key component for modeling chemical
  manufacturing processes.
\newblock {\em Organic Process Research \& Development}, 19(8):908--914, 2015.

\bibitem{moller2020model}
Johannes M{\"o}ller, Tanja~Hern{\'a}ndez Rodr{\'\i}guez, Jan M{\"u}ller, Lukas
  Arndt, Kim~B Kuchem{\"u}ller, Bj{\"o}rn Frahm, Regine Eibl, Dieter Eibl, and
  Ralf P{\"o}rtner.
\newblock Model uncertainty-based evaluation of process strategies during
  scale-up of biopharmaceutical processes.
\newblock {\em Computers \& Chemical Engineering}, 134:106693, 2020.

\bibitem{Gelman_2004}
A.~Gelman, J.~B. Carlin, H.~S. Stern, and D.~B. Rubin.
\newblock {\em Bayesian Data Analysis}.
\newblock Taylor and Francis Group, LLC, New York, 2nd edition, 2004.

\bibitem{powell2012optimal}
Warren~B Powell and Ilya~O Ryzhov.
\newblock {\em Optimal Learning}, volume 841.
\newblock John Wiley \& Sons, 2012.

\bibitem{murphy2007conjugate}
Kevin~P Murphy.
\newblock Conjugate bayesian analysis of the gaussian distribution.
\newblock Technical report, University of British Columbia, 2007.

\bibitem{ferguson2000optimal}
Thomas~S. Ferguson.
\newblock Optimal stopping and applications, 2000.
\newblock [Online; accessed 10-Apr-2022].

\bibitem{kearns2002sparse}
Michael Kearns, Yishay Mansour, and Andrew~Y Ng.
\newblock A sparse sampling algorithm for near-optimal planning in large markov
  decision processes.
\newblock {\em Machine Learning}, 49(2-3):193--208, 2002.

\bibitem{wang2005bayesian}
Tao Wang, Daniel Lizotte, Michael Bowling, and Dale Schuurmans.
\newblock Bayesian sparse sampling for on-line reward optimization.
\newblock In {\em Proceedings of the 22nd International Conference on Machine
  Learning}, pages 956--963, 2005.

\bibitem{ross2008model}
St{\'e}phane Ross and Joelle Pineau.
\newblock Model-based bayesian reinforcement learning in large structured
  domains.
\newblock In {\em Conference on Uncertainty in Artificial Intelligence}, volume
  2008, page 476, 2008.

\end{thebibliography}

\newpage

\appendix

\section{Summary of notations}
\label{sec: notationtable}

\begin{table}[h]
  \centering
{\footnotesize
    \begin{tabular}{|p{3cm} p{12cm}|} 
\hline     
    $p_t$   & protein amount at time (decision epoch) $t$ \\
    $p_0$   & the starting amount of protein \\
    $\bar{P}$ & upper limit on the accumulated protein level \\
    $\phi$ & the specific growth rate of protein (under deterministic growth rate assumption)  \\
    $i_t$   & impurity amount at time $t$ \\
    $i_0$   & the starting amount of impurity \\
    $\bar{I}$ & the impurity value at which the batch is considered as failed \\
    $\psi$ & the specific growth rate of impurity (under deterministic growth rate assumption) \\
    $T$ & the harvesting time \\ 
    $\bar{T}$ & the upper bound on the harvesting time \\
    $\Phi_t$ & the random variable representing the specific growth rate of protein in the time interval from time $t$ to $t+1$ \\ 
    $\phi_{t}$ & the realization of $\Phi_t$ \\    
    $\widetilde{\Phi}_{t}$ & the predictive protein growth-rate random variable at time $t$ \\
    $\Psi_t$ & the random variable representing the specific growth rate of impurity in the time interval from time $t$ to $t+1$ \\
    $\psi_{t}$ & the realization of $\Psi_t$ \\
    $\widetilde{\Psi}_{t}$ & the predictive impurity growth-rate random variable at time $t$ \\
        $\hat{\sigma}_{t}^{(\cdot)2}$ & variance of the growth rate due to inherent stochasticity \\ $\check{\sigma}_{t}^{(\cdot)2}$ & variance of the growth rate due to the model risk \\
    $\mathcal{N}(a,b)$ & a normal distribution with mean $a$ and variance $b$ \\
    $\pmb{\theta}^c$ & the true (unknown) parameters of the underlying stochastic fermentation model  \\
    $\pmb{\theta}$ & the random variables modeling the decision-maker's uncertainty in  $\pmb{\theta}^c$ \\
    $\alpha_{t}^{(p)}, \nu_{t}^{(p)}, \lambda_{t}^{(p)}, \beta_{t}^{(p)}$ & the prior parameters for protein growth rate at time $t$ \\
    $\alpha_{t}^{(i)}, \nu_{t}^{(i)}, \lambda_{t}^{(i)}, \beta_{t}^{(i)}$ & the prior parameters for impurity growth rate at time $t$ \\
    $\mathcal{S}_t = (p_t, i_t)$ & the physical state at time $t$ \\ 
    $\mathcal{I}_t$ & the knowledge state at time $t$ \\
    $\mathcal{H}_t$ & hyper state (combination of physical and knowledge states) at time $t$ \\
    $a_t$ & action at time $t$ \\
    $\{C, H \}$ & action space, denoting continue and harvest decisions, respectively \\
    $\Delta$ & absorbing stopping state after the harvest decision \\
    $c_u$ & the cost associated with continuing the fermentation one more time unit  \\
    $r_f$ & the cost of harvesting a failed batch due to losing the batch  \\
    $r_h(p_t,i_t)$ & the reward from harvesting a non-failed batch with protein  $p_t$ and impurity  $i_t$. \\
    $R(p_t, i_t; a_t)$ & the reward at time $t$ by taking action $a_t$ at states $p_t$ and $i_t$\\
    $\gamma$ & the discount factor \\
    $\pi^*$ & optimal policy that maximizes the expected total discounted reward \\
    $V_t(\mathcal{H}_t)$ & the expected total discounted reward starting  with hyper state $\mathcal{H}_t$ at time $t$ under the optimal policy $\pi^*$ \\
    $Q_t(p_{t}, i_{t}, \mathcal{I}_{t}; C)$ & Q-function for the total reward associated with the continue action at time $t$ and following the optimal policy thereafter
    \\
    $\mathcal{D}_t$ & historical data on the realized protein and impurity growth rates available at time~$t$ \\
    $J_t$ & size of the historical data $\mathcal{D}_t$\\
    $\widehat{\rho}^c(\pi)$ & estimated expected total reward for policy $\pi$
    \\
    $\widehat{\mbox{SD}}^c(\pi)$ & estimated standard deviation of total reward for policy $\pi$
    \\
    \hline
    \end{tabular}%
    }
  \label{tab: nomen}%
\end{table}%

\section{Solution Approach}

\subsection{Reinforcement Learning under Model Risk}
\label{sec:optimizationAlgorithm}

\begin{figure}[h]
	\centering
	\includegraphics[scale=0.575]{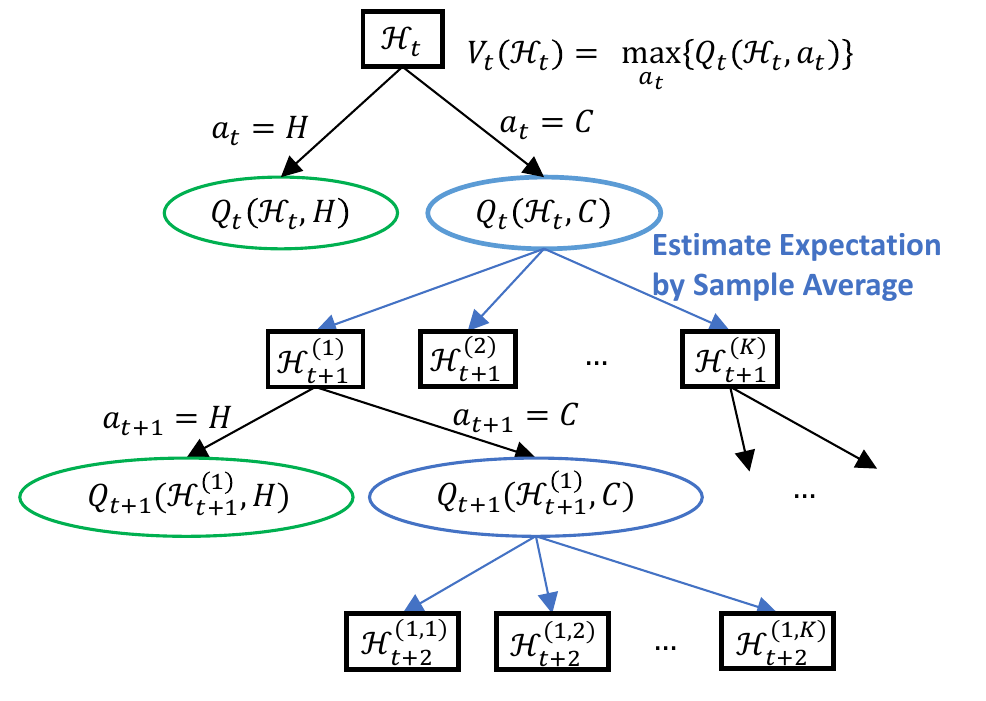}
	\caption{Illustration of the lookahead tree for Bayesian sparse sampling.}
	\label{fig:sparse_tree}
	\vspace{-0.in}
\end{figure}

The solution procedure of Reinforcement Learning under Model Risk (RL with MR) is provided in Algorithm~\ref{alg:online_control}. We initialize in Step~(1).
In Step~(2), the Q-function needs to be estimated. For this purpose, we apply the sparse sampling idea of \cite{kearns2002sparse}, which generates future trajectories (i.e., state-action pairs over time) for modeling the uncertainty in the hyper state evolution and chooses the optimal action by constructing a look-ahead tree with these trajectories. Similar to \cite{wang2005bayesian} and \cite{ross2008bayesian}, we generate the future states by using the Bayesian updated posterior predictive distributions. To be specific, given the current hyper state $\mathcal{H}_t =(\mathcal{S}_t, \mathcal{I}_t)$ with $\mathcal{S}_t=(p_{t}, i_{t})$, the samples of next hyper state can be generated through the transition probability $\mbox{Pr}(\mathcal{S}_{t+1}, \mathcal{I}_{t+1} | \mathcal{S}_t, \mathcal{I}_t, a_t) = \mbox{Pr}(\mathcal{S}_{t+1} | \mathcal{S}_t, \mathcal{I}_t, a_t)\mbox{Pr}( \mathcal{I}_{t+1} | \mathcal{S}_{t+1}, \mathcal{S}_t, \mathcal{I}_t, a_t)$ in two steps. First, we generate the $K$ scenarios of physical states $(p_{t+1}, i_{t+1})$ by sampling the protein growth rates $\widetilde{\phi}_{t}^{(k)}$ and $\widetilde{\psi}_{t}^{(k)}$ from the posterior predictive distributions in \eqref{eq.predictive} and \eqref{eq.predictiveImp}, and then using the equations $p_{t+1}^{(k)} = p_t\cdot e^{\widetilde{\phi}_{t}^{(k)}}$ and $i_{t+1}^{(k)} = i_t\cdot e^{\widetilde{\psi}_{t}^{(k)}}$ for $k=1,2,\ldots,K$.  Second, based on the collected samples of protein and impurity growth rates $\widetilde{\phi}_{t}^{(k)}$ and $ \widetilde{\psi}_{t}^{(k)}$, the knowledge state can be updated by using \eqref{eq.update2} and \eqref{eq.update3}. In Figure~\ref{fig:sparse_tree}, we illustrate the main idea of Bayesian sparse sampling.

\begin{sloppypar}

\begin{algorithm}[hbt!]
\caption{Reinforcement Learning under Model Risk (RL with MR)} \label{alg:online_control}

{\footnotesize
\KwIn{Initial physical states $\mathcal{S}_0 = (p_0, i_0)$ and the initial knowledge states $\mathcal{I}_0$.}

\KwOut{The stopping time $T$ and total reward $Total.R$.}

\SetKwFunction{Fq}{Qfunction}
\SetKwFunction{Fv}{Valuefunction}

(1) Set the total reward $Total.R = 0$. 

\For{$t=0,1,\ldots,\bar{T}$}{
    
    \uIf{$p_t < \bar{P}$ and $i_t < \bar{I}$ and $t < \bar{T}$}{
        (2) Estimate the Q-function in \eqref{eqn: Qcont} by calling $\widehat{Q}_t(p_{t}, i_{t}, \mathcal{I}_{t}; C)=$ \Fq{$p_{t}, i_{t}, \mathcal{I}_{t}, C$}\;
	
        (3) Choose the `harvest' action if $Q_t(p_{t}, i_{t}, \mathcal{I}_{t}; H) \ge \widehat{Q}_t(p_{t}, i_{t}, \mathcal{I}_{t}, C)$, and `continue' otherwise\;
	}
	\uElse{
	    (4) Choose the `harvest' action\;
	}
	
	\uIf{`continue' action is chosen}{    
	    (5) Continue the fermentation process one more period, $Total.R = Total.R -  c_u$, and observe the realized physical states $(p_{t+1}, i_{t+1})$ in real life\;
	    (6) Obtain the updated knowledge states $\mathcal{I}_{t+1}$ as described in Section~\ref{subsec:onlineLearning}\;
	    
	}
	\uElse{
	    (7) Terminate the fermentation process, set the stopping time $T = t$ and $Total.R = Total.R +  R(p_t,i_t; H)$, break the `for loop'\;
	}
}

(8) Output the total reward $Total.R$\;
}
\end{algorithm}

\end{sloppypar}

Bayesian sparse sampling considers a look-ahead tree starting with the root at the current hyper state $\mathcal{H}_t$,  followed by two potential actions: harvest or continue fermentation. The Q-function value of harvest decision is known and given by \eqref{eqn: Qharv}. For the continue action, the expected value $\mbox{E}\left[
\max_{a_{t+1}} Q_{t+1}(p_{t+1}, i_{t+1}, \mathcal{I}_{t+1};
a_{t+1})\right]$ in \eqref{eqn: Qcont} is estimated by using sample average approximation, which grows the tree into $K$ following nodes that each representing a sample of the next hyper state $\mathcal{H}_{t+1}^{(k)} = (p_{t+1}^{(k)}, i_{t+1}^{(k)}, \mathcal{I}_{t+1}^{(k)})$ for $k=1,2,\ldots,K$. The same procedure is repeated for each node until all the leaf nodes reach the end of the fermentation, i.e. the harvest decision is made.
The Q-function (and the value function) estimates are computed from leaf nodes rolling-backing up to the root. The procedure is summarized in Algorithm~\ref{alg:sparse_sampling}. 

\begin{algorithm}[hbt!]
\caption{Bayesian Sparse Sampling for Estimating the Q-Function} \label{alg:sparse_sampling}

{\footnotesize
\KwIn{The current hyper state $\mathcal{H}_t = (\mathcal{S}_t, \mathcal{I}_t)$ with physical states $\mathcal{S}_t = (p_t, i_t)$ and knowledge states $\mathcal{I}_t$; the action $a_t$, and the number of samples $K$.} 

\KwOut{Estimated value of the Q-function $\widehat{Q}_t(p_t, i_t, \mathcal{I}_{t}, a_t)$.}

\SetKwFunction{Fq}{Qfunction}
\SetKwFunction{Fv}{Valuefunction}

\SetKwProg{Fn}{Function}{:}{}
\Fn{\Fq{$p_{t}, i_{t}, \mathcal{I}_{t}$, $a_t$}}{
    \uIf{$a_t = H$}{
        (A1) \KwRet harvest reward $R(p_t, i_t, H)$ as in \eqref{eq.reward}\;
    }
    \uElse{
        (A2) Generate $K$ samples of next hyper state $\mathcal{H}_{t+1}^{(k)} = (p_{t+1}^{(k)}, i_{t+1}^{(k)}, \mathcal{I}_{t+1}^{(k)})$, $k=1, 2, \ldots, K$ from $\mbox{Pr}(\mathcal{S}_{t+1}, \mathcal{I}_{t+1} | \mathcal{S}_t, \mathcal{I}_t, a_t)$ as discussed in Section~\ref{sec:optimizationAlgorithm}\;
        
        (A3) For each $k$, call $\widehat{V}_{t+1}^{(k)}(\mathcal{H}_{t+1}^{(k)}) = $ \Fv{$p_{t+1}^{(k)}, i_{t+1}^{(k)}, \mathcal{I}_{t+1}^{(k)}$} to estimate the value function of hyper state $\mathcal{H}_{t+1}^{(k)}$ at the next decision epoch $t+1$\;
        
        (A4) \KwRet $\widehat{Q}_{t}(\mathcal{H}_{t}, C) = -c_u + \dfrac{\gamma}{K} \sum_{k=1}^K\widehat{V}_{t+1}^{(k)}(\mathcal{H}_{t+1}^{(k)})$\;
    }
}
\medskip
\SetKwProg{Fn}{Function}{:}{}
\Fn{\Fv{$p_{t}, i_{t}, \mathcal{I}_{t}$}}{
    \uIf{$p_t \geq \bar{P}$ or $i_{t} \ge \bar{I}$ or $t = \bar{T}$}{
        (B1) \KwRet harvest reward $R(p_t, i_t, H)$ as in \eqref{eq.reward}\;
    }
    \uElse{
        (B2) Generate $K$ samples of next hyper state $\mathcal{H}_{t+1}^{(k)} = (p_{t+1}^{(k)}, i_{t+1}^{(k)}, \mathcal{I}_{t+1}^{(k)})$, $k=1, 2, \ldots, K$ from $\mbox{Pr}(\mathcal{S}_{t+1}, \mathcal{I}_{t+1} | \mathcal{S}_t, \mathcal{I}_t, a_t)$ as discussed in Section~\ref{sec:optimizationAlgorithm}\;
        
        (B3) For each $k$, call $\widehat{V}_{t+1}^{(k)}(\mathcal{H}_{t+1}^{(k)}) = $ \Fv{$p_{t+1}^{(k)}, i_{t+1}^{(k)}, \mathcal{I}_{t+1}^{(k)}$} to estimate the value function of hyper state $\mathcal{H}_{t+1}^{(k)}$ at the next decision epoch $t+1$\;
    
        (B4) \KwRet  $\widehat{V}_{t}(\mathcal{H}_{t}) = \max \left\{R(p_t, i_t, H),~ -c_u + \dfrac{\gamma}{K} \sum_{k=1}^K \widehat{V}_{t+1}^{(k)}(\mathcal{H}_{t+1}^{(k)}) \right\}$\;
    }
}

}
\end{algorithm}

It is known that the online sparse sampling algorithm (which is implemented by Algorithms~\ref{alg:online_control} and \ref{alg:sparse_sampling} for our problem) is guaranteed to compute a near-optimal action at any state of the MDP model \citep{kearns2002sparse}. The theoretical results on the quality of the near-optimal action depend on the number of samples $K$ at each node and the depth of the look-ahead tree. In our implementation of the sparse-sampling algorithm, we do not set a specific value for the depth of the look-ahead tree, but instead continue branching until all the leaves represent a harvest action (i.e., step (A3) in Algorithm~\ref{alg:sparse_sampling} ends up with step (B1) for each $k$). Notice that the tree grows until at most decision epoch $\bar{T}$ at which the fermentation process must be terminated (if not done yet). In our implementation, the number of samples $K$ is the only parameter that controls the computation time and quality of the solution. In literature, it is typical to set the value of $K$ not too large considering the branches grow exponentially; e.g.,  \cite{kearns2002sparse} and  \cite{ross2008model} set $K$ equal to 5 in their implementation of Bayesian sparse sampling. In our numerical study, we set $K = 10$, as it is observed that the results do not change by increasing the value of $K$ further. 

The Bayesian sparse sampling generates computational results for the ``RL with MR'' strategy. The analysis starts with the non-informative prior that $\alpha_{0}^{(p)} = \nu_{0}^{(p)} = \lambda_{0}^{(p)} = \beta_{0}^{(p)} = \alpha_{0}^{(i)} = \nu_{0}^{(i)} = \lambda_{0}^{(i)} = \beta_{0}^{(i)} = 0$. For each case (i.e., $J_0=3, 10, 20$), we performed 100 macro-replications to assess the performance of the proposed reinforcement learning framework. In each macro-replication, we generated $J_0$ number of process data $\mathcal{D}_0$ from the underlying true distribution of the protein and impurity growth rates, and then updated the knowledge states. After that, with the updated knowledge state, we estimated the Q-function for the continue action over the physical state space $(p_t, i_t) \in [1.5, 30]\times [2.0, 50]$ at time $t$, based on Bayesian sparse sampling in Algorithm~\ref{alg:sparse_sampling}. We obtained the mean decision boundary and the corresponding 95\% confidence band based on $100$ macro-replications.

\subsection{Performance Metrics}
\label{subsec: performance metrics}
We adopt the following approach to evaluate the performance of each benchmark strategy presented in Section~\ref{Sec:CaseStudy}. For any policy $\pi$ obtained by the considered benchmark strategies, we evaluate the policy $\pi$ through the expected total reward achieved for the underlying {\it true} process, as defined in Section~\ref{subsec:hybridModeling}, 
\begin{equation*}
\rho^c(\pi) = \mbox{E}_{\tau^c}\left[ \sum_{t=0}^{T} R(\mathcal{S}_t, a_t) \bigg| \mathcal{S}_0, \pi  \right],
\end{equation*}
where the expectation is taken with respect to the stochastic trajectory $\tau^c \equiv (\mathcal{S}_0, a_0, \mathcal{S}_1, a_1, \ldots, \mathcal{S}_{T-1}, a_{T-1}, \mathcal{S}_T)$ with stopping time $T$, following the underlying physical state transition $\mbox{Pr}(\mathcal{S}_{t+1} | \mathcal{S}_t, a_t; \pmb{\theta}^c)$ and decision policy $a_t = \pi_t(\mathcal{S}_t)$. Based on the given policy $\pi$, we can generate trajectory realizations $\tau^c_n \equiv (\mathcal{S}_0^{(n)}, a_0^{(n)}, \mathcal{S}_1^{(n)}, a_1^{(n)}, \ldots, \mathcal{S}_{T_n-1}^{(n)}, a_{T_n-1}^{(n)}, \mathcal{S}_{T_n}^{(n)})$ with $n = 1, 2, \ldots, N$ from the underlying model, and the expected total reward can be estimated by,
\begin{equation*}
\widehat{\rho}^c(\pi) = \dfrac{1}{N} \sum_{n=1}^N \sum_{t=0}^{T_n} R\left( \mathcal{S}_t^{(n)}, \pi_t(\mathcal{S}_t^{(n)}) \right),
\end{equation*}
where $T_n$ denotes the stopping time in the trajectory realization $n$. Further, to assess the process stability, we measure the batch-to-batch variations under a given policy through the standard deviation (SD) of the total reward,
\begin{equation*}
\mbox{SD}^c(\pi) = \mbox{SD}_{\tau^c}\left[ \sum_{t=0}^{T} R(\mathcal{S}_t, a_t) \bigg| \mathcal{S}_0, \pi  \right],
\end{equation*}
which is estimated by sample SD from $N$ replications of simulation experiments, i.e.,
\begin{equation*}
\widehat{\mbox{SD}}^c(\pi) = \sqrt{ \dfrac{1}{N-1} \sum_{n=1}^N \left[ \sum_{t=0}^{T_n} R\left( \mathcal{S}_t^{(n)}, \pi_t(\mathcal{S}_t^{(n)})\right) - \widehat{\rho}^c(\pi) \right]^2}.
\end{equation*}
For each strategy, we run $N = 100$ replications to compare the mean and the standard deviation of the expected total reward.  

\section{Additional Numerical Analysis } \label{sec:additionalExp}

\subsection{Analysis on Costs and Rewards}  
\label{sec:campaign}
We provide a sensitivity analysis with respect to \textit{(i)} the unit reward obtained per protein amount, i.e., $r_h(p,i) = 5p - i$, $r_h(p,i) = 10p - i$ and $r_h(p,i) = 15p - i$; and \textit{(ii)} the failure penalty cost, i.e., $r_f = 400, 880, 1000$. These different reward and cost configurations are defined to capture a wide range of practically-relevant settings. Other parameters remain the same as discussed in Section~\ref{subsec:overview}. The results are shown in Table~\ref{table:compare_sensitivity}. 

\begin{table}[h!] 
\caption{The mean and standard deviation of the total reward achieved by different strategies under various configurations on unit rewards and failure costs.}
\label{table:compare_sensitivity}
\centering
\resizebox{\textwidth}{!}{%
\begin{tabular}{|cc|cc|cc|cc|cc|cc|}
\hline
              &              & \multicolumn{2}{c|}{Fixed Threshold} & \multicolumn{2}{c|}{RL ignoring MR} & \multicolumn{2}{c|}{Myopic Policy} & \multicolumn{2}{c|}{RL with MR} & \multicolumn{2}{c|}{Perfect Info MDP} \\ \hline
Protein Value & Failure Cost & $\widehat{\rho}^c(\pi)$               & $\widehat{\mbox{SD}}^c(\pi)$               & $\widehat{\rho}^c(\pi)$               & $\widehat{\mbox{SD}}^c(\pi)$               & $\widehat{\rho}^c(\pi)$               & $\widehat{\mbox{SD}}^c(\pi)$               & $\widehat{\rho}^c(\pi)$               & $\widehat{\mbox{SD}}^c(\pi)$               & $\widehat{\rho}^c(\pi)$               & $\widehat{\mbox{SD}}^c(\pi)$               \\ \hline
5             & 400          & 29.68             & 142.86           & 66.22            & 58.89            & 67.84            & 59.08           & 71.28          & 33.90          & 72.20             & 34.01             \\
10            & 400          & 140.60            & 184.12           & 155.00           & 135.88           & 175.80           & 122.88          & 181.34         & 110.70         & 185.71            & 87.15             \\
15            & 400          & 251.51            & 226.89           & 298.34           & 140.28           & 285.80           & 173.85          & 299.47         & 123.00         & 314.25            & 105.63            \\ \hline
5             & 880          & -13.52            & 279.42           & 61.42            & 102.12           & 60.88            & 102.09          & 71.28          & 33.90          & 71.80             & 33.87             \\
10            & 880          & 97.40             & 318.19           & 143.47           & 196.39           & 168.21           & 166.31          & 171.63         & 126.68         & 177.23            & 125.86            \\
15            & 880          & 208.31            & 358.48           & 268.61           & 227.41           & 273.80           & 228.98          & 281.01         & 194.74         & 295.97            & 152.93            \\ \hline
5             & 1000         & -24.32            & 313.76           & 60.22            & 113.53           & 59.49            & 113.49          & 71.28          & 33.90          & 71.33             & 34.23             \\
10            & 1000         & 86.60             & 352.19           & 139.87           & 215.71           & 165.18           & 181.73          & 170.43         & 137.03         & 176.03            & 136.27            \\
15            & 1000         & 197.51            & 392.07           & 246.12           & 246.75           & 270.20           & 247.66          & 271.32         & 209.92         & 294.77            & 162.50            \\ \hline
\end{tabular} }
\end{table}

For each different revenue and cost configuration, Table~\ref{table:compare_sensitivity} presents the mean $\widehat{\rho}^c(\pi)$ and standard deviation $\widehat{\mbox{SD}}^c(\pi)$ of total reward realized by $100$ batches. Each column corresponds to a harvesting strategy. We considered the size of historical data $J_0 = 10$ for Myopic, RL-ignoring-MR and RL-with-MR strategies. For each strategy, we observe from Table~\ref{table:compare_sensitivity} that the average and standard deviation of the total reward increase as the unit reward of protein increases; and the average reward decreases but the standard deviation increases as the failure cost increases. This trend matches our intuition. For the cases with lower protein rewards or/and lower failure costs, we observe from Table~\ref{table:compare_sensitivity} that the performance of the proposed RL-with-MR strategy (under $J_0=10$) is closer to that of the perfect information setting PI-MDP. For practitioners, this implies that the potential impact of model risk is higher for high-revenue (and/or cost) drugs in comparison to their low-revenue (and/or cost) counterparts.

\subsection{Analysis on the Frequency of Decision Epochs } \label{sec:SensitivityN}

\begin{figure}[h!]
	\centering
	\includegraphics[scale=0.45]{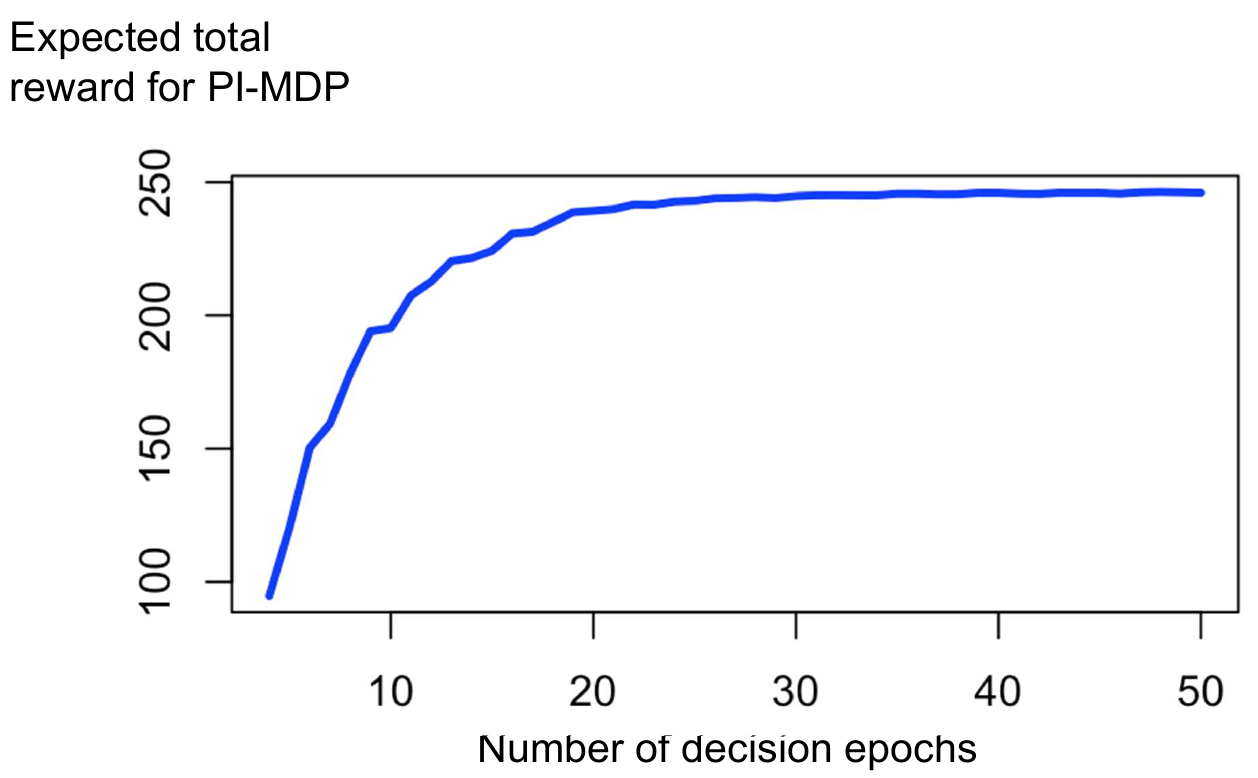}
	\caption{The expected total reward under the PI-MDP strategy as a function of the number of decisions epochs (varied from 4 to 50).}
	\label{fig:discretization}
	\vspace{-0.in}
\end{figure}

Recall that the case study presented in Section~\ref{Sec:CaseStudy} assumed that harvesting decisions can be made every 6 hours (i.e., $8$ decision epochs during a maximum of $48$ hours of fermentation). In this section, we evaluate the potential benefits of adopting more frequent decision epochs. For this purpose, we consider the parameter setting described in Section~\ref{Sec:CaseStudy} and focus on the PI-MDP strategy. Figure~\ref{fig:discretization} shows how the expected total reward changes when the frequency of decisions increases from $4$ to 50 decision epochs during a finite planning horizon of $48$ hours (i.e., the opportunity to make the harvesting decision increases from every 12 hours to every 0.96 hours). 

We observe from Figure~\ref{fig:discretization} that the expected total reward under the PI-MDP strategy converges around $20$ decision epochs (i.e., when there is an opportunity to make the harvesting decision every 2.4 hours), approximating the optimal reward to be collected under a continuous-time harvesting policy. The analysis shows a diminishing return structure in the frequency of decision epochs. That is, the benefit from more frequent harvesting opportunities is a decreasing function of the frequency. 

In industry, the frequency of decisions is driven by several constraints related to production planning and resource availability. To be specific, biomanufacturing operations can involve many 
interdependent production steps. Therefore, manufacturers often adopt a predetermined production ``rhythm" to orchestrate these interdependent activities. These rhythms are predetermined, fixed, and set at the factory level \citep{martagan2022}. In addition, not all biomanufacturing facilities may be equipped with the necessary infrastructure (e.g., sensors, information technology) to collect and process data in frequent time intervals. 
Combined with the limited availability of operators and other resources, these complex factory dynamics affect the frequency of decision epochs for feasible implementation in daily practice. Due to these reasons and to be consistent with practice, our case study assumed the possibility of harvesting decisions every six hours, but note that 
our discrete-time model is applicable to any finite amount of time between decision epochs.

\section{Proofs}
\noindent \textbf{Proof of Proposition~\ref{theo:var_decomp}}
\label{sec:pf_var_decomp}

(i) Notice that $\phi_j$, $j=1,\ldots,J_t$, are i.i.d. samples from underlying true protein growth rate distribution $\mathcal{N}(\mu^{(p)}, \sigma^{(p)2})$. According to the properties of Gaussian sample mean and sample variance, we have $\bar{\phi} \sim \mathcal{N}(\mu^{(p)}, \dfrac{\sigma^{(p)2}}{J_t})$, $\dfrac{\sum_{j=1}^{J_t}(\phi_j - \bar{\phi})^2}{\sigma^{(p)2}} \sim \chi^2(J_t-1)$, and they are independent of each other. 
Thus, $\widetilde{\sigma}_{t}^{(p)2} = \dfrac{J_t+1}{(J_t-2)J_t}\sum_{j=1}^{J_t}(\phi^{(j)} - \bar{\phi})^2 \sim \dfrac{(J_t+1)\sigma^{(p)2}}{(J_t-2)J_t} \chi^2(J_t-1)$. Based on the mean and variance of Chi-square distribution, we have
\begin{equation}
\mbox{E}\left[\widetilde{\sigma}_{t}^{(p)2}\right] = \dfrac{(J_t^2-1)\sigma^{(p)2}}{(J^2_t-2J_t)} = \sigma^{(p)2} + \dfrac{(2J_t-1)\sigma^{(p)2}}{(J_t^2-2J_t)},~~
\mbox{Var}\left[\widetilde{\sigma}_{t}^{(p)2}\right] = \dfrac{2(J_t^3+J_t^2-J_t-1)\sigma^{(p)4}}{J_t^4 - 4J_t^3 + 4J_t^2}. \nonumber
\end{equation}

(ii) Recall that the predictive growth rate $\widetilde{\Phi}_{t}$ is a compound random variable, i.e., $\widetilde{\Phi}_{t}\sim \mathcal{N}(\mu^{(p)}, \sigma^{(p)2})$ where $\left(\mu^{(p)}, \sigma^{(p)2}\right) \sim \mathcal{N}(\alpha_{t}^{(p)}, \sigma^{(p)2}/\nu_{t}^{(p)})\cdot {\mbox{Inv}\Gamma}(\lambda_{t}^{(p)}, \beta_{t}^{(p)})$ at decision epoch $t$. For brevity, let $\boldsymbol{\theta}^{(p)}$ denote the random variables $\left(\mu^{(p)}, \sigma^{(p)2}\right)$. The variance of the compound random variable $\widetilde{\Phi}_{t}$ can be decomposed as follows:
%
\begin{eqnarray}
\lefteqn{ \widetilde{\sigma}^{(p)2}_t \triangleq \mbox{Var}\left[\widetilde{\Phi}_{t}  \right] = \mbox{E}_{\boldsymbol{\theta}^{(p)}}\left[ \mbox{Var}\left(\widetilde{\Phi}_{t} \Big| \boldsymbol{\theta}^{(p)} \right) \right] + \mbox{Var}_{\boldsymbol{\theta}^{(p)}}\left[ \mbox{E}\left(\widetilde{\Phi}_{t} \Big| \boldsymbol{\theta}^{(p)} \right) \right] } 
\label{eq:prop1-1}  \\
& = &\mbox{E}_{\boldsymbol{\theta}^{(p)}}\left[ \sigma^{(p)2} \right] + \mbox{Var}_{\boldsymbol{\theta}^{(p)}}\left[ \mu^{(p)} \right] \nonumber\\
& = & \mbox{E}_{\boldsymbol{\theta}^{(p)}}\left[ \sigma^{(p)2} \right] +  
\mbox{E}_{\boldsymbol{\theta}^{(p)}}\left[ \mbox{Var}\left(\mu^{(p)} \Big| \sigma^{(p)2} \right) \right] +
\mbox{Var}_{\boldsymbol{\theta}^{(p)}}\left[ \mbox{E}\left(\mu^{(p)2} \Big| \sigma^{(p)2} \right)  \right]
\label{eq:prop1-2} \\
& = & \mbox{E}_{\boldsymbol{\theta}^{(p)}}\left[ \sigma^{(p)2} \right] +  
\mbox{E}_{\boldsymbol{\theta}^{(p)}}\left[  \sigma^{(p)2}/\nu_{t}^{(p)} \right] +
\mbox{Var}_{\boldsymbol{\theta}^{(p)}}\left[ {\alpha}^{(p)}_t \right]
\nonumber \\
& = & \dfrac{\beta_{t}^{(p)}}{\lambda_{t}^{(p)} - 1} +  \dfrac{\beta_{t}^{(p)}}{(\lambda_{t}^{(p)} - 1)\nu_{t}^{(p)}}. \label{eq.vardecomp_p}
\end{eqnarray}
where the equations in \eqref{eq:prop1-1} and \eqref{eq:prop1-2} hold by law of total variance. We note that $\mbox{Var}_{\boldsymbol{\theta}^{(p)}}\left[ {\alpha}^{(p)}_t \right]$ is zero because $\alpha^{(p)}_t$ is a constant. The first term in the right-hand side of \eqref{eq.vardecomp_p} (i.e., $\hat{\sigma}_{t}^{(p)2} \triangleq  \dfrac{\beta_{t}^{(p)}}{\lambda_{t}^{(p)} - 1}$) corresponds to $\mbox{E}_{\boldsymbol{\theta}^{(p)}}\left[ \sigma^{(p)2} \right]$, measuring the expected variability of the protein growth rate due to the inherent stochasticity of the fermentation. The second term (i.e., $\check{\sigma}_{t}^{(p)2} \triangleq \dfrac{\beta_{t}^{(p)}}{(\lambda_{t}^{(p)} - 1)\nu_{t}^{(p)}}$) corresponds to $\mbox{E}_{\boldsymbol{\theta}^{(p)}}\left[ \mbox{Var}\left(\mu^{(p)} \Big| \sigma^{(p)2} \right) \right]$ and it represents the expected variability of the protein growth rate due to model risk.


\noindent \textbf{Proof of Proposition~\ref{theo:physical}}
\label{sec:pf_th1}

Notice that, at time $\bar{T}$, we have  
\begin{align}
	V_{\bar{T}}(p_{\bar{T}}, i_{\bar{T}}, \mathcal{I}_{\bar{T}})
	= 
	\begin{cases}
	r_h(p_{\bar{T}},i_{\bar{T}}),
	~~ &\mbox{~if~} i_{\bar{T}} < \bar{I} \\
	-r_f,
	~~ &\mbox{~if~}  i_{\bar{T}} = \bar{I},
	\end{cases} \nonumber
\end{align}
which is non-increasing in $i_{\bar{T}}$ and non-decreasing in $p_{\bar{T}}$. Based on the value function given in Section~\ref{sec:RL-MR}, we can prove Proposition~\ref{theo:physical} through backward induction. 

As the induction hypothesis, assume that $V_{t+1}(p_{t+1}, i_{t+1}, \mathcal{I}_{t+1})$ is non-increasing in $i_{t+1}$ and non-decreasing in $p_{t+1}$ for a given $t \in \{0,1,\ldots,\bar{T} -1\}$. Recall from \eqref{eq: valuefunction} that the value function at decision epoch $t$ is given by 
\begin{equation*}
	V_t(p_{t}, i_{t}, \mathcal{I}_{t})= \begin{cases} 
	{\max}\left\{ r_h(p_t, i_t), -c_u + \gamma \mbox{E}\left[
	V_{t+1}(p_{t+1}, i_{t+1}, \mathcal{I}_{t+1})\right]\right\} & {\rm if \ } p_t < \bar{P} {\rm \ and \ } i_t < \bar{I} \\
	r_h(p_t, i_t) & {\rm if \ } p_t = \bar{P} {\rm \ and \ } i_t  < \bar{I} \\
	- r_f & {\rm if \ } i_t  = \bar{I},
	\end{cases}
\end{equation*}
where $\mbox{E}\left[V_{t+1}(p_{t+1}, i_{t+1}, \mathcal{I}_{t+1})\right] = \mbox{E}[V_{t+1}(p_te^{\widetilde{\Phi}_t}, i_te^{\widetilde{\Psi}_t}, \mathcal{I}_{t+1})]$. For any $i_t^{+} \ge i_t$, we know that 
\begin{align*}
	&   \mbox{E}[V_{t+1}(p_te^{\widetilde{\Phi}_t}, i_te^{\widetilde{\Psi}_t}, \mathcal{I}_{t+1})] = -r_f \int_{\ln{\bar{I}} - \ln{i_t}}^{\infty}f_{t}^{(i)}(\psi_t)d\psi_t \nonumber \\
&+ 
\int_{-\infty}^{\ln{\bar{I}} - \ln{i_t}} \int_{-\infty}^{\ln{\bar{P}} - \ln{p_t}} V_{t+1}(p_te^{\phi_t}, i_te^{\psi_t}, \mathcal{I}_{t+1})f_{t}^{(p)}(\phi_t)f_{t}^{(i)}(\psi_t)d\phi_t d\psi_t \nonumber \\
&+ 
\int_{-\infty}^{\ln{\bar{I}} - \ln{i_t}} \int_{\ln{\bar{P}} - \ln{p_t}}^{\infty} r_h(\bar{P},i_te^{\psi_t})
f_{t}^{(p)}(\phi_t)f_{t}^{(i)}(\psi_t)d\phi_t d\psi_t \nonumber \\
=& - r_f \int_{\ln{\bar{I}} - \ln{i_t}}^{\infty}f_{t}^{(i)}(\psi_t)d\psi_t \nonumber \\
&
+ 
\int_{-\infty}^{\ln{\bar{I}} - \ln{i_t^+}} \int_{-\infty}^{\ln{\bar{P}} - \ln{p_t}} V_{t+1}(p_te^{\phi_t}, i_te^{\psi_t}, \mathcal{I}_{t+1})f_{t}^{(p)}(\phi_t)f_{t}^{(i)}(\psi_t)d\phi_t d\psi_t \nonumber \\
&
+
\int_{-\infty}^{\ln{\bar{I}} - \ln{i_t^+}} \int_{\ln{\bar{P}} - \ln{p_t}}^{\infty} r_h(\bar{P},i_te^{\psi_t})
f_{t}^{(p)}(\phi_t)f_{t}^{(i)}(\psi_t)d\phi_t d\psi_t \nonumber \\
&+
\int_{\ln{\bar{I}} - \ln{i_t^+}}^{\ln{\bar{I}} - \ln{i_t}} \int_{-\infty}^{\ln{\bar{P}} - \ln{p_t}} V_{t+1}(p_te^{\phi_t}, i_te^{\psi_t}, \mathcal{I}_{t+1})f_{t}^{(p)}(\phi_t)f_{t}^{(i)}(\psi_t)d\phi_t d\psi_t \nonumber \\
&+
\int_{\ln{\bar{I}} - \ln{i_t^+}}^{\ln{\bar{I}} - \ln{i_t}} \int_{\ln{\bar{P}} - \ln{p_t}}^{\infty} r_h(\bar{P},i_te^{\psi_t})
f_{t}^{(p)}(\phi_t)f_{t}^{(i)}(\psi_t)d\phi_t d\psi_t \nonumber \\
\ge & - r_f \int_{\ln{\bar{I}} - \ln{i_t}}^{\infty}f_{t}^{(i)}(\psi_t)d\psi_t \nonumber \\
&
+ 
\int_{-\infty}^{\ln{\bar{I}} - \ln{i_t^+}} \int_{-\infty}^{\ln{\bar{P}} - \ln{p_t}} V_{t+1}(p_te^{\phi_t}, i_te^{\psi_t}, \mathcal{I}_{t+1})f_{t}^{(p)}(\phi_t)f_{t}^{(i)}(\psi_t)d\phi_t d\psi_t \nonumber \\
&
+
\int_{-\infty}^{\ln{\bar{I}} - \ln{i_t^+}} \int_{\ln{\bar{P}} - \ln{p_t}}^{\infty} r_h(\bar{P},i_te^{\psi_t})
f_{t}^{(p)}(\phi_t)f_{t}^{(i)}(\psi_t)d\phi_t d\psi_t \nonumber \\
&-r_f
\int_{\ln{\bar{I}} - \ln{i_t^+}}^{\ln{\bar{I}} - \ln{i_t}} \int_{-\infty}^{\ln{\bar{P}} - \ln{p_t}}  f_{t}^{(p)}(\phi_t)f_{t}^{(i)}(\psi_t)d\phi_t d\psi_t \nonumber \\
&-r_f
\int_{\ln{\bar{I}} - \ln{i_t^+}}^{\ln{\bar{I}} - \ln{i_t}} \int_{\ln{\bar{P}} - \ln{p_t}}^{\infty} 
f_{t}^{(p)}(\phi_t)f_{t}^{(i)}(\psi_t)d\phi_t d\psi_t \nonumber \\
\ge& - r_f \int_{\ln{\bar{I}} - \ln{i_t^+}}^{\infty}f_{t}^{(i)}(\psi_t)d\psi_t + 
\int_{-\infty}^{\ln{\bar{I}} - \ln{i_t^+}} \int_{-\infty}^{\ln{\bar{P}} - \ln{p_t}} V_{t+1}(p_te^{\phi_t}, i_t^+ e^{\psi_t}, \mathcal{I}_{t+1})f_{t}^{(p)}(\phi_t)f_{t}^{(i)}(\psi_t)d\phi_t d\psi_t \nonumber \\
&
+
\int_{-\infty}^{\ln{\bar{I}} - \ln{i_t^+}} \int_{\ln{\bar{P}} - \ln{p_t}}^{\infty} r_h(\bar{P},i_t^+ e^{\psi_t})
f_{t}^{(p)}(\phi_t)f_{t}^{(i)}(\psi_t)d\phi_t d\psi_t \nonumber \\
=& \ \mbox{E}[V_{t+1}(p_te^{\widetilde{\Phi}_t}, i_t^{+}e^{\widetilde{\Psi}_t}, \mathcal{I}_{t+1})]
\end{align*}
where the first inequality holds because  $V_{t+1}(p_t e^{\phi_t}, i_t e^{\psi_t}, \mathcal{I}_{t+1}) \ge -r_f$ and $r_h(\bar{P},i_t e^{\psi_t}) \ge -r_f$ (this follows from the assumption $r_f>c_2 \bar{I}$), and the second inequality holds because $r_h(\bar{P},i_t e^{\psi_t}) \geq r_h(\bar{P},i_t^+ e^{\psi_t})$ and the induction hypothesis together with $i_t^+ e^{\psi_t} \ge i_t e^{\psi_t}$ imply that 
$\mbox{E}[V_{t+1}(p_t e^{\widetilde{\Phi}_t}, i_t e^{\widetilde{\Psi}_t}, \mathcal{I}_{t+1})]  \ge \mbox{E}[V_{t+1}(p_t e^{\widetilde{\Phi}_t}, i_t^{+}e^{\widetilde{\Psi}_t}, \mathcal{I}_{t+1})]$. 
Consequently, $V_{t}(p_t, i_t, \mathcal{I}_t)$ is also non-increasing in $i_t$.

On the other hand, for any $p_t^{+} \ge p_t$, we know that 
\begin{align}
\mbox{E}&[V_{t+1}(p_te^{\widetilde{\Phi}_t}, i_te^{\widetilde{\Psi}_t}, \mathcal{I}_{t+1})] = -r_f \int_{\ln{\bar{I}} - \ln{i_t}}^{\infty}f_{t}^{(i)}(\psi_t)d\psi_t \nonumber \\
&+ \int_{-\infty}^{\ln{\bar{I}} - \ln{i_t}} \int_{-\infty}^{\ln{\bar{P}-\ln_{p_t}}} V_{t+1}(p_te^{\phi_t}, i_te^{\psi_t}, \mathcal{I}_{t+1})f_{t}^{(p)}(\phi_t)f_{t}^{(i)}(\psi_t)d\phi_t d\psi_t \nonumber \\
&+ \int_{-\infty}^{\ln{\bar{I}} - \ln{i_t}} \int_{\ln{\bar{P}-\ln_{p_t}}}^{\infty} r_h(\bar{P}, i_te^{\psi_t})f_{t}^{(p)}(\phi_t)f_{t}^{(i)}(\psi_t)d\phi_t d\psi_t \nonumber \\
& = 
-r_f \int_{\ln{\bar{I}} - \ln{i_t}}^{\infty}f_{t}^{(i)}(\psi_t)d\psi_t \nonumber \\
&+ \int_{-\infty}^{\ln{\bar{I}} - \ln{i_t}} \int_{-\infty}^{\ln{\bar{P}-\ln_{p_t}^+}} V_{t+1}(p_te^{\phi_t}, i_te^{\psi_t}, \mathcal{I}_{t+1})f_{t}^{(p)}(\phi_t)f_{t}^{(i)}(\psi_t)d\phi_t d\psi_t \nonumber \\
&+ \int_{-\infty}^{\ln{\bar{I}} - \ln{i_t}} \int_{\ln{\bar{P}-\ln_{p_t}^+}}^{\ln{\bar{P}-\ln_{p_t}}} V_{t+1}(p_te^{\phi_t}, i_te^{\psi_t}, \mathcal{I}_{t+1})f_{t}^{(p)}(\phi_t)f_{t}^{(i)}(\psi_t)d\phi_t d\psi_t \nonumber \\
&+ \int_{-\infty}^{\ln{\bar{I}} - \ln{i_t}} \int_{\ln{\bar{P}}-\ln_{p_t}^+}^{\infty} r_h(\bar{P}, i_te^{\psi_t})f_{t}^{(p)}(\phi_t)f_{t}^{(i)}(\psi_t)d\phi_t d\psi_t \nonumber \\
&- \int_{-\infty}^{\ln{\bar{I}} - \ln{i_t}} \int_{\ln{\bar{P}}-\ln_{p_t}^+}^{\ln{\bar{P}}-\ln_{p_t}} r_h(\bar{P}, i_te^{\psi_t})f_{t}^{(p)}(\phi_t)f_{t}^{(i)}(\psi_t)d\phi_t d\psi_t \nonumber \\
& \leq 
-r_f \int_{\ln{\bar{I}} - \ln{i_t}}^{\infty}f_{t}^{(i)}(\psi_t)d\psi_t \nonumber \\
&+ \int_{-\infty}^{\ln{\bar{I}} - \ln{i_t}} \int_{-\infty}^{\ln{\bar{P}-\ln_{p_t}^+}} V_{t+1}(p_te^{\phi_t}, i_te^{\psi_t}, \mathcal{I}_{t+1})f_{t}^{(p)}(\phi_t)f_{t}^{(i)}(\psi_t)d\phi_t d\psi_t \nonumber \\
&+ \int_{-\infty}^{\ln{\bar{I}} - \ln{i_t}} \int_{\ln{\bar{P}}-\ln_{p_t}^+}^{\infty} r_h(\bar{P}, i_te^{\psi_t})f_{t}^{(p)}(\phi_t)f_{t}^{(i)}(\psi_t)d\phi_t d\psi_t \nonumber \\
& \leq 
-r_f \int_{\ln{\bar{I}} - \ln{i_t}}^{\infty}f_{t}^{(i)}(\psi_t)d\psi_t \nonumber \\
&+ \int_{-\infty}^{\ln{\bar{I}} - \ln{i_t}} \int_{-\infty}^{\ln{\bar{P}-\ln_{p_t}^+}} V_{t+1}(p_t^+ e^{\phi_t}, i_te^{\psi_t}, \mathcal{I}_{t+1})f_{t}^{(p)}(\phi_t)f_{t}^{(i)}(\psi_t)d\phi_t d\psi_t \nonumber \\
&+ \int_{-\infty}^{\ln{\bar{I}} - \ln{i_t}} \int_{\ln{\bar{P}}-\ln_{p_t}^+}^{\infty} r_h(\bar{P}, i_te^{\psi_t})f_{t}^{(p)}(\phi_t)f_{t}^{(i)}(\psi_t)d\phi_t d\psi_t \nonumber \\
=& \mbox{E}[V_{t+1}(p_t^{+}e^{\widetilde{\Phi}_t}, i_te^{\widetilde{\Psi}_t}, \mathcal{I}_{t+1})]
\end{align}
where the first inequality holds because $r_h(\bar{P}, i_te^{\psi_t}) \geq V_{t+1}(p_te^{\phi_t}, i_te^{\psi_t}, \mathcal{I}_{t+1})$ (this holds due to the fact that the value function at a fixed impurity level can never exceed the {\it maximum} harvesting reward achievable at that impurity level), and the  second inequality holds because of the induction hypothesis and $p_t e^{\phi_t} \le p_t^{+} e^{\phi_t}$. Consequently, we have $V_{t}(p_t, i_t, \mathcal{I}_t)$ is also non-decreasing in $p_t$.


\noindent \textbf{Proof of Proposition~\ref{theo:policy_i}}
\label{sec:pf_th3}

According to Proposition~\ref{theo:policy_i}, if $\pi^{\star}(p_t, i_t^-, \mathcal{I}_t) = H$, then $\pi^{\star}(p_t, i_t^+, \mathcal{I}_t) = H$ for any $i^+>i^-$. 
Assume that $\pi^{\star}(p_t, i_t^-, \mathcal{I}_t) = H$ but $\pi^{\star}(p_t, i_t^+, \mathcal{I}_t) = C$ as the contradiction hypothesis. Then, it holds that
\begin{align*}
	r_h(p_t, i_t^-) &\ge -c_u + \gamma \mbox{E}\left[V_{t+1}(p_te^{\widetilde{\Phi}_t}, i_t^- e^{\widetilde{\Psi}_t}, \mathcal{I}_{t+1})  \right] \\
	r_h(p_t, i_t^+) &\le -c_u + \gamma \mbox{E}\left[V_{t+1}(p_te^{\widetilde{\Phi}_t}, i_t^+ e^{\widetilde{\Psi}_t}, \mathcal{I}_{t+1})  \right].
\end{align*}
Thus, we have
\begin{align}
	r_h(p_t, i_t^-) & - r_h(p_t, i_t^+) \ge \gamma \mbox{E}\left[ V_{t+1}(p_te^{\widetilde{\Phi}_t}, i_t^- e^{\widetilde{\Psi}_t}, \mathcal{I}_{t+1})  \right] - \gamma \mbox{E}\left[ V_{t+1}(p_te^{\widetilde{\Phi}_t}, i_t^+ e^{\widetilde{\Psi}_t}, \mathcal{I}_{t+1})  \right] \nonumber \\
	=& -\gamma r_f \int_{\ln{\bar{I}} - \ln{i_t^-}}^{\infty}f_{t}^{(i)}(\psi_t)d\psi_t + \gamma r_f \int_{\ln{\bar{I}} - \ln{i_t^+}}^{\infty}f_{t}^{(i)}(\psi_t)d\psi_t \nonumber \\
    &+ 
    \gamma \int_{-\infty}^{\ln{\bar{I}} - \ln{i_t^-}} \int_{-\infty}^{\ln{\bar{P}} - \ln{p_t}} V_{t+1}(p_te^{\phi_t}, i_t^- e^{\psi_t}, \mathcal{I}_{t+1})f_{t}^{(p)}(\phi_t)f_{t}^{(i)}(\psi_t)d\phi_t d\psi_t \nonumber \\ 
    &+ 
    \gamma \int_{-\infty}^{\ln{\bar{I}} - \ln{i_t^-}} \int_{\ln{\bar{P}} - \ln{p_t}}^{\infty} r_h(\bar{P}, i_t^- e^{\psi_t}, \mathcal{I}_{t+1})f_{t}^{(p)}(\phi_t)f_{t}^{(i)}(\psi_t)d\phi_t d\psi_t \nonumber \\
    &- \gamma \int_{-\infty}^{\ln{\bar{I}} - \ln{i_t^+}} \int_{-\infty}^{\ln{\bar{P}} - \ln{p_t}} V_{t+1}(p_te^{\phi_t}, i_t^+ e^{\psi_t}, \mathcal{I}_{t+1})  f_{t}^{(p)}(\phi_t)f_{t}^{(i)}(\psi_t)d\phi_t d\psi_t \nonumber\\
    &- \gamma \int_{-\infty}^{\ln{\bar{I}} - \ln{i_t^+}} \int_{\ln{\bar{P}} - \ln{p_t}}^{\infty} r_h(\bar{P}, i_t^+ e^{\psi_t}, \mathcal{I}_{t+1})  f_{t}^{(p)}(\phi_t)f_{t}^{(i)}(\psi_t)d\phi_t d\psi_t \nonumber\\
	\ge& \gamma r_f \left[ \int_{-\infty}^{\ln{\bar{I}} - \ln{i_t^-}}f_{t}^{(i)}(\psi_t)d\psi_t - \int_{-\infty}^{\ln{\bar{I}} - \ln{i_t^+}}f_{t}^{(i)}(\psi_t)d\psi_t \right] \label{ineq.theo2.2}  \\
    &+ \gamma \int_{-\infty}^{\ln{\bar{I}} - \ln{i_t^-}} \int_{-\infty}^{\infty} r_h(0, i_t^- e^{\psi_t})f_{t}^{(p)}(\phi_t)f_{t}^{(i)}(\psi_t)d\phi_t d\psi_t \nonumber \\ 
    &- \gamma \int_{-\infty}^{\ln{\bar{I}} - \ln{i_t^+}} \int_{-\infty}^{\infty} r_h(\bar{P}, i_t^- e^{\psi_t})  f_{t}^{(p)}(\phi_t)f_{t}^{(i)}(\psi_t)d\phi_t d\psi_t  \nonumber  \\
    \ge& \gamma r_f \left[ \int_{-\infty}^{\ln{\bar{I}} - \ln{i_t^-}}f_{t}^{(i)}(\psi_t)d\psi_t - \int_{-\infty}^{\ln{\bar{I}} - \ln{i_t^+}}f_{t}^{(i)}(\psi_t)d\psi_t \right] \label{ineq.theo2.3}  \\
    &+ \gamma \int_{-\infty}^{\ln{\bar{I}} - \ln{i_t^-}} r_h(0, i_t^- e^{\psi_t})f_{t}^{(i)}(\psi_t) d\psi_t - \gamma \int_{-\infty}^{\ln{\bar{I}} - \ln{i_t^-}} r_h(\bar{P}, i_t^- e^{\psi_t})  f_{t}^{(i)}(\psi_t) d\psi_t \nonumber  \\
	= & \gamma r_f  \int_{\ln{\bar{I}} - \ln{i_t^+}}^{\ln{\bar{I}} - \ln{i_t^-}}f_{t}^{(i)}(\psi_t)d\psi_t - \gamma c_1\bar{P} \int_{-\infty}^{\ln{\bar{I}} - \ln{i_t^-}} f_{t}^{(i)}(\psi_t) d\psi_t  \nonumber
\end{align}
which contradicts the condition, so that we have $\pi^{\star}(p_t, i_t^+) = H$. Notice that, inequality~\eqref{ineq.theo2.2} holds because:  (i) $V_{t+1}(0, i_t^- e^{\psi_t}, \mathcal{I}_{t+1}) \ge r_h(0, i_t^- e^{\psi_t})$ by the definition of the value function (i.e., at any decision epoch, the value function at a particular state can never be less than the harvesting reward at that state) and $V_{t+1}(p_t e^{\phi_t}, i_t^- e^{\psi_t}, \mathcal{I}_{t+1}) \ge V_{t+1}(0, i_t^- e^{\psi_t}, \mathcal{I}_{t+1})$ by the non-decreasing behavior of the value function in $p_t$ for any protein growth-rate realization $\phi_t$, and (ii) $V_{t+1}(p_te^{\phi_t}, i_t^+ e^{\psi_t}, \mathcal{I}_{t+1}) \le r_h(\bar{P}, i_t^+ e^{\psi_t}) \le r_h(\bar{P}, i_t^- e^{\psi_t})$ 
as the reward function $r_h$ is non-increasing in $i_t$. Finally, the inequality~\eqref{ineq.theo2.3} holds because $\ln{\bar{I}} - \ln{i_t^+} \le \ln{\bar{I}} - \ln{i_t^-}$.

\noindent \textbf{Proof of Proposition~\ref{prop: myopic_under_PI}}
\label{sec:pf_Sec4theo1}

We prove that the optimality of the myopic policy under perfect information by using backward induction. To start with, notice that, at time $\bar{T} - 1$, the optimal action is to harvest if and only if $(p_{\bar{T} - 1}, i_{\bar{T} - 1}) \in A$. Suppose the same applies at any time $t+1$ less than $\bar{T}-1$, i.e.,  the optimal action is to harvest if and only if $(p_{t+1}, i_{t+1}) \in A$. We want to show the optimal action at time $t$ is to harvest if and only if $(p_t, i_t) \in A$. This would complete the proof. To be able to show this result, we first postulate that the following two results hold:
\begin{itemize}
    \item[(i)]  if $(p, i)\in A$, then $(p^+, i)\in A$ for any $p^+ > p$; and 
    \item[(ii)]  if $(p, i)\in A$, then $(p, i^+)\in A$ for any $i^+ > i$. 
\end{itemize}
We will later prove that the results (i) and (ii) both hold under the conditions given in Proposition~\ref{prop: myopic_under_PI}.

We start our analysis by noting that, for $(p_t, i_t) \in A$, it holds that 
\begin{eqnarray}
V_{t}(p_t,i_t) & = &  \max\{ r_h(p_t, i_t), -c_u +  \gamma  V_{t+1}(p_{t+1}, i_{t+1}) \} \nonumber \\
& = &  \max\{ r_h(p_t, i_t), -c_u +  \gamma  \mbox{E}[R(p^\prime, i^\prime; H) | p_t, i_t] \} \label{eq:proofThm4-1} \\
& = &  r_h(p_t, i_t), \label{eq:proofThm4-2}
\end{eqnarray}
where \eqref{eq:proofThm4-1} follows from the fact that, first, under positive growth assumption, we have $p_{t+1} \ge p_t$ and $i_{t+1} \ge i_t$, we have $(p_{t+1}, i_t) \in A$ by applying condition (i), and we have $(p_{t+1}, i_{t+1}) \in A$ by subsequently applying condition (ii), and then, by the induction hypothesis that asserts the optimal action is to harvest when $(p_{t+1}, i_{t+1}) \in A$. Furthermore, \eqref{eq:proofThm4-2} holds because we know $(p_t, i_t) \in A$.

On the other hand, for $(p_t, i_t) \notin A$, the policy that continues the fermentation process one more time period and then takes the harvest decision has an expected reward of 
\[
-c_u +  \gamma \,  \mbox{E}[R(p^\prime, i^\prime; H) | p_t, i_t],
\]
which is strictly greater than $r_h(p_t, i_t)$ (because $(p_t,i_t) \notin A)$. Thus, we establish that $V_{t}(p_t,i_t)=r_h(p_t, i_t)$ for $(p_t,i_t) \in A$, and $V_{t}(p_t,i_t)>r_h(p_t, i_t)$ for $(p_t,i_t) \notin A$, and the result follows.

Now what remains to show is that the results in (i) and (ii) indeed hold. To start with, suppose the result (i) does not hold, i.e.,
\begin{align*}
r_h(p, i) \ge -c_u +  \gamma \, \mbox{E}[R(p^\prime, i^\prime; H) | p, i], \\
r_h(p^+, i) \le -c_u +  \gamma  \, \mbox{E}[R(p^\prime, i^\prime; H) | p^+, i],
\end{align*}
implying that
\begin{equation}
\dfrac{1}{\gamma}\left[r_h(p^+, i) - r_h(p, i)\right] \le \mbox{E}[R(p^\prime, i^\prime; H) | p^+, i] - \mbox{E}[R(p^\prime, i^\prime; H) | p, i] \label{eq.pf_lemma_1}.
\end{equation}
The left-hand side (LHS) of \eqref{eq.pf_lemma_1} is given by
\begin{equation*}
{\rm LHS} = \dfrac{c_1}{\gamma}(p^+ - p).
\end{equation*}
On the other hand, it holds for the right-hand side (RHS) of \eqref{eq.pf_lemma_1} that
\begin{align*}
&{\rm RHS} = c_1 \mbox{Pr}(i^\prime < \bar{I} | i) \left( \mbox{E}[p^\prime | p^+] - \mbox{E}[p^\prime | p] \right) \\
=& c_1 \Phi\left( \dfrac{\ln{\bar{I}} - \ln{i} - \mu_c^{(i)}}{\sigma_c^{(i)}} \right) \Bigg\{ p^+ e^{\mu_c^{(p)} + \sigma_c^{(p)2}/2} \Phi\left( \dfrac{\ln{\bar{P}} - \ln{p^+} - \mu_c^{(p)} - \sigma_c^{(p)2} }{\sigma_c^{(p)}} \right) + \bar{P} \Phi\left( \dfrac{\ln{\bar{P}} - \ln{p^+} - \mu_c^{(p)} }{\sigma_c^{(p)}} \right) \\
&- p e^{\mu_c^{(p)} + \sigma_c^{(p)2}/2} \Phi\left( \dfrac{\ln{\bar{P}} - \ln{p} - \mu_c^{(p)} - \sigma_c^{(p)2} }{\sigma_c^{(p)}} \right) - \bar{P} \Phi\left( \dfrac{\ln{\bar{P}} - \ln{p} - \mu_c^{(p)} }{\sigma_c^{(p)}} \right)
\Bigg\} \\
<& c_1 \Phi\left( \dfrac{\ln{\bar{I}} - \ln{i_0} - \mu_c^{(i)}}{\sigma_c^{(i)}} \right) \Bigg\{ p^+ e^{\mu_c^{(p)} + \sigma_c^{(p)2}/2} \Phi\left( \dfrac{\ln{\bar{P}} - \ln{p} - \mu_c^{(p)} - \sigma_c^{(p)2} }{\sigma_c^{(p)}} \right) + \bar{P} \Phi\left( \dfrac{\ln{\bar{P}} - \ln{p} - \mu_c^{(p)} }{\sigma_c^{(p)}} \right) \\
&- p e^{\mu_c^{(p)} + \sigma_c^{(p)2}/2} \Phi\left( \dfrac{\ln{\bar{P}} - \ln{p} - \mu_c^{(p)} - \sigma_c^{(p)2} }{\sigma_c^{(p)}} \right) - \bar{P} \Phi\left( \dfrac{\ln{\bar{P}} - \ln{p} - \mu_c^{(p)} }{\sigma_c^{(p)}} \right)
\Bigg\} \\
=& c_1 (p^+ - p) e^{\mu_c^{(p)} + \sigma_c^{(p)2}/2} \Phi\left( \dfrac{\ln{\bar{I}} - \ln{i_0} - \mu_c^{(i)}}{\sigma_c^{(i)}} \right) \Phi\left( \dfrac{\ln{\bar{P}} - \ln{p} - \mu_c^{(p)} - \sigma_c^{(p)2} }{\sigma_c^{(p)}} \right),
\end{align*}
where the inequality holds because $i_0 \le i$ and $p < p^+$. Therefore, we can simplify the inequality \eqref{eq.pf_lemma_1} as
\begin{equation*}
\dfrac{1}{\gamma} < e^{\mu_c^{(p)} + \sigma_c^{(p)2}/2} \Phi\left( \dfrac{\ln{\bar{I}} - \ln{i_0} - \mu_c^{(i)}}{\sigma_c^{(i)}} \right) \Phi\left( \dfrac{\ln{\bar{P}} - \ln{p} - \mu_c^{(p)} - \sigma_c^{(p)2} }{\sigma_c^{(p)}} \right).
\end{equation*}
However, this contradicts for $p \ge \underline{p}$ with $\underline{p}$ given in Proposition~\ref{prop: myopic_under_PI}. Thus, we know that the result (i) holds.

Next, we need to show the result (ii) also holds. We will again use a contradiction as the proof technique. So, for the moment, suppose that the result (ii) does not hold, i.e.,
\begin{align*}
r_h(p, i) \ge -c_u +  \gamma \, \mbox{E}[R(p^\prime, i^\prime; H) | p, i], \\
r_h(p, i^+) \le -c_u +  \gamma \, \mbox{E}[R(p^\prime, i^\prime; H) | p, i^+],
\end{align*}
implying that  
\begin{equation}
\dfrac{r_h(p, i) - r_h(p, i^+)}{\gamma} \ge \mbox{E}[R(p^\prime, i^\prime; H) | p, i] - \mbox{E}[R(p^\prime, i^\prime; H) | p, i^+]. \label{eq.pf_lemma_2}
\end{equation}
The left-hand side of \eqref{eq.pf_lemma_2} is
\begin{equation*}
{\rm LHS} = \dfrac{c_2}{\gamma}(i^+ - i).
\end{equation*}
On the other hand, it holds for the right-hand side of \eqref{eq.pf_lemma_2} that
\begin{align*}
&{\rm RHS} = r_f [\mbox{Pr}(i^\prime \geq \bar{I} | i^+) - \mbox{Pr}(i^\prime \geq \bar{I} | i)] +
\left( c_0 + c_1 \mbox{E}[p^\prime | p] \right) [\mbox{Pr}(i^\prime < \bar{I} | i) - \mbox{Pr}(i^\prime < \bar{I} | i^+)]  \\
& - c_2 \mbox{E}[i^\prime |i^\prime < \bar{I}, i] \mbox{Pr}(i^\prime < \bar{I} | i) + c_2 \mbox{E}[i^\prime | i^\prime < \bar{I}, i^+] \mbox{Pr}(i^\prime < \bar{I} | i^+) \\
>& r_f \left[ \Phi\left( \dfrac{\ln{\bar{I}} - \ln{i} - \mu_c^{(i)}}{\sigma_c^{(i)}} \right) - \Phi\left( \dfrac{\ln{\bar{I}} - \ln{i^+} - \mu_c^{(i)}}{\sigma_c^{(i)}} \right) \right] \\
& - c_2 i e^{\mu_c^{(i)} + \sigma_c^{(i)2}/2} \Phi\left( \dfrac{\ln{\bar{I}} - \ln{i} - \mu_c^{(i)} - \sigma_c^{(i)2} }{\sigma_c^{(i)}} \right) + 
c_2 i^+ e^{\mu_c^{(i)} + \sigma_c^{(i)2}/2} \Phi\left( \dfrac{\ln{\bar{I}} - \ln{i^+} - \mu_c^{(i)} - \sigma_c^{(i)2} }{\sigma_c^{(i)}} \right) \\
>&  r_f \left[ \Phi\left( \dfrac{\ln{\bar{I}} - \ln{i} - \mu_c^{(i)}}{\sigma_c^{(i)}} \right) - \Phi\left( \dfrac{\ln{\bar{I}} - \ln{i^+} - \mu_c^{(i)}}{\sigma_c^{(i)}} \right) \right] \\
& - c_2 i e^{\mu_c^{(i)} + \sigma_c^{(i)2}/2} \left[ \Phi\left( \dfrac{\ln{\bar{I}} - \ln{i} - \mu_c^{(i)} - \sigma_c^{(i)2} }{\sigma_c^{(i)}} \right) - \Phi\left( \dfrac{\ln{\bar{I}} - \ln{i^+} - \mu_c^{(i)} - \sigma_c^{(i)2} }{\sigma_c^{(i)}} \right) \right] \\
\ge&  r_f \left[ \Phi\left( \dfrac{\ln{\bar{I}} - \ln{i} - \mu_c^{(i)}}{\sigma_c^{(i)}} \right) - \Phi\left( \dfrac{\ln{\bar{I}} - \ln{i^+} - \mu_c^{(i)}}{\sigma_c^{(i)}} \right) \right] \\
& - c_2 \bar{I} e^{\mu_c^{(i)} + \sigma_c^{(i)2}/2} \left[ \Phi\left( \dfrac{\ln{\bar{I}} - \ln{i} - \mu_c^{(i)} - \sigma_c^{(i)2} }{\sigma_c^{(i)}} \right) - \Phi\left( \dfrac{\ln{\bar{I}} - \ln{i^+} - \mu_c^{(i)} - \sigma_c^{(i)2} }{\sigma_c^{(i)}} \right) \right],
\end{align*}
where the first inequality holds because $\left( c_0 + c_1 \mbox{E}[p^\prime | p] \right) [\mbox{Pr}(i^\prime \le \bar{I} | i) - \mbox{Pr}(i^\prime \le \bar{I} | i^+)]>0$, 
the second inequality holds since $i<i^+$, and the last inequality holds since $i \le \bar{I}$. After plugging in the expressions of LHS and RHS, we have
\begin{align*}
\dfrac{c_2}{\gamma}(i^+ - i) >& r_f \left[ \Phi\left( \dfrac{\ln{\bar{I}} - \ln{i} - \mu_c^{(i)}}{\sigma_c^{(i)}} \right) - \Phi\left( \dfrac{\ln{\bar{I}} - \ln{i^+} - \mu_c^{(i)}}{\sigma_c^{(i)}} \right) \right] \nonumber\\
&- c_2 \bar{I} e^{\mu_c^{(i)} + \sigma_c^{(i)2}/2} \left[ \Phi\left( \dfrac{\ln{\bar{I}} - \ln{i} - \mu_c^{(i)} - \sigma_c^{(i)2} }{\sigma_c^{(i)}} \right) - \Phi\left( \dfrac{\ln{\bar{I}} - \ln{i^+} - \mu_c^{(i)} - \sigma_c^{(i)2} }{\sigma_c^{(i)}} \right) \right].
\end{align*}
However, this contradicts with the condition given in Proposition~\ref{prop: myopic_under_PI}. Thus, we know that the result (ii) also holds. Hence the proof is complete.

\noindent \textbf{Derivation of Inequality~\eqref{eq:sec4lem2approx}}

We let
\[
x = \dfrac{\ln{\bar{I}} - \ln{i} - \mu_c^{(i)}}{\sigma_c^{(i)}},~~x^+ = \dfrac{\ln{\bar{I}} - \ln{i^+} - \mu_c^{(i)}}{\sigma_c^{(i)}},
\]
where $i^+ > i$, $x > x^+$. Then we can rewrite the inequality in \eqref{eq:sec4lem2} as
\begin{equation}
\dfrac{c_2\bar{I}}{\gamma}(e^{-\sigma_c^{(i)}x^+ - \mu_c^{(i)}} - e^{-\sigma_c^{(i)}x - \mu_c^{(i)}}) \le 
r_f[\Phi(x) - \Phi(x^+)] - c_2 \bar{I} e^{\mu_c^{(i)} + \sigma_c^{(i)2}/2} [\Phi(x-\sigma_c^{(i)}) - \Phi(x^+-\sigma_c^{(i)})].
\label{eq:condlem2}
\end{equation}
We apply the first-order Taylor series expansion on each exponential term in the left-hand side and on the normal cumulative distribution functions in the right-hand side of \eqref{eq:condlem2}, and further approximate the expansion by only taking the dominant elements, for example
\begin{align*}
e^{-\sigma_c^{(i)}x - \mu_c^{(i)}} = \sum_{n=0}^{\infty} \dfrac{(-\sigma_c^{(i)}x - \mu_c^{(i)})^n}{n!} \approx 1 -\sigma_c^{(i)}x - \mu_c^{(i)}
\end{align*}
and 
\begin{align*}
\Phi(x) &= \dfrac{1}{\sqrt{2\pi}} \int_{-\infty}^x e^{-z^2/2}dz = \dfrac{1}{2} + \dfrac{1}{2}\mbox{erf}(\dfrac{x}{\sqrt{2}}) = \dfrac{1}{2} + \dfrac{1}{2} \cdot \dfrac{2}{\sqrt{\pi}} \sum_{n=0}^{\infty} \dfrac{(-1)^n (x/\sqrt{2})^{2n+1})}{n!(2n+1)} \approx \dfrac{1}{2} + \dfrac{x}{\sqrt{2\pi}}
\end{align*}
where 
$\mbox{erf}(w) = \dfrac{2}{\sqrt{\pi}} \int_{0}^w e^{-z^2}dz$ is the error function. Similar approximation can be applied to $e^{-\sigma_c^{(i)}x^+ - \mu_c^{(i)}}$, $\Phi(x^+)$, $\Phi(x-\sigma_c^{(i)})$ and $\Phi(x^+ - \sigma_c^{(i)})$. Thus, we can approximate \eqref{eq:condlem2} as 
\begin{align}
\dfrac{c_2\bar{I}}{\gamma}\sigma_c^{(i)}(x - x^+) \le & \dfrac{r_f}{\sqrt{2\pi}} (x - x^+) - \dfrac{c_2 \bar{I}}{\sqrt{2\pi}} e^{\mu_c^{(i)} + \sigma_c^{(i)2}/2} [(x - \sigma_c^{(i)}) - (x^+ - \sigma_c^{(i)})], \nonumber 
\end{align}
which leads to
\begin{align}
\dfrac{c_2\bar{I}}{\gamma} \sigma_c^{(i)} (x - x^+) \le & \left[ \dfrac{r_f}{\sqrt{2\pi}} - \dfrac{c_2 \bar{I}}{\sqrt{2\pi}} e^{\mu_c^{(i)} + \sigma_c^{(i)2}/2} \right] (x - x^+). \nonumber 
\end{align}
Therefore, it follows that:
\begin{align*}
\dfrac{r_f}{\sqrt{2\pi}} \ge & \dfrac{c_2\bar{I}}{\gamma} \sigma_c^{(i)} 
+ \dfrac{c_2 \bar{I}}{\sqrt{2\pi}} e^{\mu_c^{(i)} + \sigma_c^{(i)2}/2} \nonumber \\
r_f \ge & c_2\bar{I} \left[ \dfrac{\sqrt{2\pi} \sigma_c^{(i)}}{\gamma} + e^{\mu_c^{(i)} + \sigma_c^{(i)2}/2} \right]. 
\end{align*}

\noindent \textbf{Proof of Proposition~\ref{theo:boudary_move_mean}}

\noindent (i)
We start with writing the function $\widetilde{h}(\alpha^{(p)}, \widetilde{\sigma}^{(p)}, \alpha^{(i)}, \widetilde{\sigma}^{(i)}; p, i)$ explicitly:
\begin{align}
\widetilde{h}(\alpha^{(p)}, \widetilde{\sigma}^{(p)}, \alpha^{(i)}, \widetilde{\sigma}^{(i)}; p, i) = & c_0 + c_u + c_1p - c_2i + \gamma r_f \left[ 1 - \Phi\left( \dfrac{\ln{\bar{I}} - \ln{i} - \alpha^{(i)}}{\widetilde{\sigma}^{(i)}} \right) \right] \nonumber \\
& - \gamma c_0 \Phi\left( \dfrac{\ln{\bar{I}} - \ln{i} - \alpha^{(i)}}{\widetilde{\sigma}^{(i)}} \right) \nonumber \\
& - \gamma c_1 \bar{P} \left[ 1 - \Phi\left( \dfrac{\ln{\bar{P}} - \ln{p} - \alpha^{(p)}}{\widetilde{\sigma}^{(p)}} \right) \right] \Phi\left( \dfrac{\ln{\bar{I}} - \ln{i} - \alpha^{(i)}}{\widetilde{\sigma}^{(i)}} \right) \nonumber \\
& - \gamma c_1p e^{\alpha^{(p)} + \widetilde{\sigma}^{(p)2}/2} \Phi\left( \dfrac{\ln{\bar{P}} - \ln{p} - \alpha^{(p)} - \widetilde{\sigma}^{(p)2} }{\widetilde{\sigma}^{(p)}} \right) \Phi\left( \dfrac{\ln{\bar{I}} - \ln{i} - \alpha^{(i)}}{\widetilde{\sigma}^{(i)}} \right) \nonumber \\
& + \gamma c_2i e^{\alpha^{(i)} + \widetilde{\sigma}^{(i)2}/2} \Phi\left( \dfrac{\ln{\bar{I}} - \ln{i} - \alpha^{(i)} - \widetilde{\sigma}^{(i)2} }{\widetilde{\sigma}^{(i)}} \right) \nonumber \\
= & c_0 + c_u + c_1p - c_2i + \gamma r_f [1- \Phi(x)] - \gamma c_0 \Phi(x) - \gamma c_1 \bar{P} [1-\Phi(y)]\Phi(x) \nonumber \\
& - \gamma c_1p e^{\alpha^{(p)} + \widetilde{\sigma}^{(p)2}/2} \Phi(y - \widetilde{\sigma}^{(p)}) \Phi(x) 
+ \gamma c_2i e^{\alpha^{(i)} + \widetilde{\sigma}^{(i)2}/2} \Phi(x - \widetilde{\sigma}^{(i)}), \label{eq.h_replaced} \nonumber
\end{align}
where the $x$ and $y$ are variable replacements defined as
\[
x = \dfrac{\ln{\bar{I}} - \ln{i} - \alpha^{(i)}}{\widetilde{\sigma}^{(i)}},~~y = \dfrac{\ln{\bar{P}} - \ln{p} - \alpha^{(p)}}{\widetilde{\sigma}^{(p)}}.
\]
We know that:
\begin{eqnarray}
\lefteqn{i e^{\alpha^{(i)} + \widetilde{\sigma}^{(i)2}/2} \phi(x - \widetilde{\sigma}^{(i)}) 
= i e^{\alpha^{(i)} + \widetilde{\sigma}^{(i)2}/2} \dfrac{e^{-(x - \widetilde{\sigma}^{(i)})^2/2}}{\sqrt{2\pi}} }
\nonumber \\
&=& i e^{\alpha^{(i)} + \widetilde{\sigma}^{(i)2}/2} \dfrac{e^{-x^2/2 +x\widetilde{\sigma}^{(i)} -\widetilde{\sigma}^{(i)2}/2}}{\sqrt{2\pi}} 
\nonumber \\
&=& i e^{\alpha^{(i)} + x\widetilde{\sigma}^{(i)}} \dfrac{e^{-x^2/2}}{\sqrt{2\pi}} 
\nonumber \\
&=& i e^{\ln{\bar{I}} - \ln{i}} \phi(x) 
\label{eq.mid100} \\
&=& \bar{I} \phi(x),  \nonumber
\end{eqnarray}
where the equality  \eqref{eq.mid100} follows by plugging in 
$x = \dfrac{\ln{\bar{I}} - \ln{i} - \alpha^{(i)}}{\widetilde{\sigma}^{(i)}}$.
Similarly, we have
\[
    p e^{\alpha^{(p)} + \widetilde{\sigma}^{(p)2}/2} \phi(y - \widetilde{\sigma}^{(p)}) = \bar{P} \phi(y).
\]

For function $\widetilde{h}(\alpha^{(p)}, \widetilde{\sigma}^{(p)}, \alpha^{(i)}, \widetilde{\sigma}^{(i)}; p, i)$,
if we only consider the parts related to $\alpha^{(i)}$, we have
\begin{align}
\widetilde{h}(\alpha^{(p)}, \widetilde{\sigma}^{(p)}, \alpha^{(i)}, \widetilde{\sigma}^{(i)}; p, i) \propto & - r_f \Phi(x) - c_0 \Phi(x) - c_1 \bar{P} [1-\Phi(y)]\Phi(x) \nonumber \\
& - c_1p e^{\alpha^{(p)} + \widetilde{\sigma}^{(p)2}/2} \Phi(y - \widetilde{\sigma}^{(p)}) \Phi(x) 
+ c_2i e^{\alpha^{(i)} + \widetilde{\sigma}^{(i)2}/2} \Phi(x - \widetilde{\sigma}^{(i)}), \label{eq:h_to_impurity_mean}
\end{align}
Taking derivatives with respect to $\alpha^{(i)}$ leads to:
\begin{align}
\dfrac{\partial \widetilde{h}}{\partial \alpha^{(i)}} \propto~& -C\phi(x)\dfrac{\partial x}{\partial \alpha^{(i)}} + c_2i e^{\alpha^{(i)} + \widetilde{\sigma}^{(i)2}/2} \Phi(x - \widetilde{\sigma}^{(i)})
+ c_2i e^{\alpha^{(i)} + \widetilde{\sigma}^{(i)2}/2} \phi(x - \widetilde{\sigma}^{(i)})\dfrac{\partial x}{\partial \alpha^{(i)}} \nonumber \\
\propto~& C\phi(x)\dfrac{1}{\widetilde{\sigma}^{(i)}} + c_2i e^{\alpha^{(i)} + \widetilde{\sigma}^{(i)2}/2} \Phi(x - \widetilde{\sigma}^{(i)})
- c_2\bar{I} \phi(x)\dfrac{1}{\widetilde{\sigma}^{(i)}} \nonumber \\
\propto~& \left(C - c_2\bar{I} \right)\dfrac{\phi(x)}{\widetilde{\sigma}^{(i)}} + c_2i e^{\alpha^{(i)} + \widetilde{\sigma}^{(i)2}/2} \Phi(x - \widetilde{\sigma}^{(i)}), \nonumber
\end{align}
where
\[
C = r_f + c_0 + c_1 \bar{P} [1-\Phi(y)] + c_1p e^{\alpha^{(p)} + \widetilde{\sigma}^{(p)2}/2} \Phi(y - \widetilde{\sigma}^{(p)}).
\]
Since $r_f \ge c_2\bar{I}$, we have $C - c_2\bar{I} > 0$, then $\dfrac{\partial \widetilde{h}}{\partial \alpha^{(i)}} > 0$.

\noindent (ii)
On the other hand, if we only consider the parts related to $\alpha^{(p)}$, we have
\begin{align}
\widetilde{h}(\alpha^{(p)}, \widetilde{\sigma}^{(p)}, \alpha^{(i)}, \widetilde{\sigma}^{(i)}; p, i) \propto &~ c_1 \bar{P} \Phi(y)\Phi(x) 
- c_1 e^{\alpha^{(p)} + \widetilde{\sigma}^{(p)2}/2} \Phi(y - \widetilde{\sigma}^{(p)}) \Phi(x) \nonumber \\
\propto &~ \bar{P} \Phi(y) - p e^{\alpha^{(p)} + \widetilde{\sigma}^{(p)2}/2} \Phi(y - \widetilde{\sigma}^{(p)}).
\label{eq:h_to_protein_mean}
\end{align}
Taking derivatives with respect to $\alpha^{(p)}$ leads to:
\begin{align*}
\dfrac{\partial \widetilde{h}}{\partial \alpha^{(i)}} \propto~& \bar{P} \phi(y)\dfrac{\partial y}{\partial \alpha^{(p)}} - p e^{\alpha^{(p)} + \widetilde{\sigma}^{(p)2}/2} \Phi(y - \widetilde{\sigma}^{(p)}) 
- p e^{\alpha^{(p)} + \widetilde{\sigma}^{(p)2}/2} \phi(y - \widetilde{\sigma}^{(p)}) \dfrac{\partial y}{\partial \alpha^{(p)}} \\
\propto~& \bar{P} \phi(y)\dfrac{\partial y}{\partial \alpha^{(p)}} - p e^{\alpha^{(p)} + \widetilde{\sigma}^{(p)2}/2} \Phi(y - \widetilde{\sigma}^{(p)}) 
- \bar{P} \phi(y)\dfrac{\partial y}{\partial \alpha^{(p)}} \\
\propto~& - p e^{\alpha^{(p)} + \widetilde{\sigma}^{(p)2}/2} \Phi(y - \widetilde{\sigma}^{(p)}) < 0.
\end{align*}



\noindent (iii)
If we only consider the parts related to $\widetilde{\sigma}^{(i)}$, we have exactly the same formula of $\widetilde{h}(\alpha^{(p)}, \widetilde{\sigma}^{(p)}, \alpha^{(i)}, \widetilde{\sigma}^{(i)}; p, i)$ as in \eqref{eq:h_to_impurity_mean}. 
Taking derivatives with respect to $\widetilde{\sigma}^{(i)}$, we have that
\begin{align}
\dfrac{\partial \widetilde{h}}{\partial \widetilde{\sigma}^{(i)}} \propto~& -C\phi(x)\dfrac{\partial x}{\partial \widetilde{\sigma}^{(i)}} + c_2i \widetilde{\sigma}^{(i)} e^{\alpha^{(i)} + \widetilde{\sigma}^{(i)2}/2} \Phi(x - \widetilde{\sigma}^{(i)})
+ c_2i e^{\alpha^{(i)} + \widetilde{\sigma}^{(i)2}/2} \phi(x - \widetilde{\sigma}^{(i)})\left( \dfrac{\partial x}{\partial \widetilde{\sigma}^{(i)}} - 1 \right) \nonumber \\
\propto~& -C\phi(x)\dfrac{\partial x}{\partial \widetilde{\sigma}^{(i)}} + c_2i \widetilde{\sigma}^{(i)} e^{\alpha^{(i)} + \widetilde{\sigma}^{(i)2}/2} \Phi(x - \widetilde{\sigma}^{(i)})
+ c_2\bar{I} \phi(x)\left( \dfrac{\partial x}{\partial \widetilde{\sigma}^{(i)}} - 1 \right) \label{eq.theo5.2} \nonumber \\
\propto~& -\left(C - c_2\bar{I} \right)\phi(x)\dfrac{\partial x}{\partial \widetilde{\sigma}^{(i)}} + c_2i e^{\alpha^{(i)} + \widetilde{\sigma}^{(i)2}/2} \left[ \widetilde{\sigma}^{(i)} \Phi(x - \widetilde{\sigma}^{(i)}) -  \phi(x - \widetilde{\sigma}^{(i)}) \right]. \nonumber
\end{align}
Similarly, we have $C - c_2\bar{I} > 0$. Therefore, if the conditions in Theorem~\ref{theo:boudary_move_mean}(iii) hold, then $\dfrac{\partial x}{\partial \widetilde{\sigma}^{(i)}} < 0$, and $\widetilde{\sigma}^{(i)} \Phi(x - \widetilde{\sigma}^{(i)}) -  \phi(x - \widetilde{\sigma}^{(i)}) > 0$. Thus, $\dfrac{\partial \widetilde{h}}{\partial \widetilde{\sigma}^{(i)}} > 0$.



\noindent (iv)
Further, if we only consider the parts related to $\widetilde{\sigma}^{(p)}$, we will have the same formula as in \eqref{eq:h_to_protein_mean}.
Taking derivatives with respect to $\widetilde{\sigma}^{(p)}$, we have the following holds, 
\begin{align*}
\dfrac{\partial \widetilde{h}}{\partial \widetilde{\sigma}^{(p)}} \propto~& \bar{P} \phi(y)\dfrac{\partial y}{\partial \widetilde{\sigma}^{(p)}} - p \widetilde{\sigma}^{(p)} e^{\alpha^{(p)} + \widetilde{\sigma}^{(p)2}/2} \Phi(y - \widetilde{\sigma}^{(p)}) 
- p e^{\alpha^{(p)} + \widetilde{\sigma}^{(p)2}/2} \phi(y - \widetilde{\sigma}^{(p)}) \left( \dfrac{\partial y}{\partial \widetilde{\sigma}^{(p)}} - 1 \right) \\
\propto~& \bar{P} \phi(y)\dfrac{\partial y}{\partial \widetilde{\sigma}^{(p)}} - p \widetilde{\sigma}^{(p)} e^{\alpha^{(p)} + \widetilde{\sigma}^{(p)2}/2} \Phi(y - \widetilde{\sigma}^{(p)}) - \bar{P} \phi(y)\dfrac{\partial y}{\partial \widetilde{\sigma}^{(p)}}
+ p e^{\alpha^{(p)} + \widetilde{\sigma}^{(p)2}/2} \phi(y - \widetilde{\sigma}^{(p)}) \\
\propto~& p e^{\alpha^{(p)} + \widetilde{\sigma}^{(p)2}/2} \left[ \phi(y - \widetilde{\sigma}^{(p)}) - \widetilde{\sigma}^{(p)} \Phi(y - \widetilde{\sigma}^{(p)}) \right].
\end{align*}
Since $p e^{\alpha^{(p)} + \widetilde{\sigma}^{(p)2}/2} > 0$, we have $\dfrac{\partial \widetilde{h}}{\partial \widetilde{\sigma}^{(p)}} < 0$ if and only if $\widetilde{\sigma}^{(p)} \Phi(y - \widetilde{\sigma}^{(p)}) - \phi(y - \widetilde{\sigma}^{(p)}) > 0$.

Therefore, given the current physical state $(p_t, i_t)$, the optimal decision is to harvest if and only if $r_0(p_t,i_t) \ge r_m(p_t,i_t)$ for all $m \in \{1,2,\ldots, \bar{T}-t\}$. Otherwise, the optimal decision is to continue the fermentation for this batch. 

\noindent \textbf{Proof of Proposition~\ref{theo:boundary_consistency}}
\label{sec:pf_theo:boundary_consistency}

Define the ``harvest function" of myopic policy under perfect information as following,
\begin{align*}
h(\mu^{(p)}_c, \sigma^{(p)}_c, \mu^{(i)}_c, \sigma^{(i)}_c; p, i) = & r_h(p, i) + c_u -  \gamma  \mbox{E}[R(p^\prime, i^\prime; H) | p, i] \\
& r_h(p, i) + c_u -  \gamma \Big\{ -r_f \mbox{Pr}(i^\prime > \bar{I} | i, \mu^{(i)}_c, \sigma^{(i)}_c) \\
& + \mbox{E}[r_h(p^\prime, i^\prime) | i^\prime \le \bar{I}, p, i, \mu^{(p)}_c, \sigma^{(p)}_c, \mu^{(i)}_c, \sigma^{(i)}_c] \Big\}
\end{align*}
then the harvest region in \eqref{eq:myopic_policy} can be written as $A = \left\{ (p, i): h(p, i; \mu^{(p)}_c, \sigma^{(p)2}_c, \mu^{(i)}_c, \sigma^{(i)2}_c) \ge 0 \right\}$. Notice that under model risk,
\begin{align*}
\widetilde{h}(\alpha^{(p)}, \widetilde{\sigma}^{(p)}, \alpha^{(i)}, \widetilde{\sigma}^{(i)}; p, i) = & r_h(p, i) + c_u -  \gamma  \mbox{E}[R(p^\prime, i^\prime; H) | p, i, \alpha^{(p)}, \widetilde{\sigma}^{(p)}, \alpha^{(i)}, \widetilde{\sigma}^{(i)}] \\
& r_h(p, i) + c_u -  \gamma \Big\{ -r_f \mbox{Pr}(i^\prime > \bar{I} | i, \alpha^{(i)}, \widetilde{\sigma}^{(i)}) \\
& + \mbox{E}[r_h(p^\prime, i^\prime) | i^\prime \le \bar{I}, p, i, \alpha^{(p)}, \widetilde{\sigma}^{(p)}, \alpha^{(i)}, \widetilde{\sigma}^{(i)}] \Big\},
\end{align*}
which can be obtained by replacing the unknown  parameters $\mu^{(i)}_c$ and $\sigma^{(i)2}_c$ with predictive mean $\alpha^{(i)}$ and predictive standard deviation $\widetilde{\sigma}^{(i)}$ for the impurity growth rate; and similarly, by replacing $\mu^{(p)}_c$ and $\sigma^{(p)2}_c$ with $\alpha^{(p)}$ and $\widetilde{\sigma}^{(p)}$ for the protein growth rate. Our objective is to show that the function $\widetilde{h}(\alpha_{t}^{(p)}, \widetilde{\sigma}_{t}^{(p)}, \alpha_{t}^{(i)}, \widetilde{\sigma}_{t}^{(i)}; p, i)$ converges to its true counterpart $h(\mu^{(p)}_c, \sigma^{(p)}_c, \mu^{(i)}_c, \sigma^{(i)}_c; p, i)$ as the length of the historical data increases. 

It is known that $\phi_j$, $j=1,2,\ldots,J_t$, are i.i.d. samples from the underlying true protein growth rate distribution $\mathcal{N}(\mu^{(p)}_c, \sigma^{(p)2}_c)$. According to the properties of Gaussian sample mean and sample variance, we have $\bar{\phi} \sim \mathcal{N}(\mu^{(p)}_c, \dfrac{\sigma^{(p)2}_c}{J_t})$, $\dfrac{\sum_{j=1}^{J_t}(\phi_j - \bar{\phi})^2}{\sigma^{(p)2}_c} \sim \chi^2(J_t-1)$, and they are independent of each other, where $\bar{\phi}=\frac{1}{J_t} \sum_{j=1}^{J_t}\phi_j$. 
According to the weak law of large numbers (WLLN), we have sample mean converge to the true mean, i.e., $\alpha_{t}^{(p)} \overset{p}{\to} \mu^{(p)}_c$ as $J_t \to \infty$. In addition, it is known from Proposition~\ref{theo:var_decomp}(i) that
\begin{equation}
\mbox{E}\left[\widetilde{\sigma}_{t}^{(p)2}\right] = \sigma^{(p)2}_c + \dfrac{(2J_t-1)\sigma^{(p)2}_c}{(J_t^2-2J_t)},~~
\mbox{Var}\left[\widetilde{\sigma}_{t}^{(p)2}\right] = \dfrac{2(J_t^3+J_t^2-J_t-1)\sigma^{(p)4}_c}{J_t^4 - 4J_t^3 + 4J_t^2}.
\end{equation}
As $J_t \to \infty$, we will have $\mbox{E}\left[\widetilde{\sigma}_{t}^{(p)2}\right] \to \sigma^{(p)2}_c$ and $\mbox{Var}\left[\widetilde{\sigma}_{t}^{(p)2}\right] \to 0$, so that $\widetilde{\sigma}_{t}^{(p)2} \overset{p}{\to} \sigma^{(p)2}_c$. Similar results also apply for impurity growth rate so that we also have $\alpha_{t}^{(i)} \overset{p}{\to} \mu^{(i)}_c$ and $\widetilde{\sigma}_{t}^{(i)2} \overset{p}{\to} \sigma^{(i)2}_c$. Thus, through the continuous mapping theorem, $\widetilde{\sigma}_{t}^{(p)} \overset{p}{\to} \sigma^{(p)}_c$ and $\widetilde{\sigma}_{t}^{(i)} \overset{p}{\to} \sigma^{(i)}_c$, and we have for any $(p_t,i_t) \in [p_0, \bar{P}] \times [i_0, \bar{I}]$ and $m=1,2,\ldots, \bar{T}-t$, $\widetilde{h}(\alpha_{t}^{(p)}, \widetilde{\sigma}_{t}^{(p)}, \alpha_{t}^{(i)}, \widetilde{\sigma}_{t}^{(i)}; p, i) \overset{p}{\to} {h}(\mu^{(p)}_c, \sigma^{(p)}_c, \mu^{(i)}_c, \sigma^{(i)}_c; p, i)$. In other words, the decision boundary under model risk will converge to the decision boundary under perfect information.

\end{document}